\documentclass[10pt,journal,compsoc]{IEEEtran}

\ifCLASSOPTIONcompsoc
  \usepackage[nocompress]{cite}
\else
  \usepackage{cite}
\fi

\ifCLASSOPTIONcompsoc
    \usepackage[caption=false, font=normalsize, labelfont=sf, textfont=sf]{subfig}
\else
\usepackage[caption=false, font=footnotesize]{subfig}
\fi

\usepackage{booktabs} 
\usepackage{algorithm}
\usepackage{algorithmic}
\usepackage{multirow}
\usepackage{amsfonts}
\usepackage{amsmath}
\usepackage{float}
\usepackage{bbm}
\usepackage{verbatim}
\usepackage{amsthm}
\usepackage[super]{nth}
\usepackage{graphicx}
\usepackage{caption}
\usepackage{color, colortbl}
\ifCLASSOPTIONcompsoc
  \usepackage[caption=false,font=normalsize,labelfont=sf,textfont=sf]{subfig}
\else
  \usepackage[caption=false,font=footnotesize]{subfig}
\fi

\theoremstyle{definition}
\newtheorem{definition}{Definition}

\definecolor{Gray}{gray}{0.9}

\begin{document}
\title{A Multi-view Multi-task Learning Framework for Multi-variate Time Series Forecasting}

\author{Jinliang~Deng,
        Xiusi~Chen,
        Renhe~Jiang,
        Xuan~Song,
        Ivor~W.~Tsang
\IEEEcompsocitemizethanks{\IEEEcompsocthanksitem J. Deng and I.W. Tsang are with Australian Artificial Intelligence Institute, University of Technology Sydney, Sydney, Australia. J. Deng is also affiliated with Department of Computer Science and Engineering, Southern University of Science and Technology, Shenzhen, China.
E-mail: jinliang.deng@student.uts.edu.au, Ivor.Tsang@uts.edu.au.
\IEEEcompsocthanksitem X. Chen is with University of California, Los Angeles, USA. Email: xchen@cs.ucla.edu.
\IEEEcompsocthanksitem R. Jiang is with Center for Spatial Information Science, University of Tokyo, Tokyo, Japan. Email: jiangrh@csis.u-tokyo.ac.jp.
\IEEEcompsocthanksitem X. Song is with SUSTech-UTokyo Joint Research Center on Super Smart City, Department of Computer Science and Engineering, Southern University of Science and Technology (SUSTech), Shenzhen, China. Email: songx@sustech.edu.cn.
}
\thanks{This work has been submitted to the IEEE for possible publication. Copyright may be transferred without notice, after which this version may no longer be accessible.}}

\IEEEtitleabstractindextext{%
\begin{abstract}
Multi-variate time series (MTS) data is a ubiquitous class of data abstraction in the real world. Any instance of MTS is generated from a hybrid dynamical system and their specific dynamics are usually unknown. The hybrid nature of such a dynamical system is a result of complex external attributes, such as geographic location and time of day, each of which can be categorized into either spatial attributes or temporal attributes. Therefore, there are two fundamental views which can be used to analyze MTS data, namely the spatial view and the temporal view. Moreover, from each of these two views, we can partition the set of data samples of MTS into disjoint forecasting tasks in accordance with their associated attribute values. Then, samples of the same task will manifest similar forthcoming pattern, which is less sophisticated to be predicted in comparison with the original single-view setting. Considering this insight, we propose a novel multi-view multi-task  (MVMT) learning framework for MTS forecasting. Instead of being explicitly presented in most scenarios, MVMT information is deeply concealed in the MTS data, which severely hinders the model from capturing it naturally. To this end, we develop two kinds of basic operations, namely task-wise affine transformation and task-wise normalization, respectively. Applying these two operations with prior knowledge on the spatial and temporal view allows the model to adaptively extract MVMT information while predicting. Extensive experiments on three datasets are conducted to illustrate that canonical architectures can be greatly enhanced by the MVMT learning framework in terms of both effectiveness and efficiency. In addition, we design rich case studies to reveal the properties of representations produced at different phases in the entire prediction procedure.
\end{abstract}

}

\maketitle

%

%

\maketitle

\section{Introduction}

Time series forecasting is a significant problem in many industrial and business applications \cite{9422199}. For instance, a public transport operator can allocate sufficient capacity to mitigate the queuing time in a region in advance, if they have the means to foresee that a particular region will suffer from a supply shortage in the next couple of hours \cite{geng2019spatiotemporal, 9096591, 9139357, 9063494}. As another example, an investor can avoid economic loss with the assistance of a robo-advisor which is able to predict a potential market crash \cite{ding2019modeling}. Due to the complex and continuous fluctuation of impacting factors, real-world time series tends to be extraordinarily non-stationary, that is, exhibiting diverse dynamics. For instance, traffic volume is largely affected by the road's condition, location, and the current time and weather condition. In the retail sector, the current season, price and brand are determinants for the sales of merchandise. The diverse dynamics impose an enormous challenge on time series forecasting. In this work, we study multi-variate time series forecasting, where multiple variables evolve with time.

\begin{figure}[tb]
    \centering
    \includegraphics[width=0.9\linewidth]{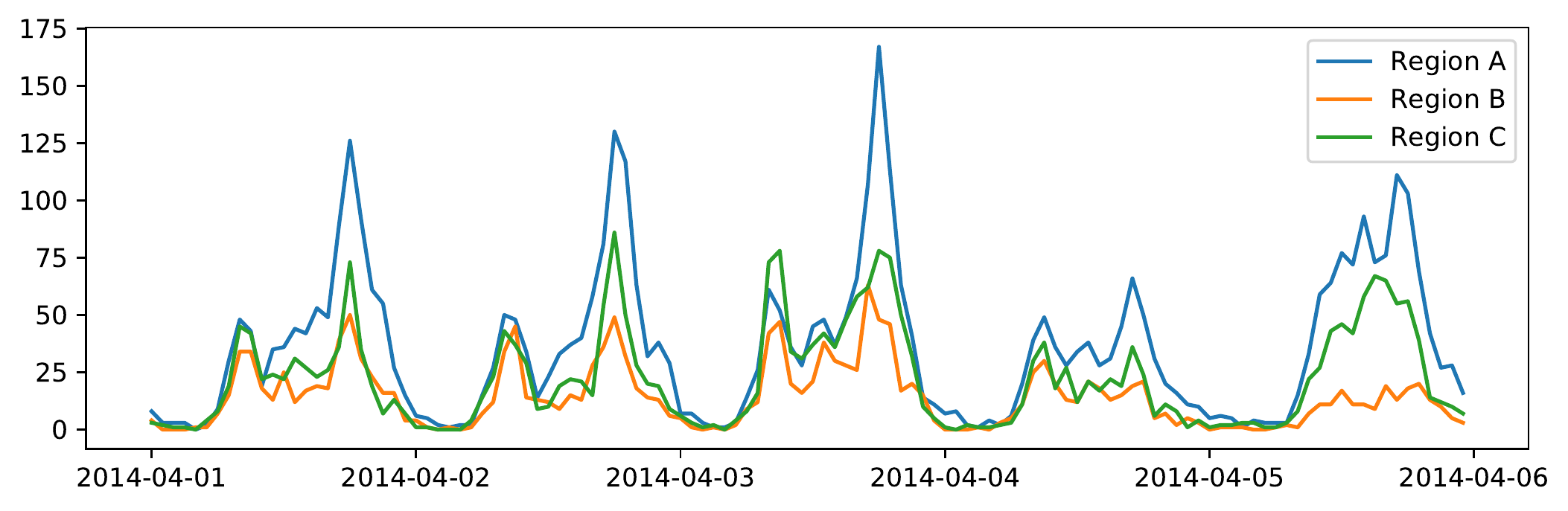}
    \caption{NYC shared bike demand.}
    \label{fig:evolution}
\end{figure}

Traditional time series forecasting algorithms, such as ARIMA and state space models (SSMs), provide a principled framework for modeling and learning time series patterns. However, these algorithms have a rigorous requirement for the stationarity of a time series, which suffer from severe limitations in practical use if most of the impacting factors are unavailable. With the recent advances in deep learning techniques, we are now capable of handling the complex dynamics as a single unit, even without any additional impacting factors. Common neural architectures applied on time series data include recurrent neural networks (RNNs), long-short term memory (LSTM) \cite{hochreiter1997long}, Transformer \cite{li2019enhancing}, Wavenet \cite{oord2016wavenet} and temporal convolution networks (TCNs) \cite{bai2018empirical}.

\begin{figure*}[thb]
\centering
\subfloat[]{
\includegraphics[width=0.32\linewidth]{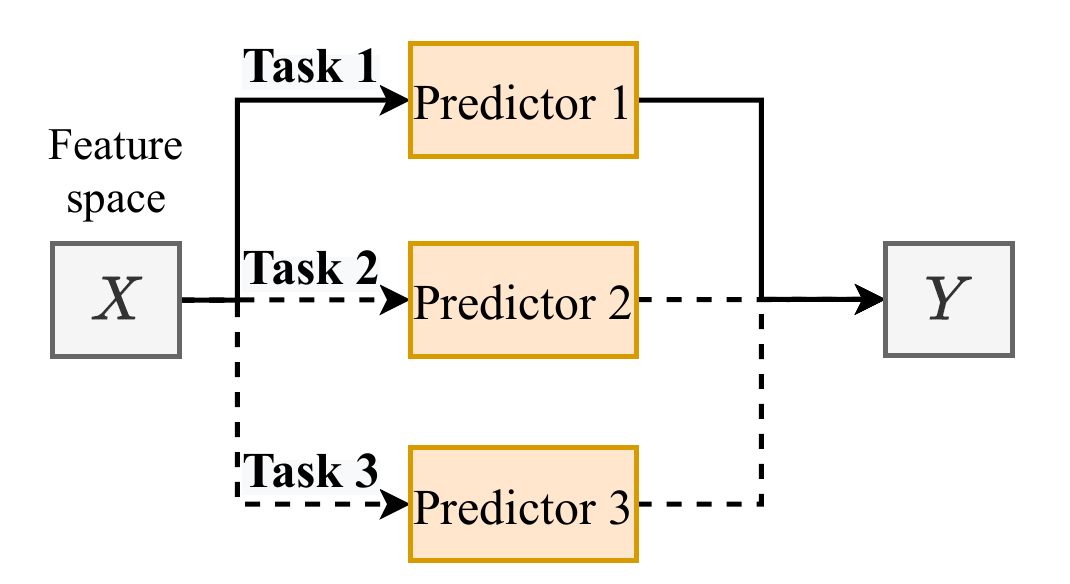}
\label{fig:mtl_1}
}
\subfloat[]{
\includegraphics[width=0.32\linewidth]{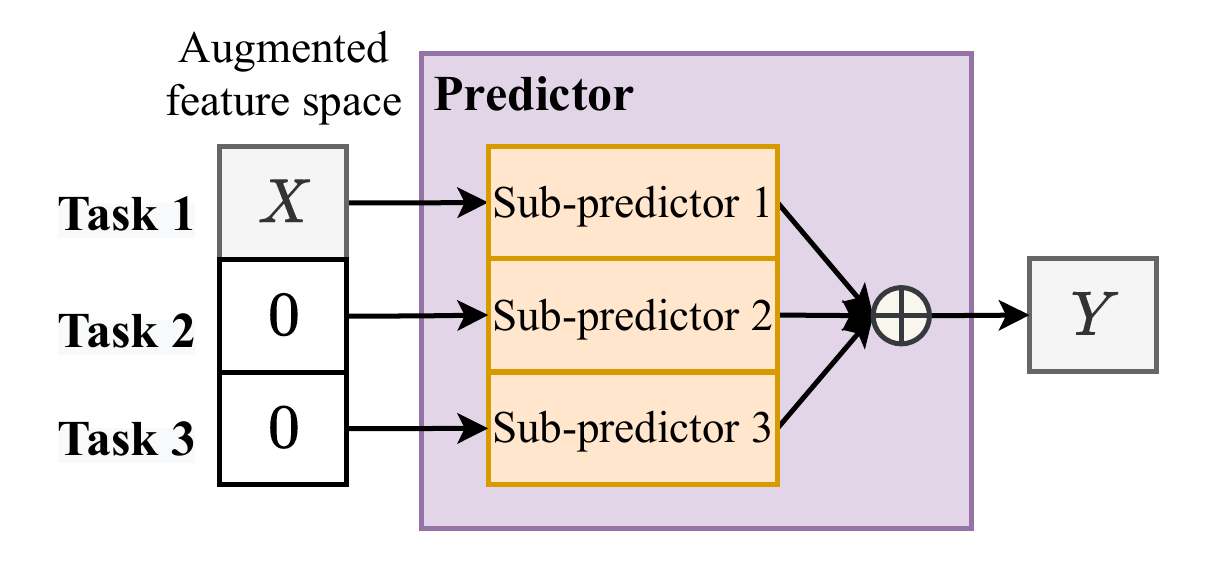}
\label{fig:mtl_2}
}
\subfloat[]{
\includegraphics[width=0.32\linewidth]{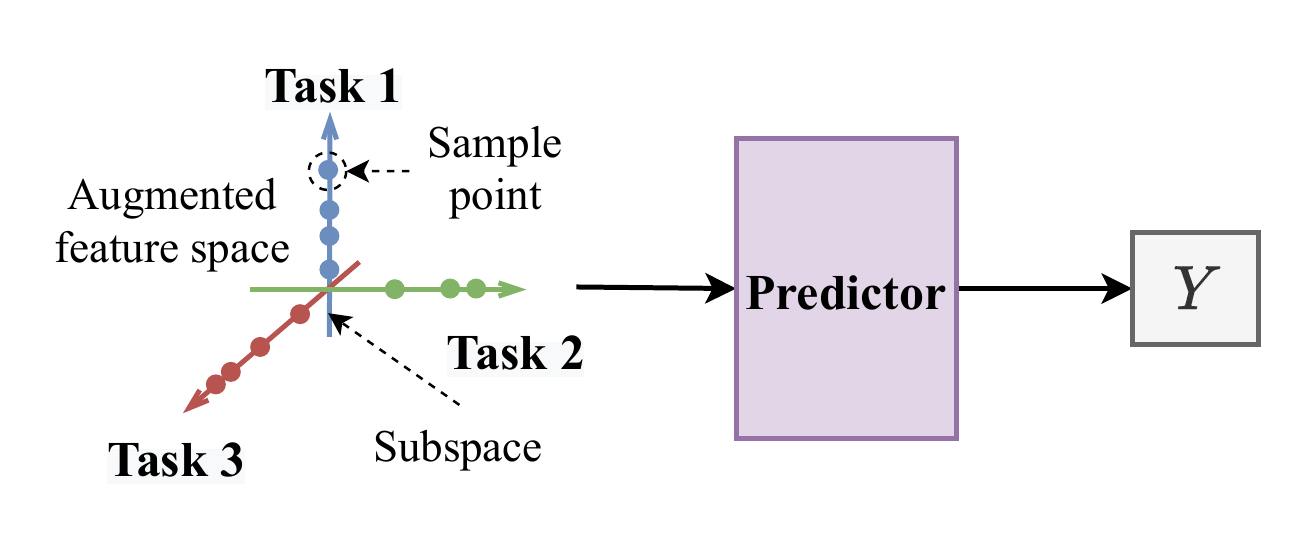}
\label{fig:mtl_3}
}
\caption{Three equivalent paradigms for for multi-task learning.}
\label{fig:issue}
\end{figure*}

In our article, we conjecture that MTS forecasting can essentially be treated as a multi-view multi-task learning problem. To the best of our knowledge, we are the first to formulate MTS in this way. There are typically two additional views in the MTS problem apart from the original spatial-temporal view, namely the temporal view and the spatial view. From each of these two views, we can divide the forecasting for samples into different tasks based on certain criteria. We take shared bike demand as a concrete example, and display the demand data collected from three regions over a five-day period in Figure \ref{fig:evolution}. In this example, from the temporal view, forecasting at a particular time over all the regions can be grouped into a task; from the spatial view, forecasting at all times over a particular region can be grouped into a task. Herein, the task partition scheme follows the principle where data points sampled from the same task are impacted by a common external factor, which makes them display common patterns. By handling each task with an exclusive predictor, the prediction complexity can be largely reduced in contrast with using a single predictor for all tasks. Sometimes, if the diversity of tasks suitably coincides with the diversity of dynamics, a linear regression model has the sufficient ability to undertake individual tasks.

\begin{figure}[tbh]
\centering
\subfloat[]{
\includegraphics[width=0.31\linewidth]{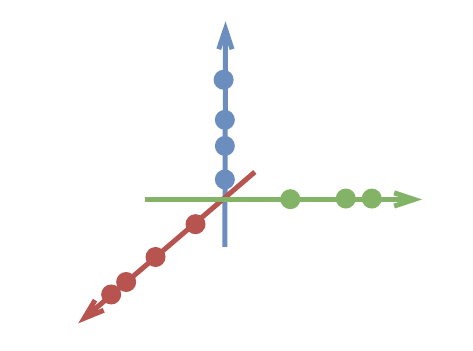}
}
\subfloat[]{
\includegraphics[width=0.31\linewidth]{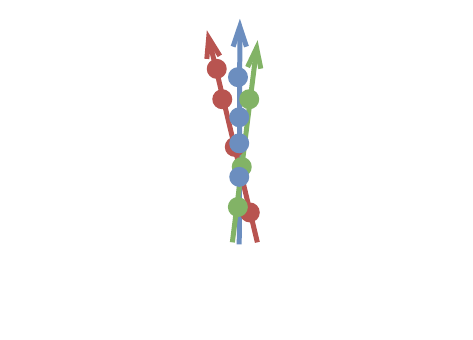}
}
\subfloat[]{
\includegraphics[width=0.31\linewidth]{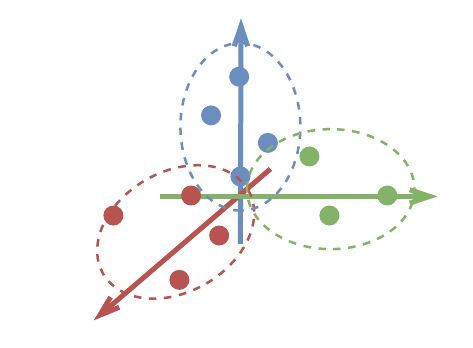}
}
\caption{(a) Desirable feature space. (b), (c) Undesirable feature spaces, where (b) has a strong inter-task correlation, and (c) has a weak intra-task correlation.}
\label{fig:geom}
\end{figure}

To achieve multi-task learning given any view, the key is to produce a feature space manifesting \textbf{inter-task weak-correlation} and \textbf{intra-task strong-correlation}. To make this idea more comprehensible, we start by displaying the canonical paradigm of multi-task learning in Fig. \ref{fig:mtl_1} with its two equivalent derivatives in Fig. \ref{fig:mtl_2} and Fig. \ref{fig:mtl_3}. In Fig. \ref{fig:mtl_2}, we create an augmented feature space with the number of dimensions being as many as three times that of the original ones. The augmented feature space is equally partitioned into three subspace where each subspace is associated with a task. Given a sample from any task, we let the corresponding subspace of its belonging task accommodate its features and the other two subspace padded with 0. It is easy to testify the equivalence between Fig. \ref{fig:mtl_1} and Fig. \ref{fig:mtl_2}. Next, let we have a closer look at Fig. \ref{fig:mtl_2}. An immediate judgement can be made from this formulation that samples from different tasks are orthogonal to each other in the augmented feature space, and samples from the same task maintain their relative positions as in the original feature space. More generally, even if orthogonality is not rigorously satisfied, each task can be captured in a more individual way provided that the correlations between different tasks diminish. In a further step, we can deduce that given the condition that the inter-task weak correlation and the intra-task strong correlation are manifested in the feature space, the predictor will automatically differentiate the task identity of every given sample. Therefore, all we need is an augmented feature space encoding the two types of relationships as shown in Figure \ref{fig:mtl_3}. In addition, we display two kinds of undesirable geometries in Figure \ref{fig:geom} to highlight the key properties of the geometry which fits the multi-task learning paradigm.

\begin{figure}[tb]
\centering
\subfloat[]{
\includegraphics[width=0.4\linewidth]{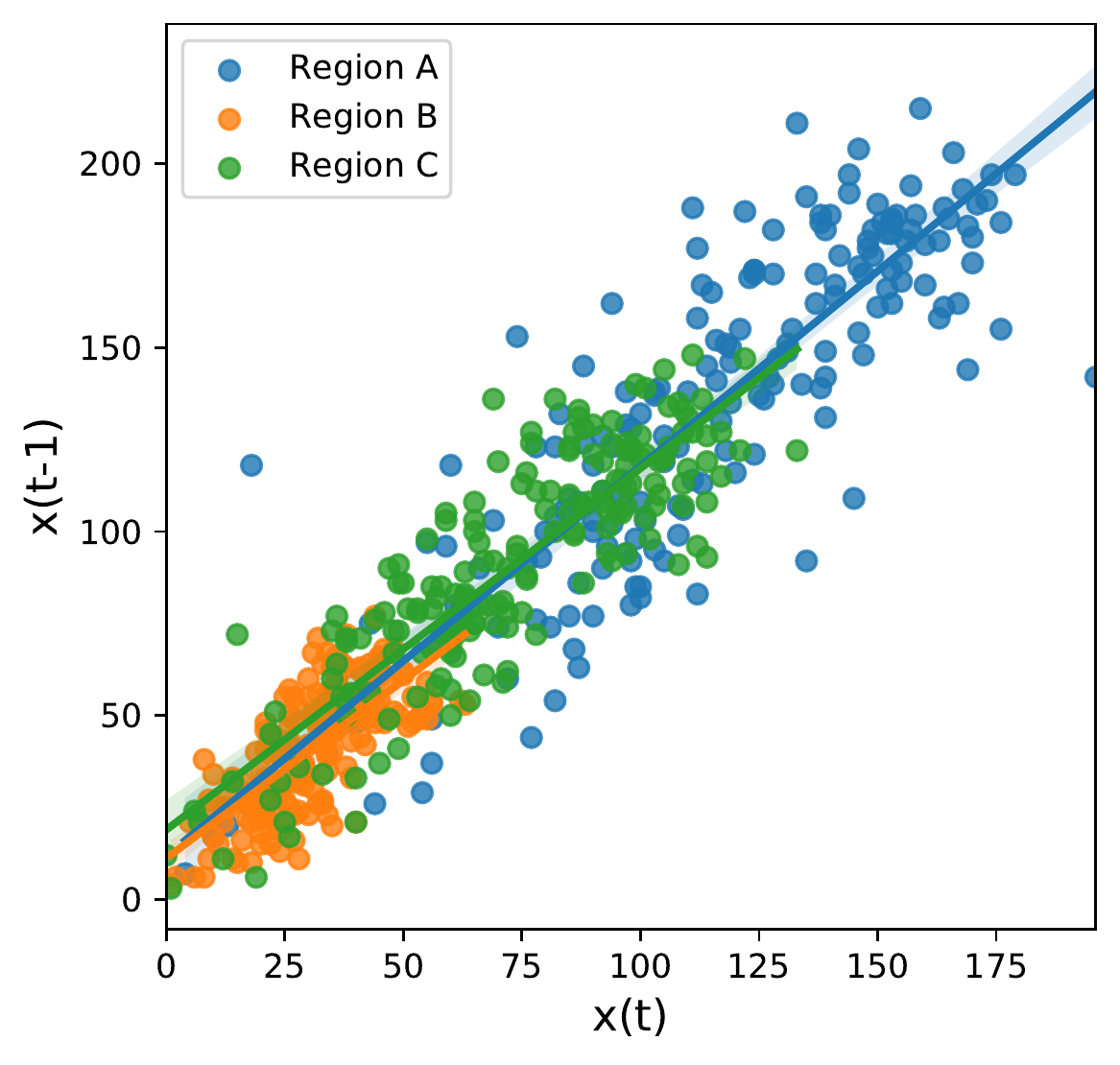}
\label{fig:x_xm1_region}
}
\subfloat[]{
\includegraphics[width=0.4\linewidth]{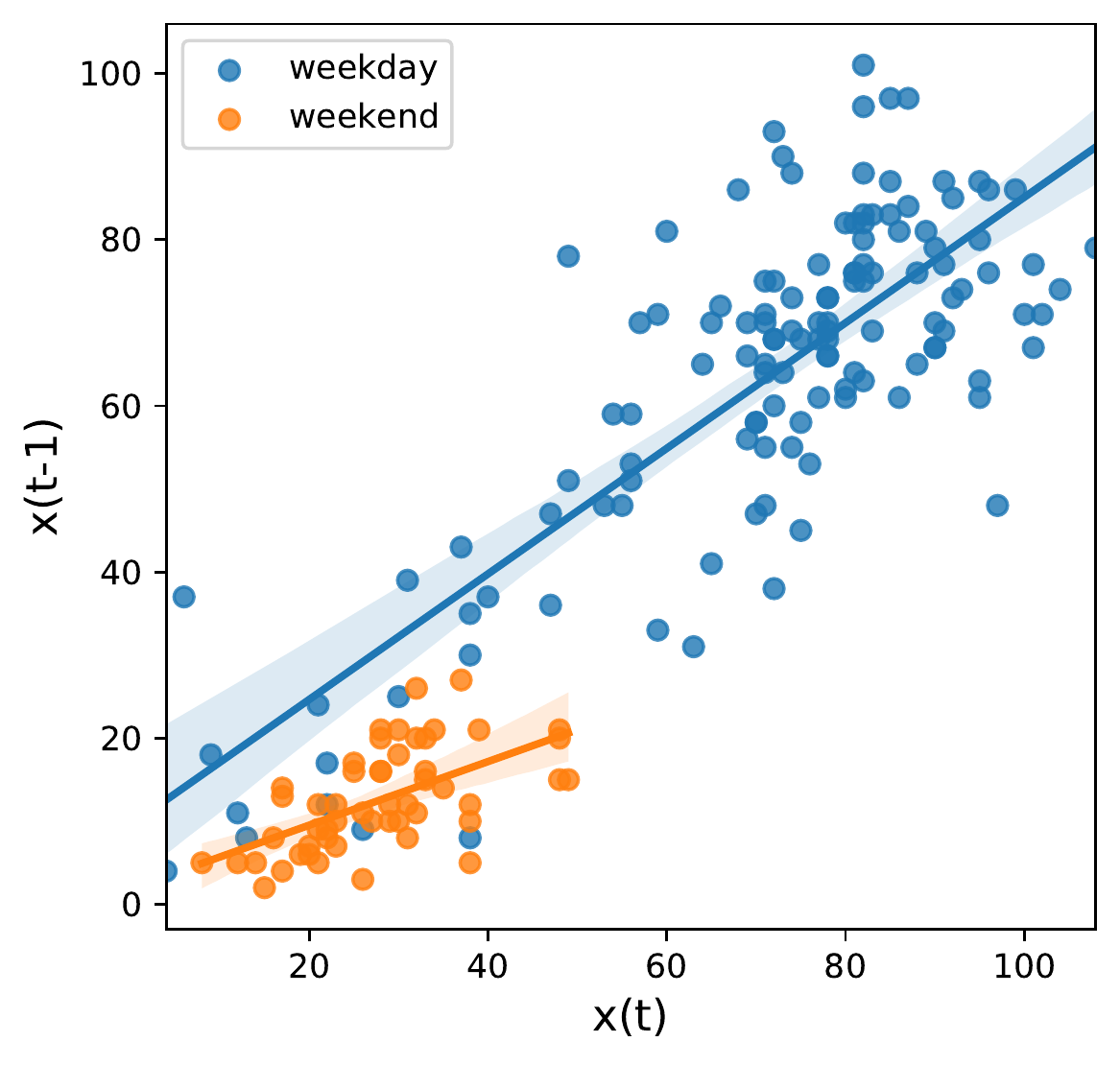}
\label{fig:x_xm1_time}
}
\caption{(a) Spatial indistinguishability; (b) Temporal indistinguishability.}
\label{fig:issue}
\end{figure}

However, only using raw time series data as input features does not obtain the inter-task weak-correlation and the intra-task strong-correlation from either of the spatial view or the temporal view. Although some tasks are inherently separated (i.e. 9am versus 6pm) based on the sequential pattern, most tasks are indistinguishable in the feature space. Following our previous work \cite{deng2021st},  there are two types of indistinguishability due to the strong inter-task correlation from the spatial view and the temporal view: (1) \textbf{Spatial indistinguishability} means that the dynamics yielded by different variables are not adequately discernible. For instance, looking at the three regions in Fig. \ref{fig:evolution}, we consider their dynamics measured between 8pm and 9pm on different days. In Fig. \ref{fig:x_xm1_region}, we plot the measurement at 8pm versus the measurement at 9pm over the three regions, where the data points are colored in accordance with their regional identities. Different clusters of dynamics are supposed to be distinguishable. However, the cluster-wise relationships (indicated by the direction of a straight line fitting the intra-cluster data points) are highly correlated, which signifies the inter-task strong correlation; (2) \textbf{Temporal indistinguishability} means that dynamics measured at specific times are not substantially discrete. In Fig. \ref{fig:x_xm1_time}, we only plot the measurement pairs of region A, and separate them based on weekday or weekend. It is obvious that these two clusters also have a strong correlation.

To address the issues, we propose two fundamental operations, \textbf{task-wise affine transformation} and \textbf{task-wise normalization}, each of which can weaken the inter-task correlation while maintaining the intra-task correlation. Task-wise affine transformation transforms the representations of each sample with task-specific affine parameters, hence task-specific characteristics can be encoded into the feature space. The limitation of this operation is that it can only be applied on the spatial view whose task partition is static over time, or in other words the set of tasks does not change with time. When it comes to the temporal view with dynamic task partition, the model cannot pre-learn the affine parameters for tasks appearing at a future time. To complement task-wise affine transformation, task-wise normalization is proposed, which can be applied not only on the spatial view but also on the temporal view. Basically, it performs normalization over the entire group of samples divided into the same task, which can also result in representations with task-specific characteristics.

We summarize our contributions as follows:
\begin{itemize}
    \item We propose a novel MVMT learning framework for time series forecasting. We account for three views in this framework, namely the original view, the spatial view and the temporal view. From each of the spatial view and the temporal view, learning is performed in a multi-task manner where each task is associated with a variable or a timestamp.
    \item We develop task-wise affine transformation and normalization to enable the feature space to be encoded with MVMT information. Either of these two operations can weaken the inter-task correlations while keeping the intra-task correlations in the representation space, which emulates the explicit partitioning of data samples as in the normal setting of multi-task learning.
    \item We conduct extensive experiments to quantitatively and qualitatively validate the effectiveness of the MVMT learning framework.
\end{itemize}


\section{Related Work}

\subsection{Time Series Forecasting}
Time series forecasting has been studied for decades. Traditional methods, such as ARIMA, can only learn the linear relationship among different timesteps, which has an inherent deficiency in fitting many real-world time series data that are highly nonlinear. With the power of deep learning models, a large volume of work in this area that has recently achieved impressive performance. For instance, \cite{qin2017dual} adopt LSTM to capture the nonlinear dynamics and long-term dependencies in time series data. However, the memorizing capacity of LSTM is still restricted, as pointed out by \cite{zhao2020rnn}. To  resolve this issue, \cite{chang2018memory, tang2020joint} create an external memory to explicitly store some representative patterns that can be frequently observed in the history, which is able to effectively guide the forecasting when similar patterns occur. \cite{lai2018modeling} makes use of a skip connection to enable the information to be transmitted from distant history. The attention mechanism is another option to deal with the vanishing memory problem \cite{tan2020data, fan2019multi}. Of these methods, Transformer is a representative architecture which consists of only attention operations \cite{vaswani2017attention}. To overcome the computation bottleneck of canonical Transformer, \cite{li2019enhancing} proposes a novel mechanism that periodically skips some timesteps when performing attention. As far as we know, Wavenet \cite{oord2016wavenet}, TCN \cite{bai2018empirical} and Transformer \cite{vaswani2017attention} are currently the superior choices for modeling long-term time series data \cite{wu2019graph, wu2020connecting, sen2019think}.

To tackle MTS, several studies \cite{sen2019think, yu2016temporal, yu2017long} assume that multi-variate time series data has a low-rank structure. Another thread of works \cite{liang2018geoman, qin2017dual} leverage the attention mechanism to learn the correlations among individual time series. Recently, \cite{wu2020connecting} inferred the inherent structure over the variables derived from self-learned encodings associated with each variable. The methods make point estimation. \cite{rangapuram2018deep, salinas2019deepar, wang2019deep} propose a confidence interval that is likely to contain the forthcoming observation. With prior knowledge on the application scenario, rich kinds of inter-variate relationship can be utilized. For instance, in real-world transportation system, there are three typical relationships, namely spatial closeness \cite{geng2019spatiotemporal, 9096591, 9139357, 9063494}, functional similarity \cite{geng2019spatiotemporal} and origin-destination connectiveness \cite{wang2019origin}. In particular, spatial closeness follows the first law of geography - "near things are more related than distant things"; functional similarity explains a part of phenomenons that although two locations are separate far away from each other, they exhibit similar movement pattern of traffic recordings over a day; origin-destination (OD) connectiveness is characterized as the quantity of the traffic flow between the OD pair \cite{wang2019origin}.


\subsection{Normalization}
Normalization was first adopted in deep image processing, and has significantly enhanced the performance of deep learning models for nearly all tasks. There are multiple normalization methods, such as batch normalization \cite{ioffe2015batch}, instance normalization \cite{ulyanov2016instance}, group normalization \cite{wu2018group}, layer normalization \cite{ba2016layer} and positional normalization \cite{li2019positional}, each of which is proposed to address a particular group of computer vision tasks. Of these, instance normalization has the greatest potential for our study, which was originally designed for image synthesis owing to its power to remove style information from the images. Researchers have found that feature statistics can capture the style of an image, and the remaining features upon normalizing the statistics are responsible for the content. Such a separable property enables the content of an image to be rendered in the style of another image, which is also known as style transfer. The style information in the image is like the scale information in time series. There is another line of work which explores the reason why the normalization trick facilitates the learning of deep neural networks \cite{lian2019revisit, bjorck2018understanding, santurkar2018does, daneshmand2020batch}. One of their major discoveries is that normalization can increase the rankness of the feature space, or in other words, it enables the model to extract more diverse features. 

\section{Preliminaries}
\label{sec:pre}

\newtheorem{assumption}{Assumption}
In this section, we introduce the definitions and the assumptions. All frequently used notations are reported in Table \ref{tab:notation}.

\begin{table*}[thb]
\centering
\caption{Notations} \label{tab:notation}
\begin{tabular}{l|l}
\hline
Notation    & Description   \\
\hline
$N, M$ & Number of variables / tasks. \\
$T, T_{in}, T_{out}$ & Length of history, number of input steps / output steps. \\
$\mathbf{X} \in \mathbb{R}^{N \times T}$  & Historical observations for forecasting. \\
$\mathbf{Z} \in \mathbb{R}^{N \times T \times d}$ & Historical latent representation. \\
$\mathbb{I} \subseteq [\![1,N]\!] \times [\![1,T]\!]$  & A set of sample indexes. \\
$\mathbb{P} = \{\mathbb{I}_1, \cdots,  \mathbb{I}_M\}$ & A task partition which is a partition of the entire set of sample indexes. \\
$\mathbf{G}^{\mathbb{P}} \in \mathbb{R}^{M \times d}$ & Global components conditioned on partition $\mathbb{P}$. \\
$\mathbf{L}^{\mathbb{P}} \in \mathbb{R}^{N \times T \times d}$ & Local components conditioned on partition $\mathbb{P}$. \\
$\mathbb{T}, \mathbb{S}$ & Task partition from the temporal / spatial view. \\
$ \mathbf{x}, \mathbf{y}, \mathbf{z}$ & Vector or matrix  that represents certain variable. \\
$+, \cdot, /$ & Element-wise addition / multiplication / division. \\
\hline
\end{tabular}
\end{table*}

\subsection{Task Partition Schemes}
\label{sec:tps}

\begin{figure}[tbh]
\centering
\subfloat[]{
\includegraphics[width=0.45\linewidth]{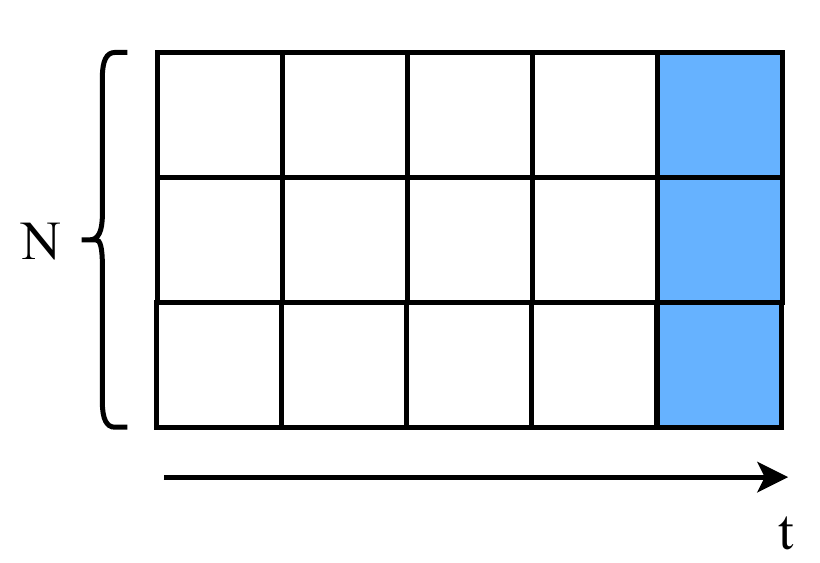}
\label{fig:t}
}
\subfloat[]{
\includegraphics[width=0.45\linewidth]{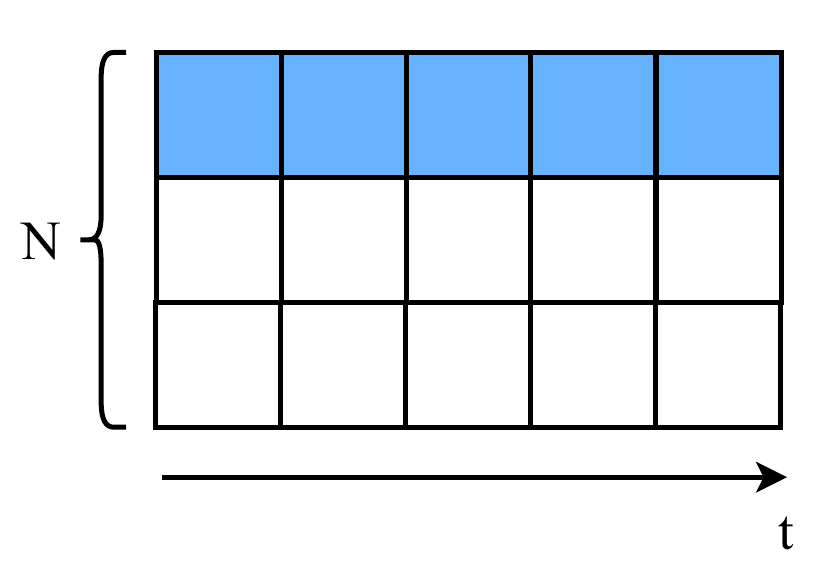}
\label{fig:v}
}
\caption{Illustration of task partition schemes from: (a) the temporal view; (b) the spatial view.}
\label{fig:nomrs}
\end{figure}

We employ two schemes to partition tasks from the spatial view and the temporal view, which are introduced as follows:

\begin{itemize}
    \item Temporal view: samples collected at the same time are put in the same task as shown in Figure \ref{fig:t}, ignoring the spatial difference. In this case, $M = T$ and $\mathbb{I}_m = \{(i, m)\}_{i=1}^{N}$.
    \item Spatial view: samples of the same variable are put in the same task as shown in Figure \ref{fig:v}, ignoring the temporal difference. In this case, $M = N$ and $\mathbb{I}_m = \{(m, j)\}_{j=1}^{T}$.
\end{itemize}

In terms of any of the aforementioned partition schemes, samples put in the same task are impacted by the same factor, thus they should manifest a strong correlation. Meanwhile, samples across tasks are impacted by different factors, which result in weakly correlated patterns. 

There are many partition schemes with different granularities. For example, we can also put the samples collected during a specified period in a task. In our work, we adopt the finest granularity, and the question of what the optimal granularity is is left to explore in future works.

\subsection{Other Preliminaries}

\begin{definition}[Time series forecasting] Time series forecasting is formulated as the following conditional distribution:
\[
P(\mathbf{Y} | \mathbf{X}) = \prod_{t=1}^{T_{out}} P(\mathbf{Y}_{:, t} | \mathbf{X}),
\]
\end{definition}

\begin{definition}[Time series factorization]
\label{def}
   Specifying a task partition scheme $\mathbb{P}$, any sample of the time series can be factorized in the following way:
\begin{align}
    \mathbf{Z}_{n, t} = \mathbf{G}^\mathbb{P}_m \mathbf{L}^\mathbb{P}_{n, t}, \label{eq:tsf}
\end{align}
where $m = \mathcal{P}(n, t)$ denotes the task belonging to the sample under $\mathbb{P}$, $\mathbf{G}^\mathbb{P}_m$ is a global component shared by all the samples from task $m$ under $\mathbb{P}$, and $\mathbf{L}^\mathbb{P}_{n, t}$ is a  local component only possessed by sample $(n, t)$. Be aware that different task partition schemes will result in different forms of factorization.
\end{definition}

\begin{assumption}
\label{assume}
We postulate that different dimensions of the local component are independent, and they follow a multi-variate normal distribution as follows:
\begin{align}
    \mathbf{L}^\mathbb{P}_{n, t} \sim \mathcal{N}(\begin{bmatrix}
    \beta^\mathbb{P}_m[0] \\
    \vdots \\
    \beta^\mathbb{P}_m[d-1]
  \end{bmatrix}, \begin{bmatrix}
    (\gamma^\mathbb{P}_m[0])^2 & & \\
    & \ddots & \\
    & & (\gamma^\mathbb{P}_m[d-1])^2
  \end{bmatrix}), \label{eq:2}
\end{align}
where $\beta^\mathbb{P}_m$ denotes the mean vector and $\beta^\mathbb{P}_m[i]$ denotes its $i^{\text{th}}$ entry; the co-variance matrix is a diagonal matrix, where the vector of entries on the main diagonal are denoted as $(\gamma^\mathbb{P}_m)^2$ and all the off-diagonal entries are $0$.
\end{assumption}

\newtheorem{remark}{Remark}

\begin{remark}
\label{remark:1}
Given a collection of samples from any task $m$, their local components are denoted as $\{\mathbf{L}^\mathbb{P}_{n, t} | \mathcal{P}(n, t) = m\}$. Intuitively, $\{\mathbf{L}^\mathbb{P}_{n, t} | \mathcal{P}(n, t) = m\}$ is expected to span a subspace whose dimension is lower than the effective dimension of the latent representation space $d$, as $\mathbf{L}^\mathbb{P}_{n, t}$ results from the lower number of environmental factors compared to $\mathbf{Z}_{n, t}$. Hence, only a part of the entries in $\gamma^\mathbb{P}_m$ have non-zero values. In addition, if $\{\mathbf{L}^\mathbb{P}_{n, t} | \mathcal{P}(n, t) = m_1\}$ and $\{\mathbf{L}^\mathbb{P}_{n, t} | \mathcal{P}(n, t) = m_2\}$ with different task identities are impacted by the same group of environmental factors,  they are supposed to span the same subspace.
\end{remark}

\begin{remark}
\label{remark:2}
Indistinguishability is attributed to the phenomenon where latent representations from different tasks span the same subspace. In the rest of this paragraph, we discuss the root cause of this phenomenon. To begin with, we can treat Eq. \ref{eq:tsf} from the view of geometric transformation. In this way, local components from the same task will experience the same transformation relying on the corresponding global component, and the ones from the different tasks will experience different transformations. In practice, there are multiple types of geometric transformations, which are made up of three basic transformations, namely scaling, translating and rotating. Scaling is the one that will not change the space spanned by the local components, so any group of transformations that only differ in scaling factors will cause the produced latent representations to be indistinguishable. A concrete instance of scaling in the real-world is the effect imposed by population: taxi demand over a region is normally proportional to the population residing in this region.
\end{remark}

\section{Methodology}
\label{sec:method}

\begin{figure}
    \centering
    \includegraphics[width=0.7\linewidth]{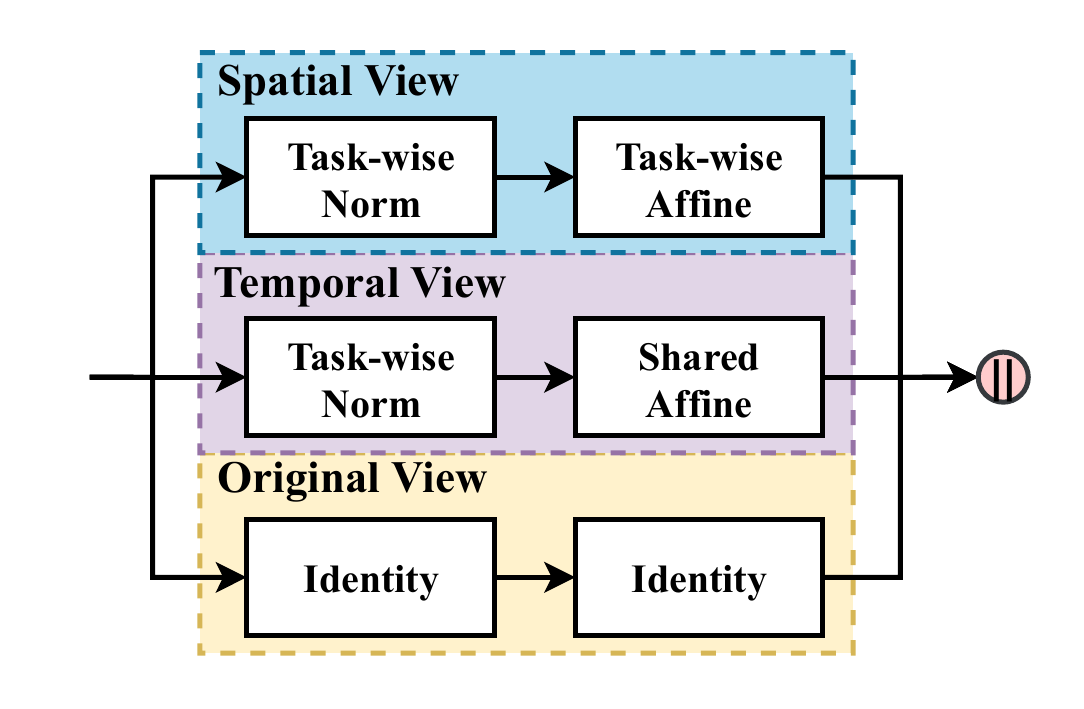}
    \caption{An overview of the MVMT block which can be incorporated into time series models, such as Wavenet.}
    \label{fig:mvmt-framework}
\end{figure}

An overview of the multi-view multi-task learning block is displayed in Figure \ref{fig:mvmt-framework}. Any representation input to this block is firstly replicated to three copies. Then, the three copies are separately transformed by specific operations respectively from the spatial view, the temporal view and the original view. Finally, the resulting three copies are concatenated together to obtain an augmented representation, which is taken to be the output of this block.

In this section, we start by introducing the proposed task-wise affine transformation and task-wise normalization respectively in Sec. \ref{sec:tat} and Sec. \ref{sec:tn}; then, in Sec. \ref{sec:wavenet}, we demonstrate how to integrate the proposed MVMT block into a widely adopted neural architecture -- Wavenet; we conclude this section by introducing the process of forecasting and learning in Sec. \ref{sec:fl}.

\subsection{Task-wise Affine Transformation}
\label{sec:tat}

Task-wise affine transformation differentiates tasks by assigning each task with an exclusive group of affine parameters. As each group of parameters is only responsible for capturing the dynamics of a single task, the indistinguishability issue can be mitigated. 

Formally, with respect to a task partition scheme $\mathbb{P}$ and its associated mapping function $\mathcal{P}$, task-wise affine transformation takes the following operations:
\begin{align*}
    \bar{\mathbf{Z}}^{\mathbb{P}}_{n, t} = \mathbf{Z}_{n, t} \mathbf{w}_m^{\mathbb{P}} + \mathbf{b}_m^{\mathbb{P}},
\end{align*}
where $m$ denotes $\mathcal{P}(n, t)$; $\mathbf{w}_m^{\mathbb{P}}$ and $\mathbf{b}_m^{\mathbb{P}}$ are two affine parameters targeting at task $m$ under the partition $\mathbb{P}$. To be noted that affine transformation is a special case of a fully connected layer, where interactions between the feature channels are taken into consideration. We will empirically show that task-wise affine transformation in conjunction with the following task-wise normalization achieves competitive performance, and introduces far fewer parameters in contrast with the task-wise fully connected layer.

Although task-wise affine transformation allows for more freedom to learn task-wise dynamics, it has severe limitation in handling cold-start tasks. In the practice of time series forecasting, tasks partitioned from the temporal view accumulate with time. For each new task encountered in the testing phase, we must identify a task that not only presents similar dynamics to the one being tested, but it has also been observed in the training data. For a time series showing regular patterns, such identification can be accomplished given prior knowledge on regularity. Nonetheless, for the ones with irregular patterns, it would be cumbersome to identify eligible tasks.

Due to the limitation, we apply task-wise affine transformation from the spatial view. The implementation takes the following form:
\begin{align}
    \bar{\mathbf{Z}}_{n, t}^{\mathbb{S}} = \mathbf{Z}_{n, t} \mathbf{w}^{\mathbb{S}}_{n} + \mathbf{b}^{\mathbb{S}}_{n}
\end{align}

Next, we introduce another thread of approaches which can address the cold-start problem.

\subsection{Task-wise Normalization}
\label{sec:tn}

Task-wise normalization explicitly encodes global components into the representation space. The global component varies from task to task, and thus can be used to separate tasks. However, the observation of any sample is a mixture of the global component and the local component, which hinders the capture of the global component. To extract the global component, we start by applying task-wise normalization to eliminate the global component from the representation, then we combine the normalized representation with the original representation to obtain an augmented representation for each sample. The augmented representation space can manifest the difference on global components, and hence the current inter-task correlation is weaker than the original.

Likewise, for a task partition scheme $\mathbb{P}$ and its associated mapping function $\mathcal{P}$, task-wise normalization is performed as follows:
\begin{align}
    \hat{\mathbf{Z}}^{\mathbb{P}}_{n, t} = \frac{\mathbf{Z}_{n, t}  -  \mu^{\mathbb{P}}_m}{\sigma^{\mathbb{P}}_m} \label{eq:4}
\end{align}
where we let $m$ denote $\mathcal{P}(n, t)$,
\begin{align}
    \mu^{\mathbb{P}}_m &= E\left[ \mathbf{Z}_{i, j} | \mathcal{P}(i, j) = m\right] \label{eq:6}\\
    &\approx \frac{1}{|\mathbb{I}_m|}\sum_{i, j}^{\mathbb{I}_m} \mathbf{Z}_{i , j} \nonumber \\
    \sigma^{\mathbb{P}}_m &= \sqrt{E\left[ \left(\mathbf{Z}_{i, j} - \mu^{\mathbb{P}}_m\right)^2 | \mathcal{P}(i, j) = m \right]} \label{eq:8}\\
    &\approx \sqrt{ \frac{1}{|\mathbb{I}_m|}\sum_{i, j}^{\mathbb{I}_m} (\mathbf{Z}_{i, j} - \mu^{\mathbb{P}}_m)^2 + \epsilon}. \nonumber
\end{align}
By implementing $\mathbb{P}$ with different task partition schemes, we gain representations normalized in different ways. For instance, the task partition from the temporal view produces the following implementation of task-wise normalization:
\begin{align*}
    \hat{\mathbf{Z}}^{\mathbb{T}}_{n, t} = \frac{\mathbf{Z}_{n, t}  -  \mu^{\mathbb{T}}_t}{\sigma^{\mathbb{T}}_t},
\end{align*}
where
\begin{align*}
    \mu^{\mathbb{T}}_t &=  \frac{1}{N}\sum_{i=1}^{N} \mathbf{Z}_{i , t},\\ 
    \sigma^{\mathbb{T}}_t &= \sqrt{ \frac{1}{N}\sum_{i=1}^{N} (\mathbf{Z}_{i, t} - \mu^{\mathbb{T}}_t)^2 + \epsilon}.
\end{align*}
For conciseness, we do not include the implementation from the spatial view.

Next, we explain why task-wise normalization works. As we indicate in Remark \ref{remark:2}, certain cases of indistinguishability are caused by task-wise transformations that differ only in scaling factors. Task-wise normalization resolves this issue by converting scaling transformation to rotating transformation. Basically, we construct rotation with a rotating angle which relies on the scaling factor. Firstly, we rewrite the normalized representation from Eq. \ref{eq:4} in the following way:


\begin{align}
    \hat{\mathbf{Z}}^{\mathbb{P}}_{n, t} &= \frac{\mathbf{Z}_{n, t} - \mu^{\mathbb{P}}_m }{\sigma^{\mathbb{T}}_m} \label{eq:10} \\
    &= \frac{\mathbf{L}^{\mathbb{P}}_{n, t} - \beta_m^\mathbb{P}}{ \gamma_m^\mathbb{P}} \label{eq:11},
\end{align}
where Eq. \ref{eq:11} is deduced by substituting Eq. \ref{eq:tsf}, Eq. \ref{eq:2}, Eq. \ref{eq:6} and Eq. \ref{eq:8} into Eq. \ref{eq:10}. Then combining the original view, the spatial view and the temporal view, we map the original representation to an augmented representation as follows:
\[
\mathbf{Z}_{n, t} \rightarrow 
\begin{bmatrix}
    \mathbf{Z}_{n, t} \\
    \hat{\mathbf{Z}}^{\mathbb{S}}_{n, t} \\
    \hat{\mathbf{Z}}^{\mathbb{T}}_{n, t} \\
\end{bmatrix} = 
\begin{bmatrix} 
    \mathbf{Z}_{n, t} \\
	\frac{\mathbf{L}^{\mathbb{S}}_{n, t} - \beta_{m_1}^\mathbb{S}}{ \gamma_{m_1}^\mathbb{S}} \\
	\frac{\mathbf{L}^{\mathbb{T}}_{n, t} - \beta_{m_2}^\mathbb{T}}{ \gamma_{m_2}^\mathbb{T}}\\
	\end{bmatrix},
\]
where $m_1 = \mathcal{S}(n, t)$ and $m_2 = \mathcal{T}(n, t)$. At this step, from the view of geometric transformation, each sample point is rotated by an angle depending on $\mathbf{G}^{\mathbb{P}}_m \gamma_m^\mathbb{P}$  from its original position and is then translated by $\beta_m^\mathbb{P} / \gamma_m^\mathbb{P}$. Thus, the difference on $\mathbf{G}^{\mathbb{P}}_m$, on $\beta_m^\mathbb{P}$ and on $\gamma_m^\mathbb{P}$ can be manifested in the new space. It is noteworthy that the new space also maintains the correlation among $\mathbf{G}^{\mathbb{P}}_m$ belonging to different tasks, which can facilitate the learning process.

\subsection{Wavenet}
\label{sec:wavenet}

\begin{figure}[thb]
    \centering
    \includegraphics[width=\linewidth]{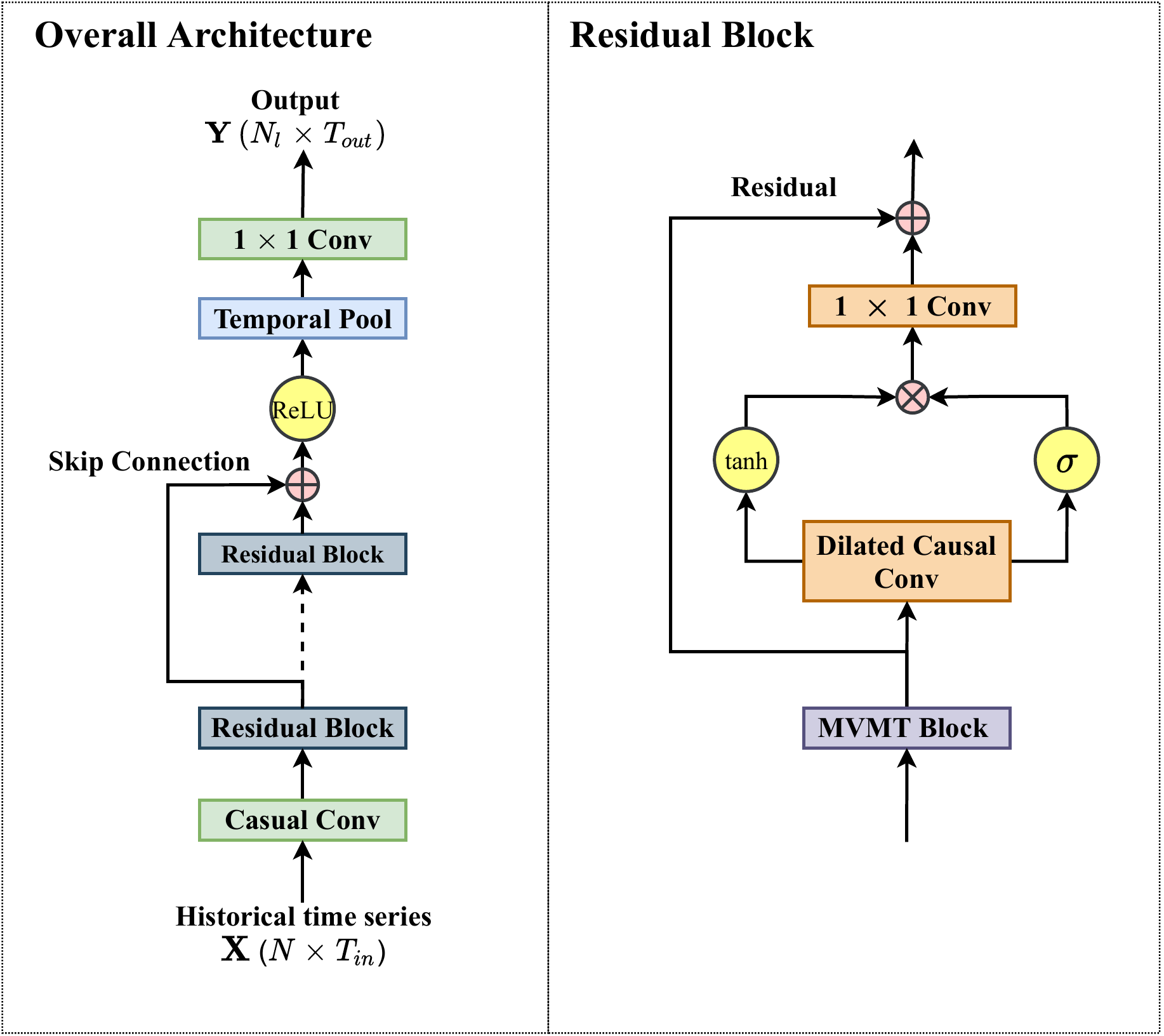}
    \caption{Overall architecture, showing two residual blocks for illustration, but multiple blocks can be stacked layer by layer. $+$ and $\times$ respectively denote element-wise addition and element-wise multiplication.}
    \label{fig:architecture}
\end{figure}

We illustrate the architecture of our work in Fig. \ref{fig:architecture}. Some key variables with their shapes are labeled at their corresponding positions along the computation path. Generally, our framework follows a structure like Wavenet\cite{bai2018empirical}, except that we incorporate an MVMT block into the residual block.

\begin{figure}[thb]
    \centering
    \includegraphics[width=0.6\linewidth]{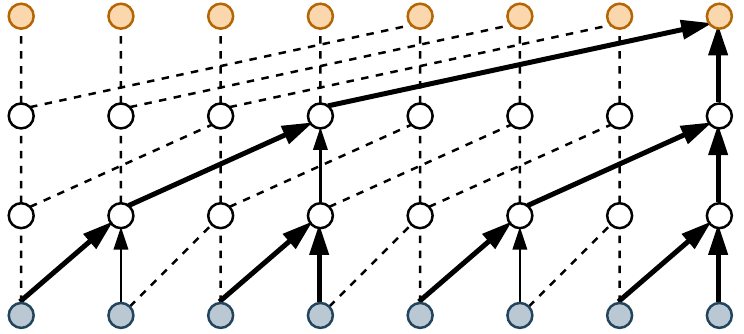}
    \caption{Dilated Causal Convolution.}
    \label{fig:my_label}
\end{figure}

We briefly introduce a dilated causal convolution where the filter is applied with skipping values. For a 1-D signal $\mathbf{z} \in \mathbb{R}^T$ and a filter $f: \{0, \dots, k - 1 \} \to \mathbb{R}$, the causal convolution on element $t$ is defined as follows:
\begin{equation}
    F(t) = (\mathbf{z} \ast f)(t) = \sum_{i=0}^{k-1}f(i) \cdot \mathbf{z}_{t-i}.
\end{equation}
This formula can be easily generalized for a multi-dimension signal but we omit its general form here for brevity. Moreover, padding (zero or replicate) with size of $k - 1$ is appended to the left tail of the signal to ensure length consistency. We can stack multiple causal convolution layers to obtain a larger receptive field for each element.

One shortcoming of using causal convolution is that either the kernel size or the number of layers increases in a linear manner with the range of the receptive field, and the linear relationship causes an explosion of parameters when modeling long history. Pooling is a natural choice to address this issue, but it sacrifices the order information presented in the signal. To this end, dilated causal convolution is used, a form which supports the exponential expansion of the receptive field. The formal computing process is written as:
\begin{equation}
    F(t) = (\mathbf{z} \ast_d f)(t) = \sum_{i=0}^{k-1}f(i) \cdot \mathbf{z}_{t-d \cdot i},
\end{equation}
where $d$ is the dilation factor. Normally, $d$ increases exponentially w.r.t. the depth of the network (i.e., $2^l$ at level $l$ of the network). If $d$ is $1$ ($2^0$), then the dilated convolution operator $\ast_d$ reduces to a regular convolution operator $\ast$.

\subsection{Forecasting and Learning}
\label{sec:fl}
We let $\mathbf{Z}^{(L)} \in \mathbb{R}^{N_l \times T_{in} \times d_z}$ denote the output from the last residual block, where each row $\mathbf{z}^{(L)} \in \mathbb{R}^{T_{in} \times d_z}$ represents a variable. Then, we employ a temporal pooling block to perform temporal aggregation for each variable. Several types of pooling operations can be applied, such as max pooling and mean pooling, depending on the problem being studied. In our case, we select the vector in the most recent time slot as the pooling result, which is treated as the representation of the entire signal. Finally, we make a separate prediction for each variable, based on the obtained representation using a shared fully connected layer.

In the learning phase, our objective is to minimize the mean squared error between the predicted values and ground truth values. We use the Adam optimizer \cite{kingma2014adam} to optimize this target.

\section{Evaluation}

In this section, we describe the extensive experiments on three common datasets to validate the effectiveness of MVMT from different aspects.

\subsection{Experimental Setting}
\subsubsection{Datasets}

\begin{table}[thb]
\small
\caption{Dataset statistics.}
\label{tab:dataset}
\begin{tabular}{l|ccc}
\hline
\textbf{Tasks}            & \textbf{Electricity}                                               & \textbf{PeMSD7}                                                    & \textbf{BikeNYC}                                                   \\ \hline
\textbf{Start time}       & 10/1/2014                                                          & 5/1/2012                                                           & 4/1/2014                                                           \\
\textbf{End time}         & 12/31/2014                                                         & 6/30/2012                                                          & 9/30/2014                                                          \\
\textbf{Sample rate}      & 1 hour                                                             & 30 minutes                                                         & 1 hour                                                             \\
\textbf{\# Timesteps}     & 2184                                                               & 2112                                                               & 4392                                                               \\
\textbf{\# Variate}       & 336                                                                & 228                                                                & 128                                                                \\ \hline
\textbf{Training size}    & 1848                                                               & 1632                                                               & 3912                                                               \\
\textbf{Validation size}  & 168                                                                & 240                                                                & 240                                                                \\
\textbf{Testing size}     & 168                                                                & 240                                                                & 240                                                                \\
\textbf{Output length}    & 3                                                                  & 3                                                                  & 3                                                                  \\ \hline
\end{tabular}
\end{table}

We validate our model on three real-world datasets, namely BikeNYC, PeMSD7 and Electricity. The statistics regarding each dataset as well as the corresponding settings of the designed task are reported in Table \ref{tab:dataset}. We standardize the values in each dataset to facilitate training and transform them back to the original scale in the testing phase.

Table \ref{tab:dataset} reports the statistics of the datasets. More details regarding the datasets are as follows.
\begin{itemize}
    \item PeMSD7 \cite{yu2018spatio}. The data is collected from the Caltrans Performance Measurement System (PeMS) using sensor stations, which are deployed to monitor traffic speed across the major metropolitan areas of the California state highway system. We further aggregate the data to $30$-minute intervals by average pooling.
    \item Electricity\footnote{https://archive.ics.uci.edu/ml/datasets/ElectricityLoadDiagrams20112014}. The original dataset contains the electricity consumption of $370$ points/clients, from which $34$ outlier points that contain extreme values are removed. Moreover, we calculate the hourly average consumption for each point, and take it as the time series being modeled.
    \item BikeNYC \cite{zhang2017deep}. Each time series in this dataset denotes the aggregate demand for shared bikes over a region in New York City. We do not consider the spatial relationship presented in the PeMSD7 and BikeNYC data, since our objective is to study the temporal patterns. 
\end{itemize}

\begin{figure}[thb]
\centering
\subfloat[PeMSD7.]{
\includegraphics[width=0.9\linewidth]{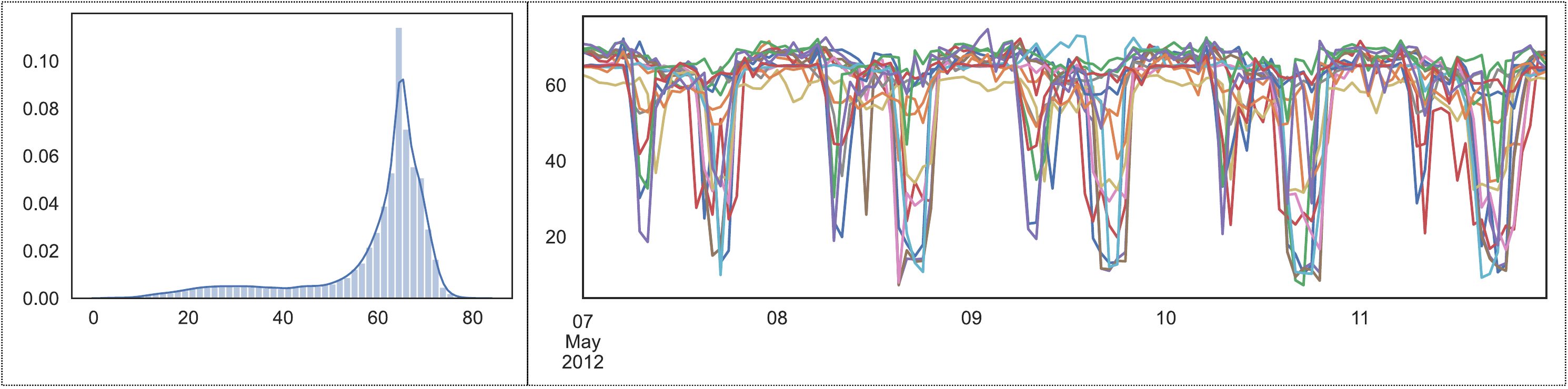}
}

\subfloat[Electricity.]{
\includegraphics[width=0.9\linewidth]{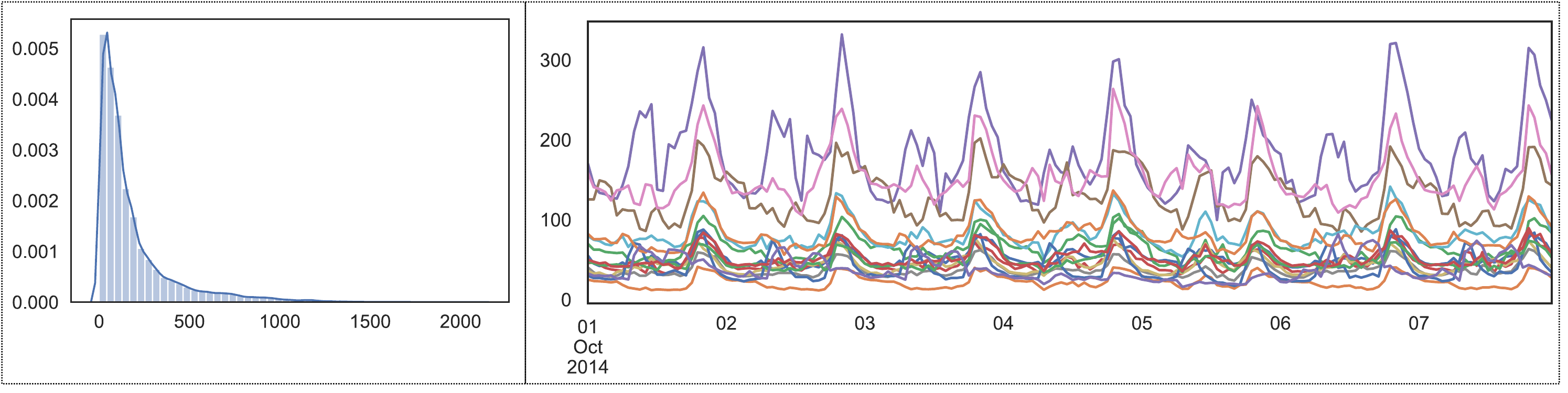}
\label{fig:elec_sample}
}

\subfloat[BikeNYC.]{
\includegraphics[width=0.9\linewidth]{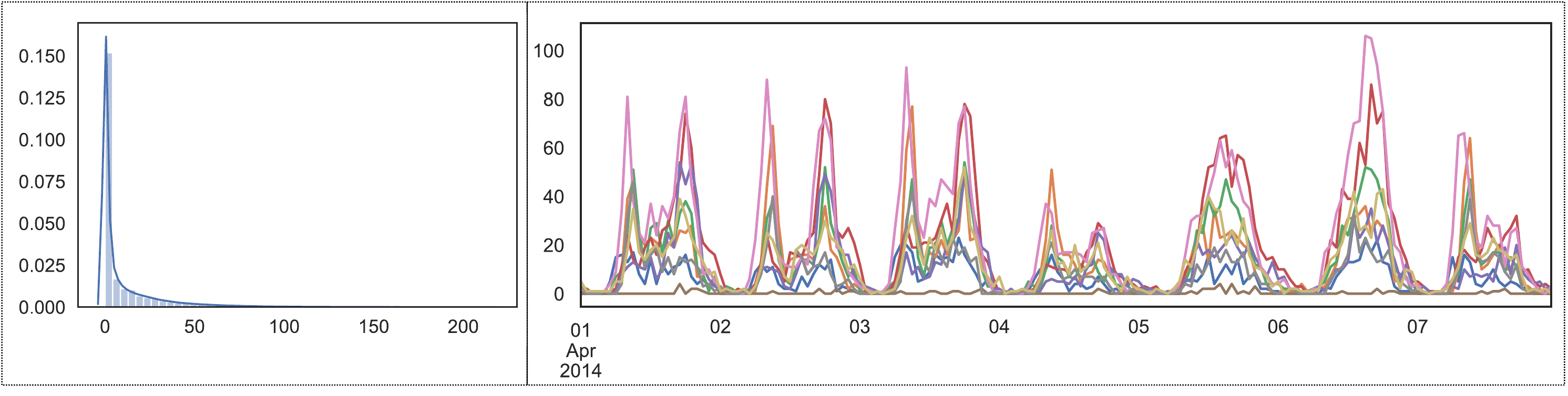}
}
\caption{For each of (a), (b) and (c), the left figure shows the probability density function of the observed values aggregated from all variables at all time steps, and the right figure displays some sample time series.}
\label{fig:data_dist}
\end{figure}

Furthermore, we display the data distribution and several exemplar time series in Fig. \ref{fig:data_dist} to gain more insights from each dataset. We can observe that each of the three types of data lays in a wide range of scale, and exhibits periodicity to some extent. However, their evolving patterns are entirely different. The electricity time series displays the greatest diversity. 

\subsubsection{Network Setting}
The batch size is $8$, and the input length of the batch sample is $16$. For the Wavenet backbone, the layer number is set to $4$, the kernel size of each DCC component is $2$, and the associated dilation rate is $2^i$, where $i$ is the index of the layer (counting from $0$). Such settings collectively enable the output from Wavenet to perceive $16$ input steps. The number of hidden channels $d_z$ in each DCC is $16$. We apply zero-padding on the left tail of the input to enable the length of the output from DCC to be $16$ as well.  The learning rate of the Adam optimizer is $0.0001$.

\subsubsection{Evaluation Metrics} We validate our model using root mean squared error (RMSE), mean absolute error (MAE) and mean absolute percentage error (MAPE). We repeat the experiment ten times for each model on each dataset and report the mean of the results. 

\subsection{Baseline Models}

\begin{table*}[t]
\centering
\caption{Performance on the BikeNYC dataset}
\label{tab:bike}
\begin{tabular}{l|ccc|ccc|ccc}
\hline
\multirow{2}{*}{Models} & \multicolumn{3}{c|}{1 hour}                                         & \multicolumn{3}{c|}{2 hour}                                         & \multicolumn{3}{c}{3 hour}                                         \\ \cline{2-10} 
                        & MAPE                 & MAE                  & RMSE                  & MAPE                 & MAE                  & RMSE                  & MAPE                 & MAE                  & RMSE                 \\ \hline
LSTNet                  & 19.6\%               & 2.55                 & 5.35                  & 21.1\%               & 2.77                 & 6.04                  & 22.6\%               & 2.99                 & 6.63                 \\
AGCRN                   & \underline{17.3\%}               & \underline{2.34}                 & \underline{4.76}                  & \underline{18.7\%}               & \underline{2.56}                 & \underline{5.51}                  & \underline{20.3\%}               & \underline{2.77}                 & 6.07                 \\
Graph Wavenet           & 18.0\%               & 2.39                 &  4.78                  & 19.4\%               & 2.65                 & 5.53                  & 20.8\%               & 2.86                 & \underline{6.05}                 \\
MTGNN                   & 19.0\%               & 2.55                 & 5.05                  & 21.1\%               & 2.88                 & 6.00                  & 22.9\%               & 3.13                 & 6.61                 \\ \hline
Transformer             & 22.8\%               & 2.98                 & 6.16                  & 27.1\%               & 3.66                 & 7.88                  & 29.7\%               & 4.12                 & 8.95                 \\
Transformer + MVMT       & 17.7\%               & 2.36                 & 4.75                  & 19.1\%               & 2.57                 & 5.51                  & 20.7\%               & 2.78                 & 6.09                 \\ \hline
TCN                     & 22.4\%               & 2.90                 & 5.98                  & 26.4\%               & 3.57                 & 7.66                  & 29.1\%               & 4.07                 & 8.76                 \\
TCN + MVMT               & \textbf{16.8\%}      & 2.30                 & 4.51                  & 18.6\%               & 2.55                 & 5.34                  & 20.4\%               & 2.77                 & 5.92                 \\ \hline
Wavenet                 & 22.1\%               & 2.86                 & 5.92                  & 26.3\%               & 3.52                 & 7.58                  & 29.0\%               & 3.97                 & 8.68                 \\
Wavenet + MVMT           &  16.9\%      & \textbf{2.23}        & \textbf{4.48}         & \textbf{18.3\%}      & \textbf{2.47}        & \textbf{5.28}         & \textbf{20.1\%}      & \textbf{2.68}        & \textbf{5.88}        \\  \hline
Improvements                                      & +2.8\%               & +4.7\%                 & +5.8\%                  & +2.1\%               & +3.5\%                 & +4.1\%                  & +0.9\%               & +3.2\%                 & +2.8\%                 \\\hline
\end{tabular}
\end{table*}

\begin{table*}[thb]
\centering
\caption{Performance on the PeMSD7 dataset.}
\label{tab:pems}
\begin{tabular}{l|ccc|ccc|ccc}
\hline
\multicolumn{1}{c|}{\multirow{2}{*}{Models}} & \multicolumn{3}{c|}{30 min}                                         & \multicolumn{3}{c|}{60 min}                                         & \multicolumn{3}{c}{90 min}                                         \\ \cline{2-10} 
\multicolumn{1}{c|}{}                        & MAPE                 & MAE                  & RMSE                  & MAPE                 & MAE                  & RMSE                  & MAPE                 & MAE                  & RMSE                 \\ \hline
LSTNet                                       & 8.10\%               & 3.88                 & 6.52                  & 8.43\%               & 4.01                 & 6.71                  & 9.06\%               & 4.30                 & 7.12                 \\
AGCRN                                        & \underline{4.87\%}              & 2.34                 & \underline{4.24}                  & \underline{6.48\%}               & \underline{3.07}                 & \underline{5.58}                  & \underline{7.27\%}               & \underline{3.43}                 & \underline{6.19}                 \\
Graph Wavenet                                & 4.90\%               & \underline{2.33}                 & 4.25                  & 6.77\%               & 3.19                 & 5.69                  & 7.57\%               & 3.57                 & 6.25                 \\
MTGNN                                        & 5.18\%               & 2.46                 & 4.48                  & 7.61\%               & 3.57                 & 6.31                  & 9.03\%               & 4.25                 & 7.26                 \\ \hline
Transformer                                  & 5.82\%               & 2.75                 & 5.03                  & 9.31\%               & 4.34                 & 7.51                  & 11.8\%               & 5.49                 & 9.02                 \\
Transformer + MVMT                            & 4.86\%               & 2.33                 & 4.26                  & 6.50\%               & 3.08                 & 5.65                  & 7.33\%               & 3.48                 & 6.31                 \\ \hline
TCN                                          & 5.80\%               & 2.75                 & 4.97                  & 9.44\%               & 4.43                 & 7.53                  & 12.0\%               & 5.61                 & 9.06                 \\
TCN + MVMT                                    & 4.91\%     & 2.34                 & 4.22                  & 6.42\%               & 3.04                 & 5.51                  & 7.12\%               & 3.38                 & 6.05                 \\ \hline
Wavenet                                      & 5.50\%               & 2.61                 & 4.80                  & 8.75\%               & 4.10                 & 7.20                  & 11.0\%               & 5.16                 & 8.61                 \\
Wavenet + MVMT                                & \textbf{4.71\%}      & \textbf{2.25}        & \textbf{4.12}         & \textbf{6.23\%}      & \textbf{2.95}        & \textbf{5.48}         & \textbf{6.97\%}      & \textbf{3.29}        & \textbf{6.01}        \\ \hline
Improvements                                      & +3.2\%               & +3.4\%                 & +3.0\%                  & +3.8\%               & +3.9\%                 & +1.7\%                  & +4.1\%               & +4.0\%                 & +2.9\%                 \\\hline
\end{tabular}
\end{table*}

\begin{table*}[thb]
\centering
\caption{Performance on the Electricity dataset.}
\label{tab:elec}
\begin{tabular}{l|ccc|ccc|ccc}
\hline
\multicolumn{1}{c|}{\multirow{2}{*}{Models}} & \multicolumn{3}{c|}{1 hour}                     & \multicolumn{3}{c|}{2 hour}                     & \multicolumn{3}{c}{3 hour}                      \\ \cline{2-10} 
\multicolumn{1}{c|}{}                        & MAPE            & MAE           & RMSE          & MAPE            & MAE           & RMSE          & MAPE            & MAE           & RMSE          \\ \hline
LSTNet                                       & 22.4\%          & 31.1          & 61.2          & 23.0\%          & 31.8          & 62.6          & 24.8\%          & 33.8          & 66.8          \\
AGCRN                                        & 12.2\%          & 17.1          & 36.3          & 16.1\%          & 22.3          & 49.1          & \underline{19.1\%}          & \underline{26.0}          & 56.6          \\
Graph Wavenet                                & \underline{10.9\%} & 15.9          & 34.9          & \underline{15.8\%}          & 22.4          & 49.9          & \underline{19.1\%}          & 26.5          & 57.7          \\
MTGNN                                        & 11.1\%          & \underline{15.8}          & \underline{32.5}          & 16.0\%          & \underline{22.2}          & \underline{46.3}          & 19.6\%          & 26.5          & \underline{54.6}          \\ \hline
Transformer                                  & 11.2\%          & 16.6          & 36.3          & 17.6\%          & 25.4          & 53.7          & 22.2\%          & 31.8          & 65.0          \\
Transformer + MVMT                            & 13.2\%          & 17.4          & 35.9          & 17.7\%          & 23.5          & 48.8          & 21.2\%          & 27.9          & 58.0          \\ \hline
TCN                                          & 11.1\%          & 16.3          & 35.5          & 17.3\%          & 25.0          & 52.4          & 21.5\%          & 30.7          & 62.0          \\
TCN + MVMT                                    & 13.2\%          & 16.7          & 31.7          & 16.5\%          & 21.5          & 42.8          & 20.1\%          & 25.4          & 50.9          \\ \hline
Wavenet                                      & \textbf{10.8\%} & 15.8          & 33.3          & 16.8\%          & 23.8          & 49.5          & 21.1\%          & 29.5          & 60.3          \\
Wavenet + MVMT                                & 12.0\%          & \textbf{15.6} & \textbf{30.9} & \textbf{15.1\%} & \textbf{20.1} & \textbf{42.2} & \textbf{17.1\%} & \textbf{23.0} & \textbf{49.2} \\ \hline
Improvements                                      & -11.0\%               & +1.2\%                 & +4.9\%                  & +4.4\%               & +9.4\%                 & +8.8\%                  & +10.4\%               & +11.5\%                 & +9.8\%                 \\\hline
\end{tabular}
\end{table*}

\begin{itemize}
    \item \textbf{MTGNN} \cite{wu2020connecting}. MTGNN constructs inter-variate relationships by introducing a graph-learning module. Specifically, the graph learning module connects each hub node with its top k nearest neighbors in a defined metric space. MTGNN's backbone architecture for temporal modeling is Wavenet.
    \item \textbf{Graph Wavenet} \cite{wu2019graph}. The architecture of Graph Wavenet is similar to MTGNN. The major difference is that the former derives a soft graph where each pair of nodes has a continuous probability of being connected.
    \item \textbf{AGCRN} \cite{bai2020adaptive}. AGCRN is also equipped with a graph-learning module to establish inter-variate relationship. Furthermore, it uses a personalized RNN to model the temporal relationship for each time series. 
    \item \textbf{Transformer} \cite{li2019enhancing}. This model captures the long-term dependencies in time series data by using an attention mechanism, where the keys and queries are yielded by causal convolution over local context to model segment-level correlation. 
    \item \textbf{LSTNet} \cite{lai2018modeling}. There are two components in LSTNet: one is a conventional autoregressive model, and the other is an LSTM with an additional skip connection over the temporal dimension.
    \item \textbf{TCN}. \cite{bai2018empirical} The architecture of TCN is similar to Wavenet, except that the nonlinear transformation in each residual block is made up of two rectified linear units (ReLU).
\end{itemize}

We also test the performance of TCN and Transformer incorporating MVMT, where MVMT is similarly applied before the causal convolution operation in each layer. We do not compare our method to linear models such as ARIMA, because the involved baseline models show superiority over the linear models as illustrated in the original work in \cite{lai2018modeling, li2019enhancing, wu2019graph, wu2020connecting}.

\subsection{Experimental Results}

\begin{figure}[tb]
\centering
\subfloat[BikeNYC]{
\includegraphics[width=0.45\linewidth]{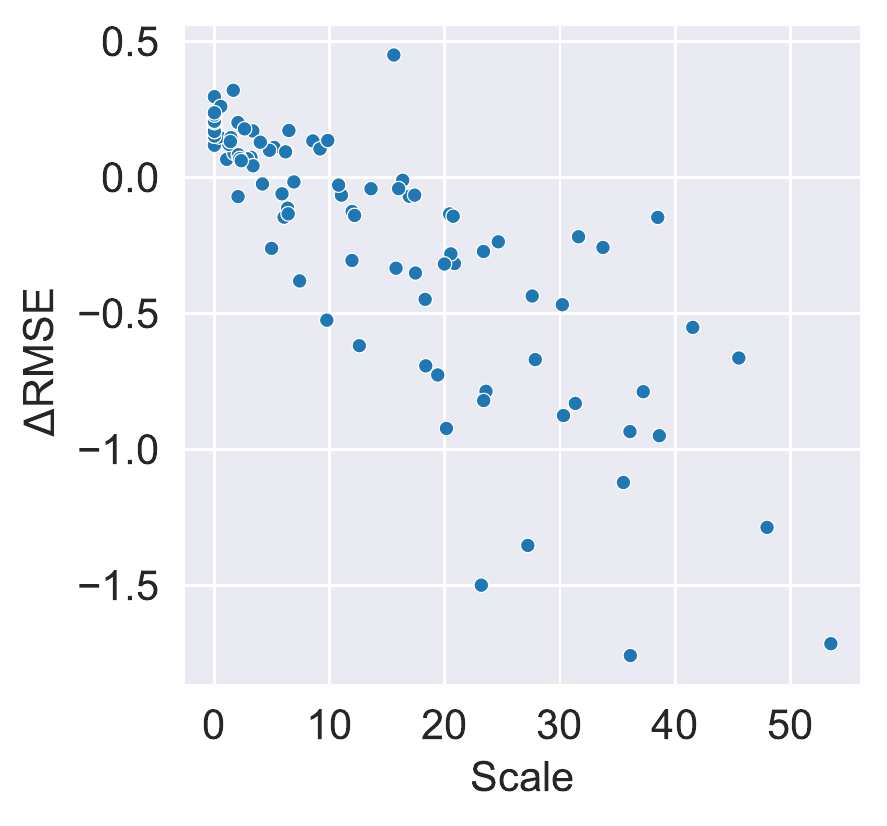}
}
\subfloat[PeMSD7]{
\includegraphics[width=0.45\linewidth]{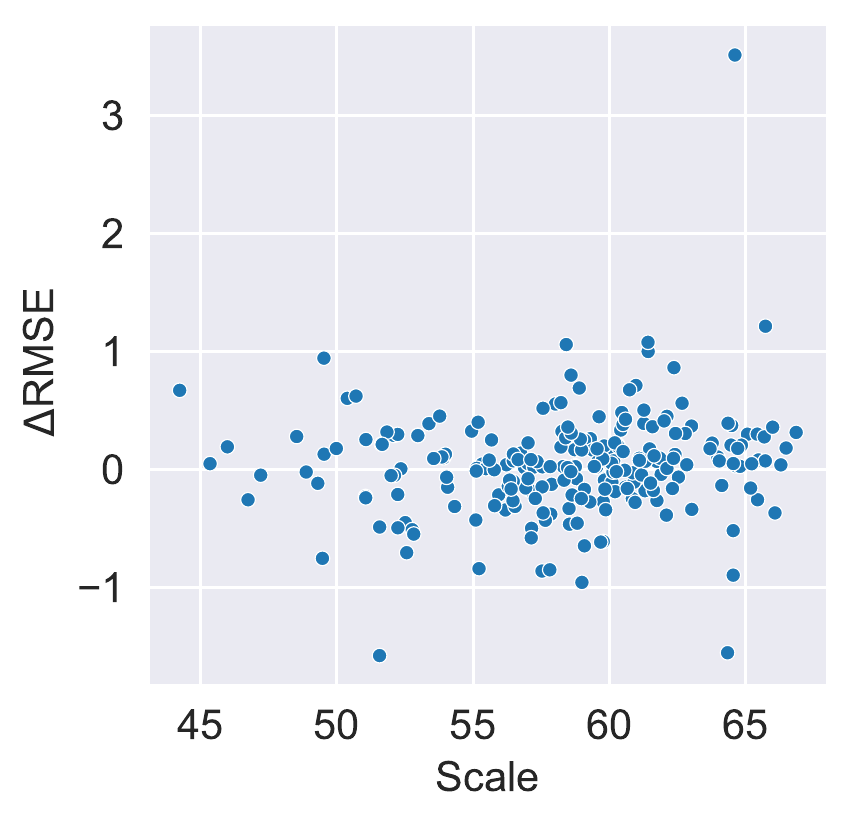}}

\subfloat[Electricity]{
\includegraphics[width=0.45\linewidth]{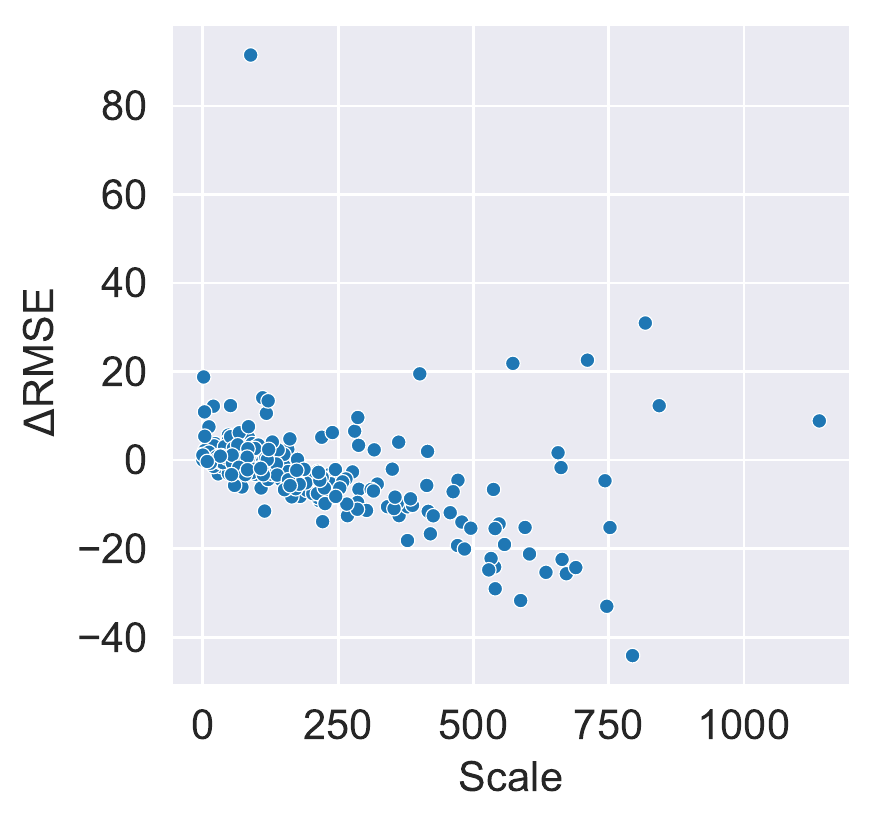}}
\caption{Variable-wise improvements over AGCRN.}
\label{fig:scale_error}
\end{figure}

The experimental results on the BikeNYC, PeMSD7 and Electricity datasets are reported in Table \ref{tab:bike}, Table \ref{tab:pems} and Table \ref{tab:elec} respectively. The improvements achieved by Wavenet + MVMT over the best benchmarks are recorded in the last row of each table.

It is obvious that Wavenet + MVMT achieves SOTA results over almost all horizons on the BikeNYC, PeMSD7 and electricity data. The reason for this is that we refine the high-frequency components from both the temporal view and the spatial view, which are generally overlooked by baseline models. Next, we reveal the cause of Wavenet + MVMT's under-performance on the electricity dataset over the first horizon with respect to MAPE. As shown in Fig. \ref{fig:elec_sample}, electricity data follows a long-tailed distribution -- a certain proportion exceeds a relatively high level. Recall that the optimization involves minimizing mean squared error, which means that more weights are placed on large errors. Moreover, every sample is treated equivalently in the estimation of global statistics. Therefore, the model can fit long-tailed samples better, but at the cost of degrading the fitness on normal samples.

We further investigate the improvement in terms of an individual variable in MTS. To prove that MVMT captures the difference in scaling factors among various variables, we characterize a variable with the mean of its historical observations. For succinctness, we calculate the variable-wise reduction on RMSE obtained by Wavenet + MVMT compared to AGCRN, and then plot the reduction against the scale of each variable in Figure \ref{fig:scale_error}. We can see that for BikeNYC and electricity, the improvement becomes more prominent as the scale grows, which meets our previous expectation. For PeMSD7, the improvement is less significant, as all the variables in this dataset vary in the same range, which signifies that they have approximately the same scaling factor.

\begin{figure}[tb]
\centering
\subfloat[BikeNYC]{
\includegraphics[width=0.45\linewidth]{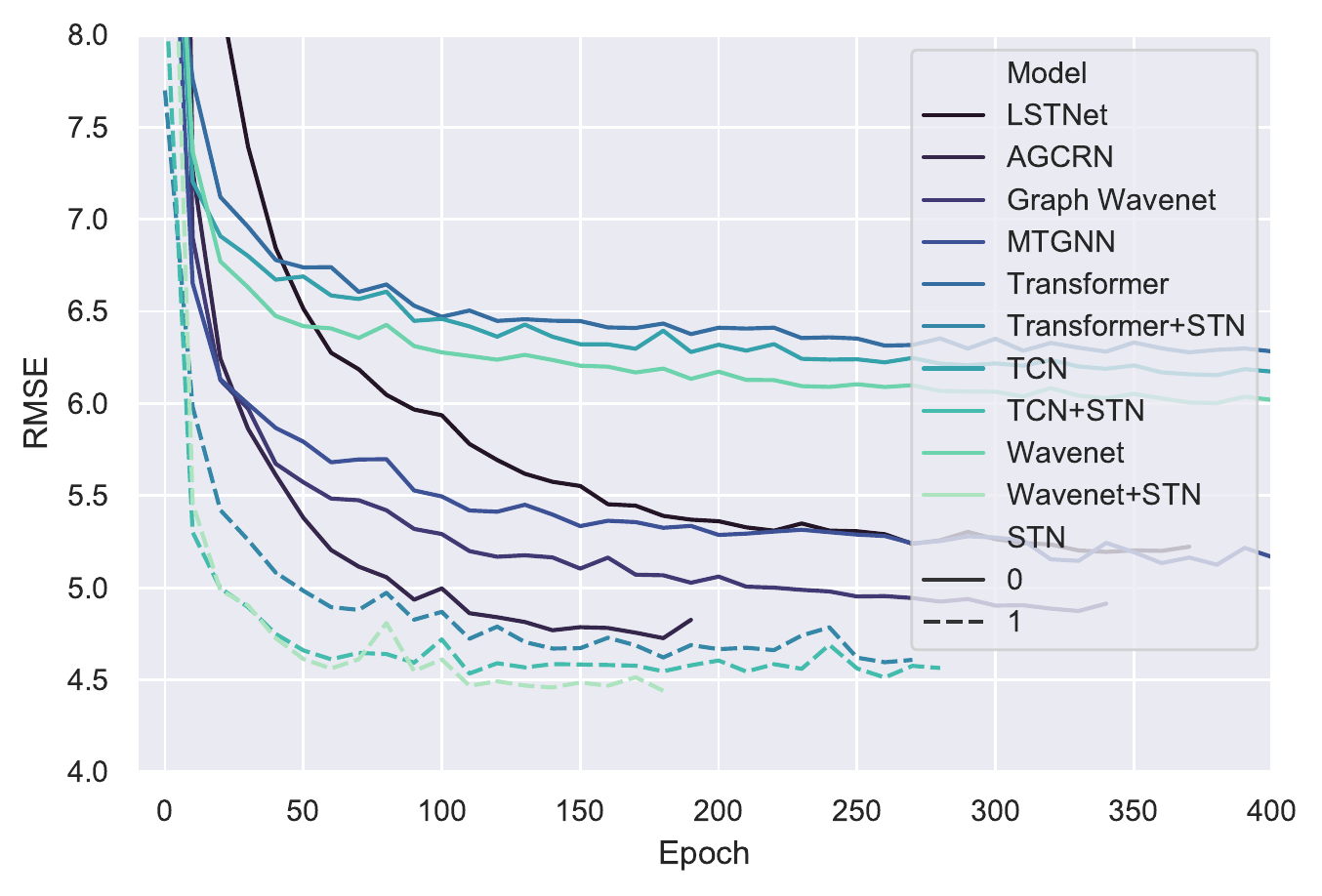}
}
\subfloat[PeMSD7]{
\includegraphics[width=0.45\linewidth]{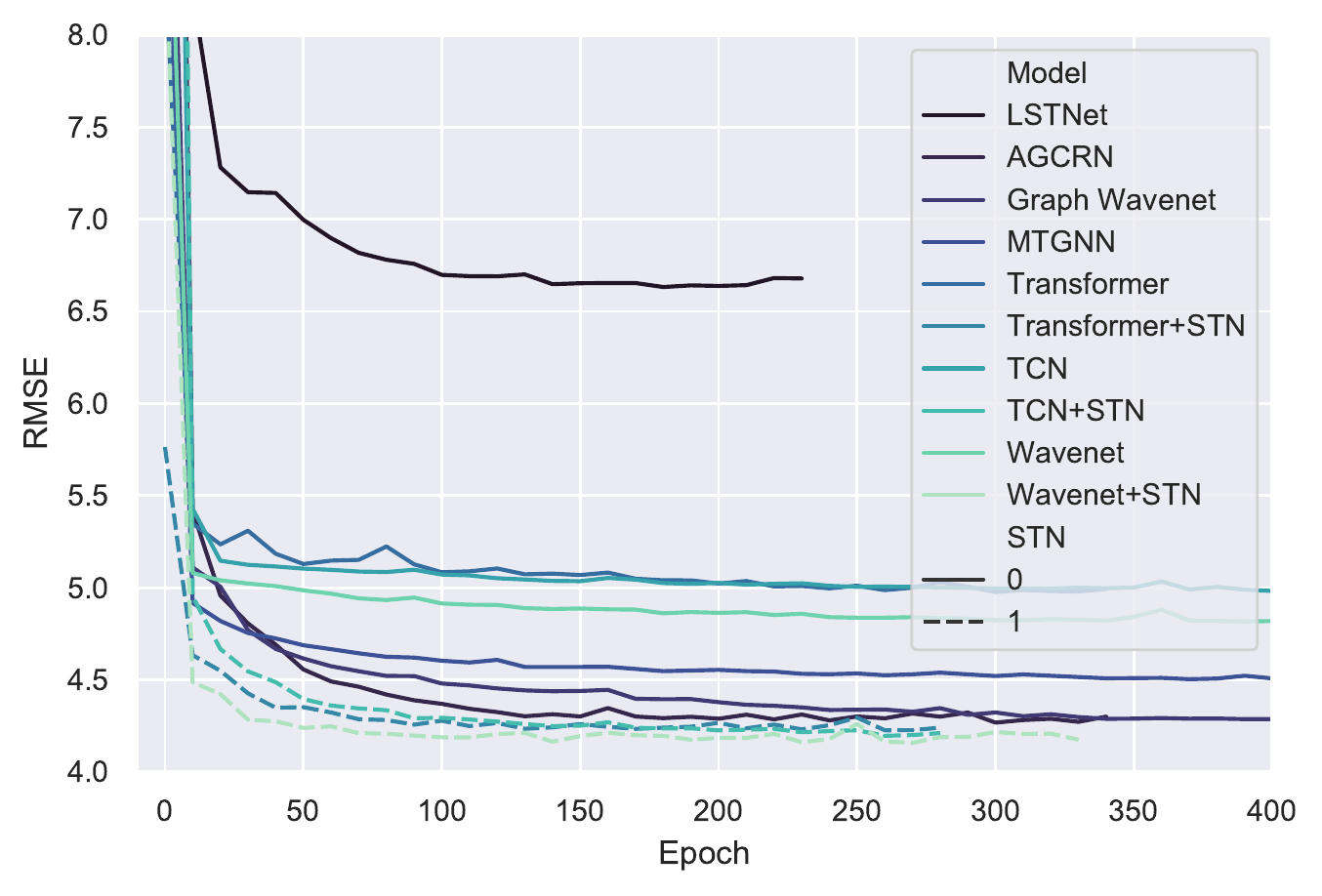}}

\subfloat[Electricity]{
\includegraphics[width=0.45\linewidth]{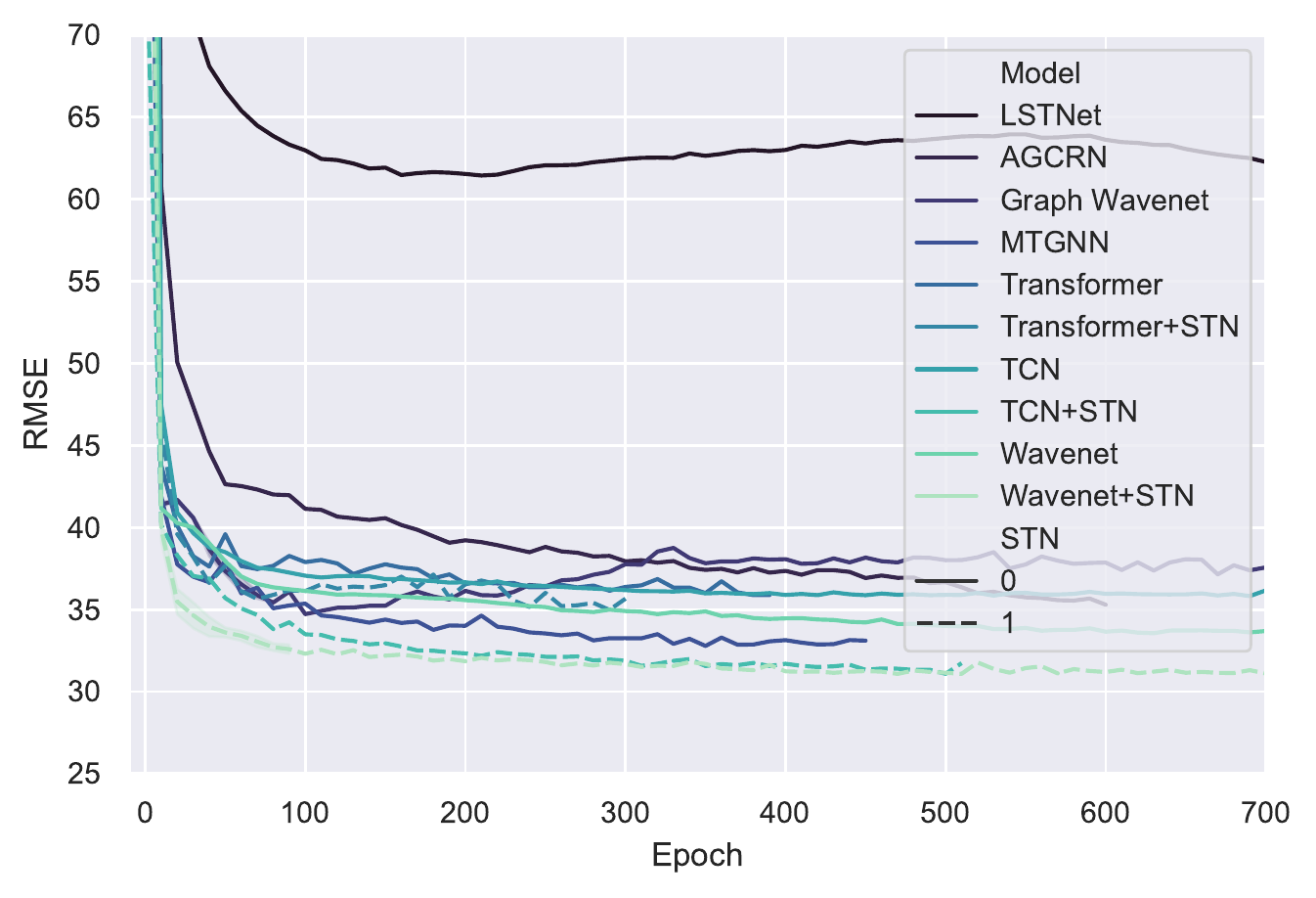}}
\caption{Loss convergence.}
\label{fig:converge}
\end{figure}

Fig. \ref{fig:converge} illustrates the process of loss convergence. It shows that with the additional MVMT module, the converging speeds of the models are accelerated by a large margin, faster than that of nearly all the baseline models.

\subsection{Ablation Study}

\begin{table}[tbh]
\tiny
\centering
\caption{Ablation Study}
\label{tab:ablation}
\begin{tabular}{cc|cccccc}
\hline
\multicolumn{2}{c}{}                                     & (a)     & (b)          & (c)  & (d)     & (e) & (f) \\ \hline
\multicolumn{1}{c|}{\multirow{3}{*}{B}}     & RMSE & \textbf{5.32}   & \underline{5.40}  & 6.34   & 6.15   & 6.48   & 7.55    \\ \cline{2-8} 
\multicolumn{1}{c|}{}                             & MAE  & \textbf{2.60}   & \underline{2.65}   & 3.04   & 2.93   & 3.07   & 3.53    \\ \cline{2-8} 
\multicolumn{1}{c|}{}                             & MAPE & \textbf{19.1\%} & \underline{19.5\%} & 22.9\% & 21.7\% & 23.4\% & 26.3\%  \\ \hline
\multicolumn{1}{c|}{\multirow{3}{*}{P}}      & RMSE & \textbf{5.25}   & \underline{5.38}   & 5.99   & 6.12   & 6.00   & 6.78    \\ \cline{2-8} 
\multicolumn{1}{c|}{}                             & MAE  & \textbf{2.88}   & \underline{3.01}   & 3.39   & 3.44   & 3.42   & 3.89    \\ \cline{2-8} 
\multicolumn{1}{c|}{}                             & MAPE & \textbf{6.02\%} & \underline{6.28\%} & 7.15\% & 7.20\% & 7.20\% & 8.23\%  \\ \hline
\multicolumn{1}{c|}{\multirow{3}{*}{E}} & RMSE & \textbf{38.9}   & \underline{41.0}   & 43.1   & 42.9   & 46.4   &     47.2    \\ \cline{2-8} 
\multicolumn{1}{c|}{}                             & MAE  & \textbf{18.8}   & 21.1   & 21.4   & \underline{20.2}   & 23.9   &    22.8     \\ \cline{2-8} 
\multicolumn{1}{c|}{}                             & MAPE & \textbf{14.2\%} & 17.7\% & 16.2\% & \underline{14.9\%} & 18.9\% &    16.4\%     \\ \hline
\end{tabular}
\end{table}

\begin{figure*}[tbh]
\centering
\subfloat[]{
\includegraphics[width=0.25\linewidth]{figure/ablation/MVMT.pdf}}
\subfloat[]{
\includegraphics[width=0.25\linewidth]{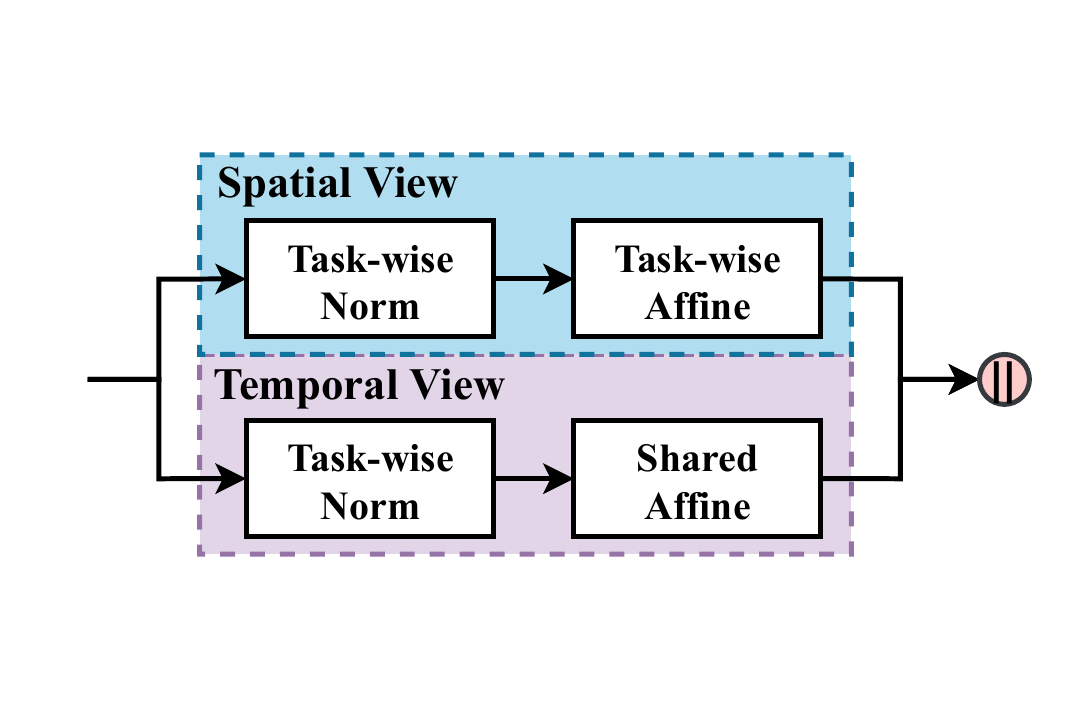}}
\subfloat[]{
\includegraphics[width=0.25\linewidth]{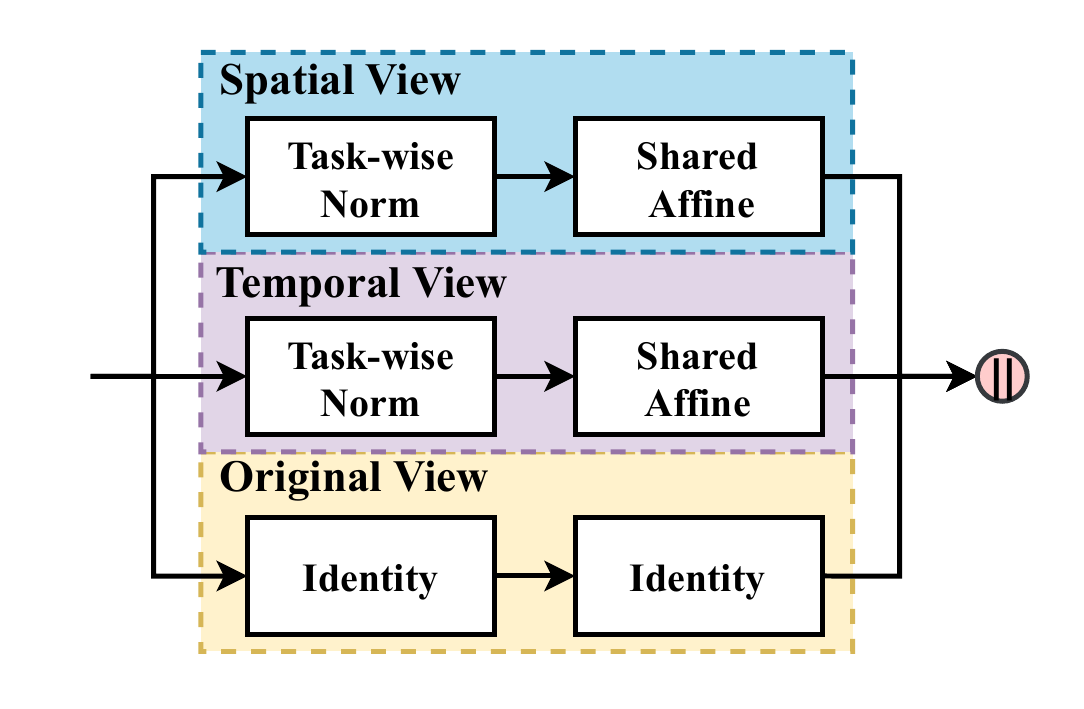}}

\subfloat[]{
\includegraphics[width=0.25\linewidth]{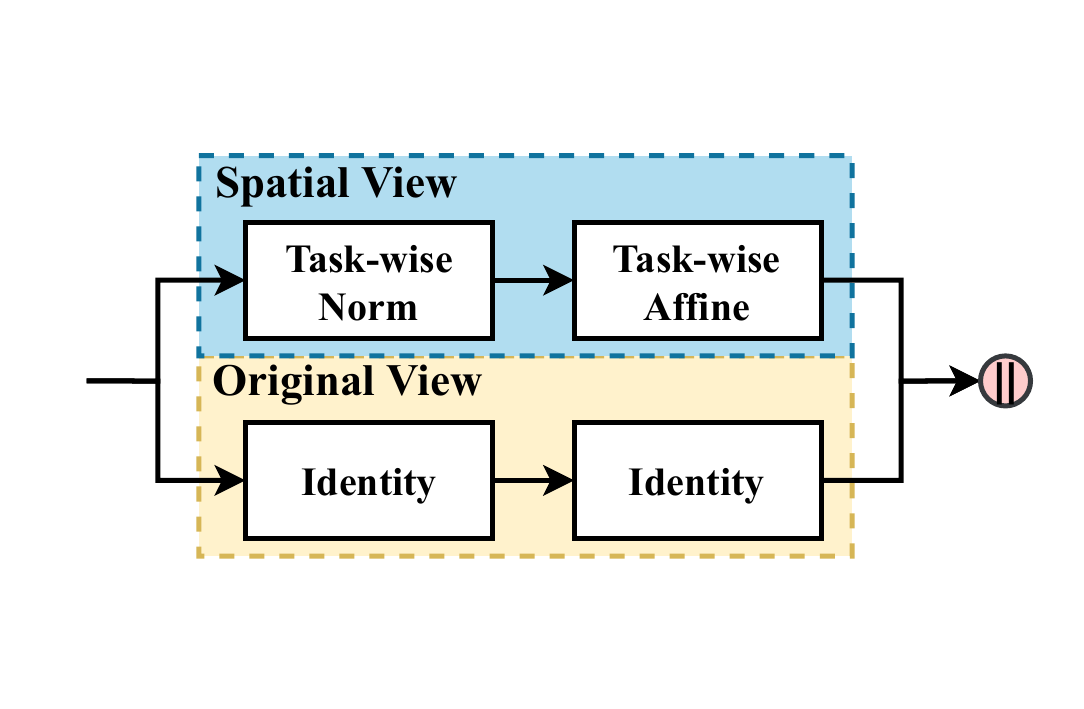}}
\subfloat[]{
\includegraphics[width=0.25\linewidth]{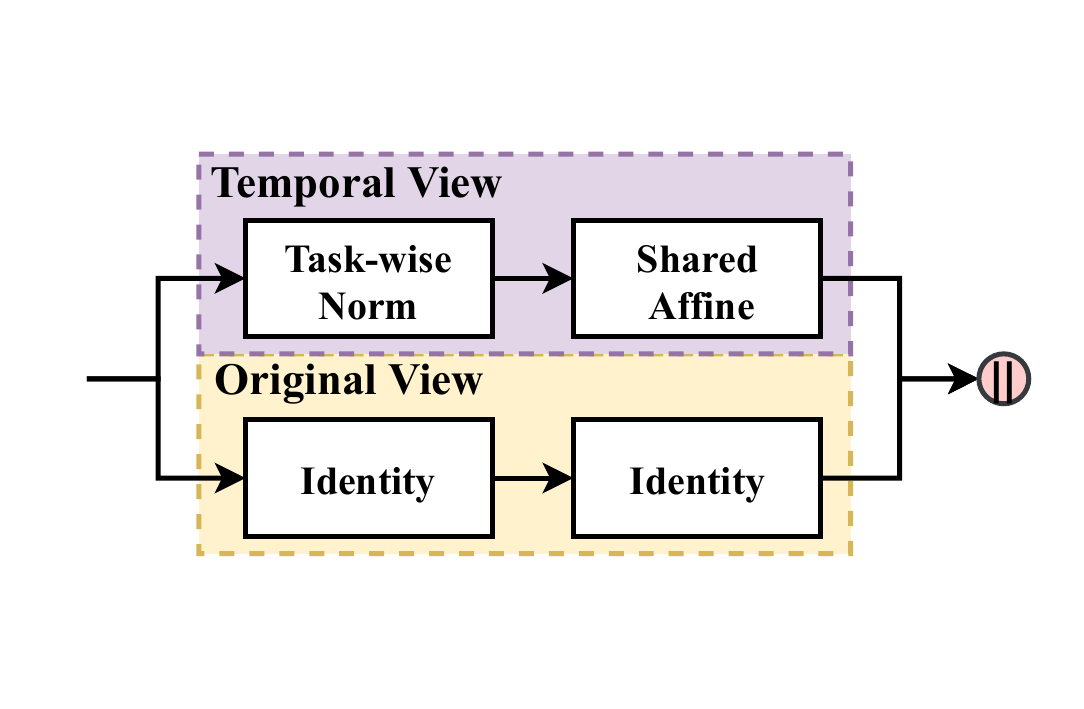}}
\subfloat[]{
\includegraphics[width=0.25\linewidth]{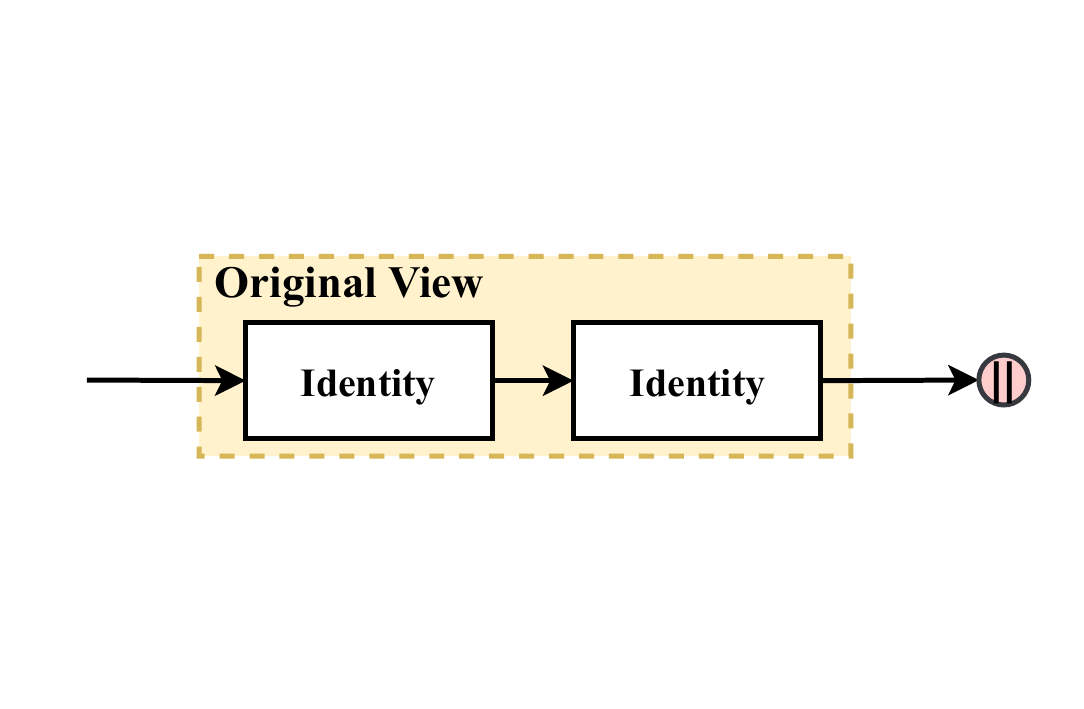}}
\caption{Variants for ablation study.}
\label{fig:ablation}
\end{figure*}

We design several variants of the MVMT block as shown in Figure \ref{fig:ablation} to validate the effectiveness of different operations.

We evaluate these variants on all the three datasets and report the results in Table \ref{tab:ablation}. From this table, we can draw the following major conclusions:
\begin{itemize}
    \item Taking either the spatial view or the temporal view can greatly enhance the capability of vanilla Wavenet, and taking both achieves the best performance.
    \item Contrasting (a) with (b), we conclude that the original view is indispensable in the MVMT block, especially when applied on the electricity data. An intuitive explanation for this indispensability is that task-wise normalization more or less loses information encoded in the original representation, especially for data which presents complex dynamics like the electricity dataset.
    \item Based on (a), (c) and (e), we find that task-wise affine transformation significantly increases the improvement contributed by the spatial branch. The reason for this is that this operation links the tasks associated with the same region over time, which is paramount for online forecasting where only recent observations are input to the model.
    \item Based on (a), (d), (e) and (f), we find that the performance gain entailed by merely taking the temporal view is limited on the electricity data. We conjecture that this is also due to the complex patterns of the data: without encoding the spatial attribute into the representations, it is difficult for the model to capture typical temporal patterns from data.
\end{itemize}

\subsection{Hyper-parameter Analysis}

\begin{figure}[tb]
\centering
\subfloat[]{
\includegraphics[width=0.4\linewidth]{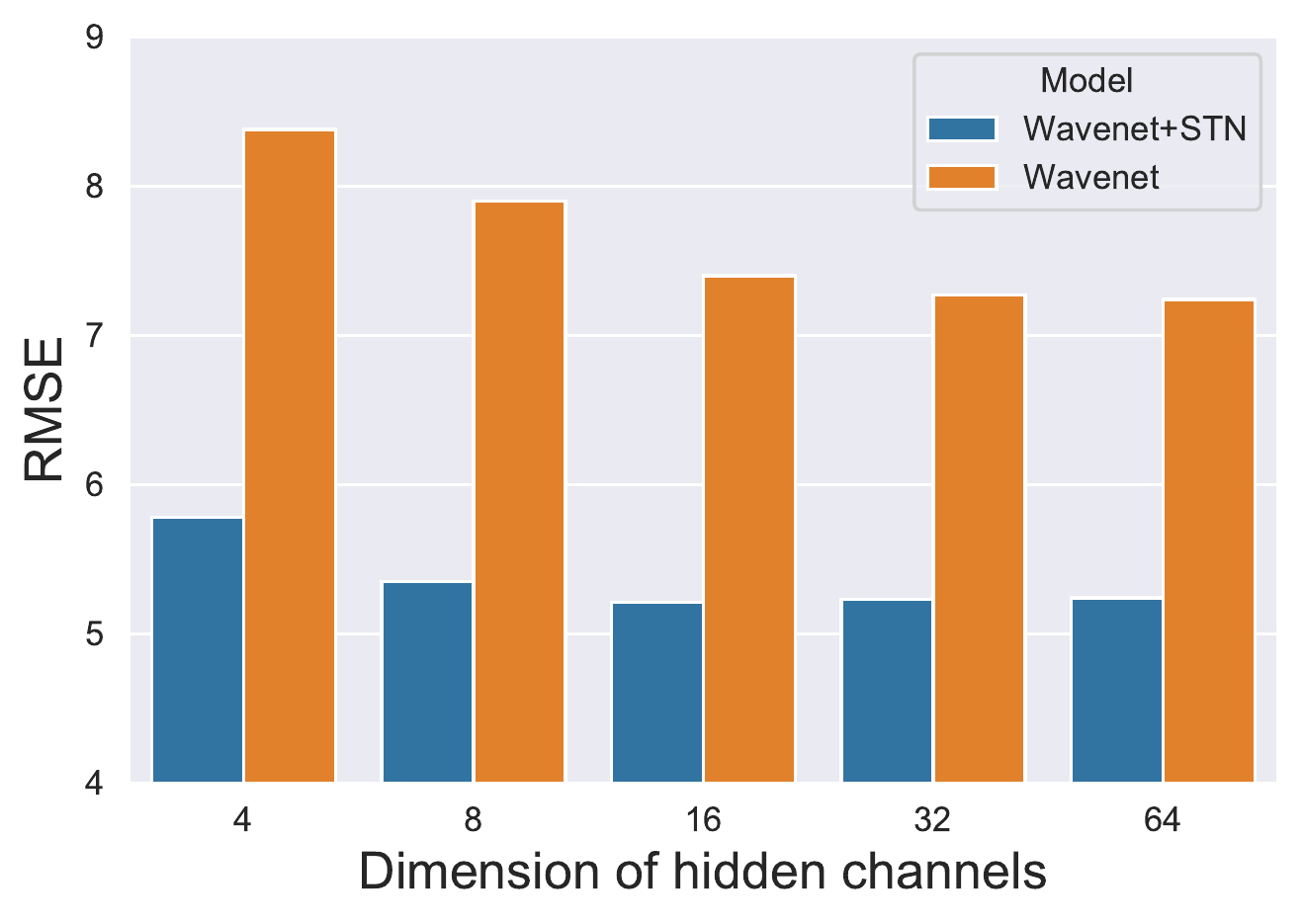}
}
\subfloat[]{\includegraphics[width=0.4\linewidth]{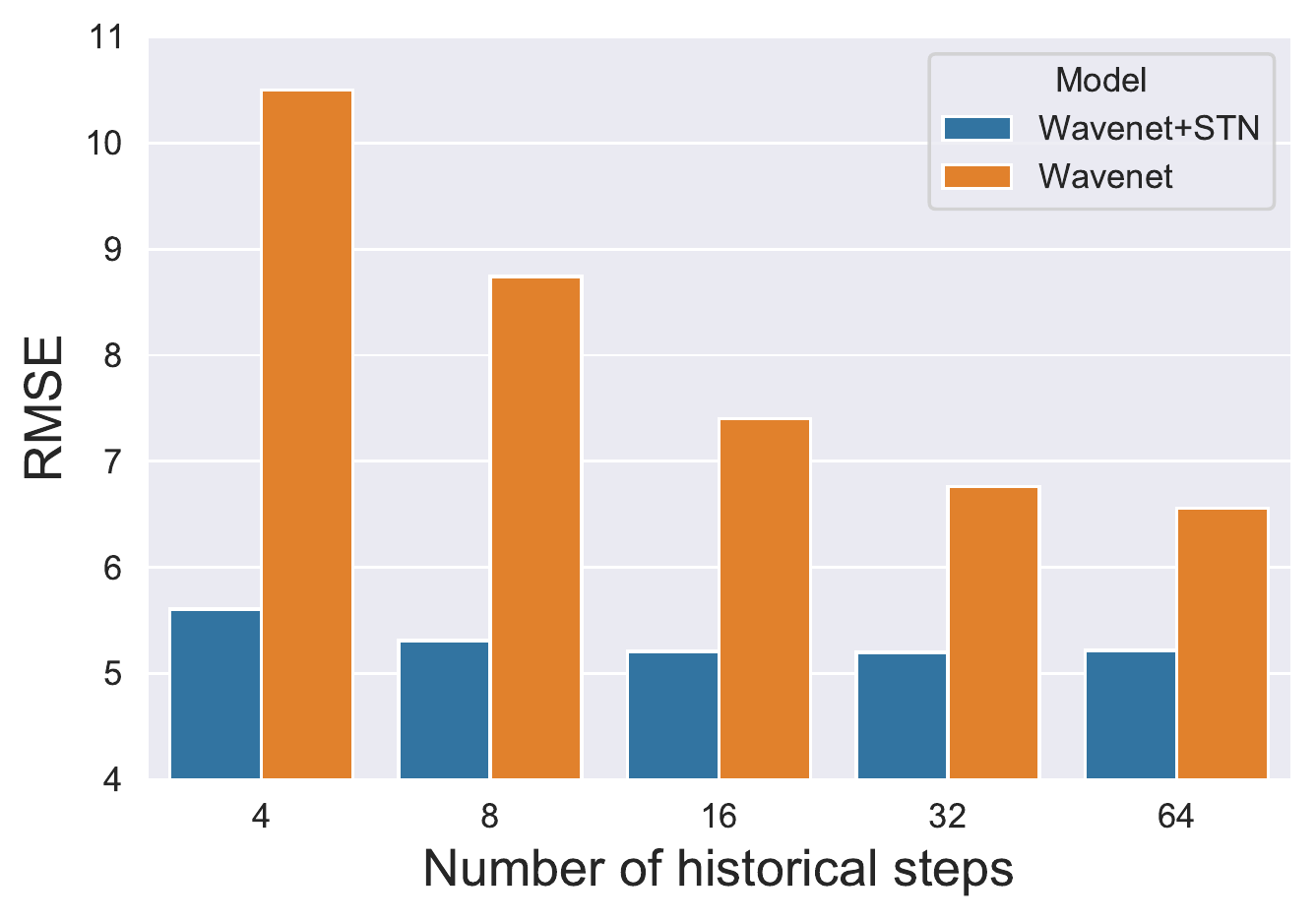}}

\subfloat[]{\includegraphics[width=0.4\linewidth]{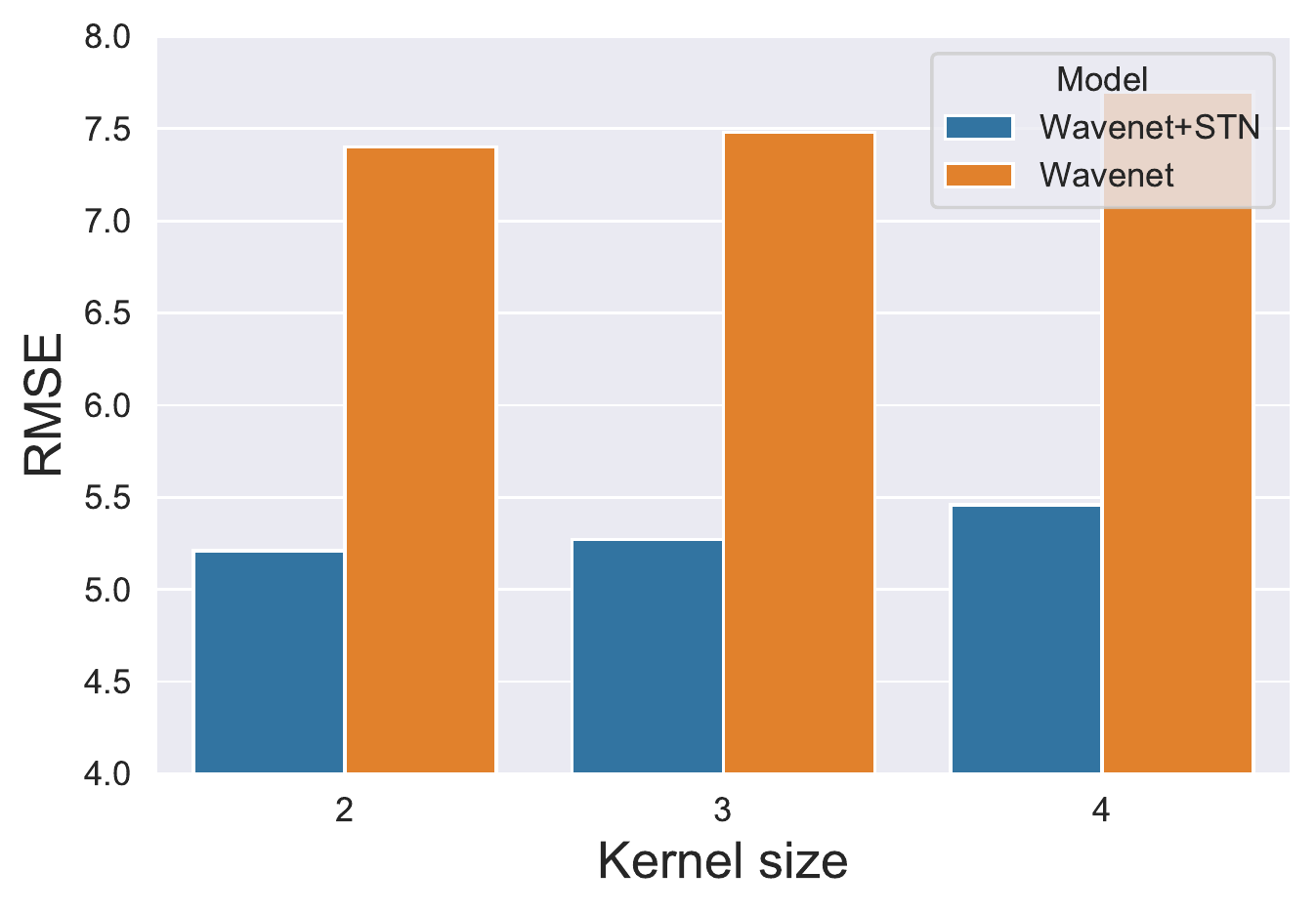}}
\subfloat[]{\includegraphics[width=0.4\linewidth]{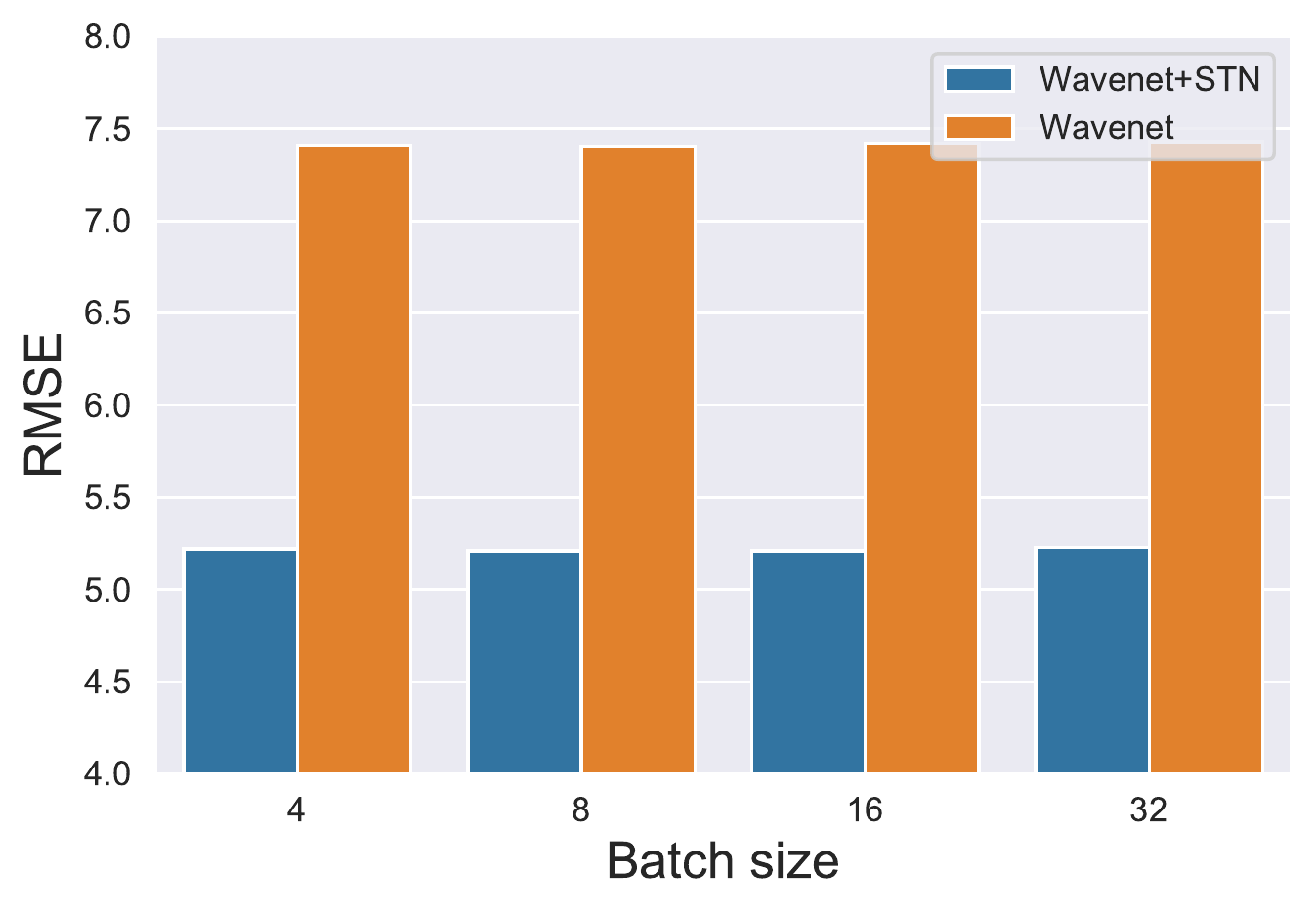}}
\caption{Hyper-parameter analysis.}
\label{fig:hyper}
\end{figure}

We further study the effect of different settings of the hyper-parameters in the proposed modules. There are four hyper-parameters need to be manually set by practitioners, namely the dimension of hidden channels $d_z$, the number of historical steps input to the model, the kernel size of DCC and the batch size. The study results are reported in Fig. \ref{fig:hyper}, from which we can draw a major conclusion: MVMT not only boosts the performance, but also increases the stability of the performance under different hyper-parameter settings.

\subsection{Case Study}

To obtain more insights on the algorithm, we conduct multiple studies to qualitatively analyze the representations generated while forecasting, including the initial representation, the intermediate representation and the final representation. The dataset we select for this investigation is BikeNYC.

\subsubsection{Initial Representation}

\begin{figure}[tb]
\centering
\subfloat[]{
\includegraphics[width=0.4\linewidth]{figure/x_xm1_region_corr.pdf}}
\subfloat[]{
\includegraphics[width=0.4\linewidth]{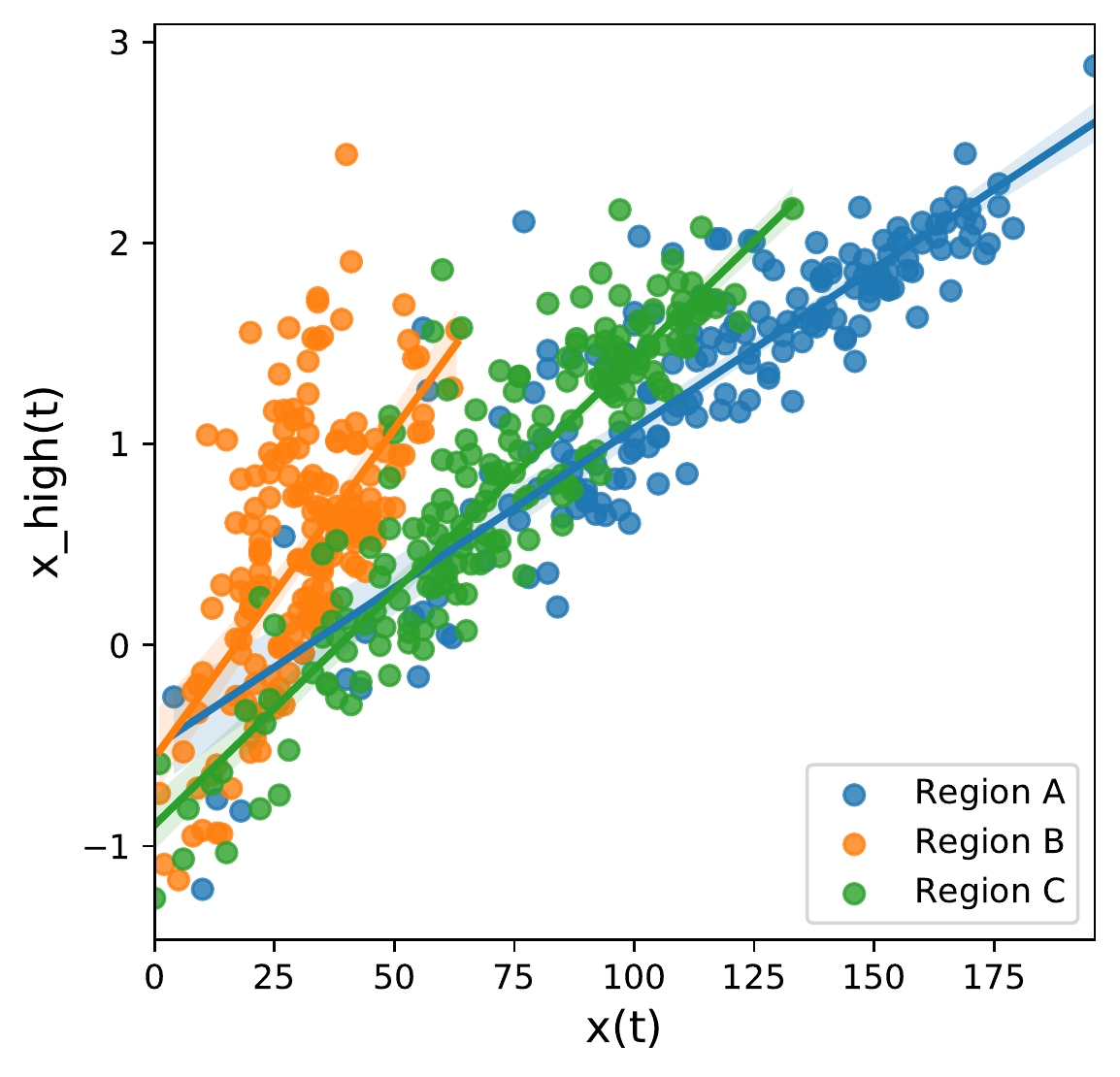}
\label{fig:x_xt}}
\caption{Initial representations produced: (a) without task-wise normalization; (b) with task-wise normalization from the temporal view.}
\end{figure}

\begin{figure}[tb]
\centering
\subfloat[]{
\includegraphics[width=0.4\linewidth]{figure/x_xm1_time_corr.pdf}}
\subfloat[]{
\includegraphics[width=0.4\linewidth]{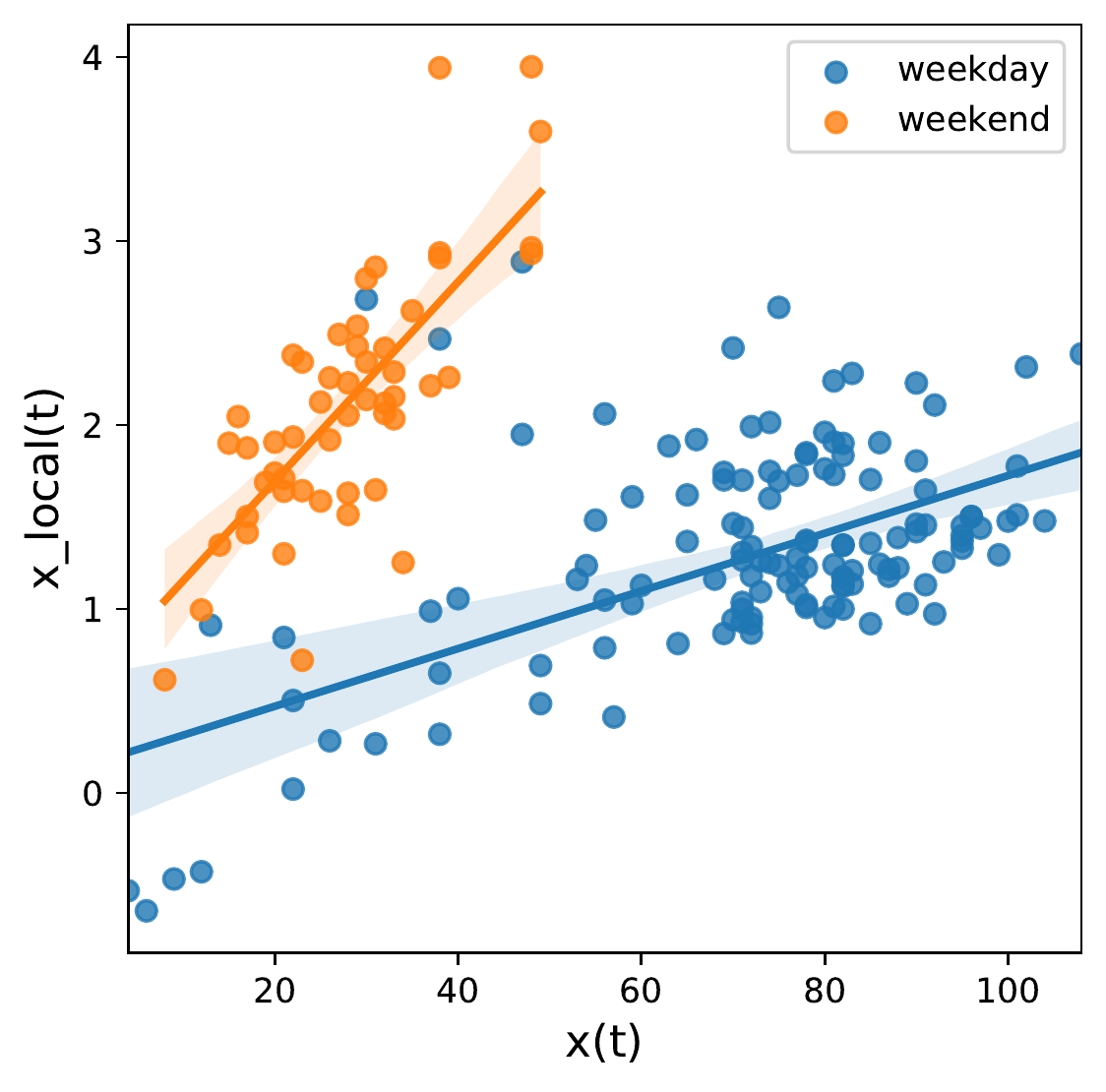}
\label{fig:x_xs}}
\caption{Initial representations produced: (a) without task-wise normalization; (b) with task-wise normalization from the spatial view.}
\end{figure}

We apply task-wise normalization over the raw input data respectively from the spatial view and the temporal view, and examine whether the issues we raise in Fig. \ref{fig:issue} are mitigated. We plot the original quantity versus the temporally normalized quantity in Fig. \ref{fig:x_xt}, and the original quantity versus the spatially normalized quantity in Fig. \ref{fig:x_xs}. It is apparent that the pairwise relationship between the original quantity and the temporally normalized quantity separates different regions, and the pairwise relationship between the original quantity and the spatially normalized quantity separates different days.

\subsubsection{Intermediate Representation}

\begin{figure}[tb]
    \centering
    \includegraphics[width=0.8\linewidth]{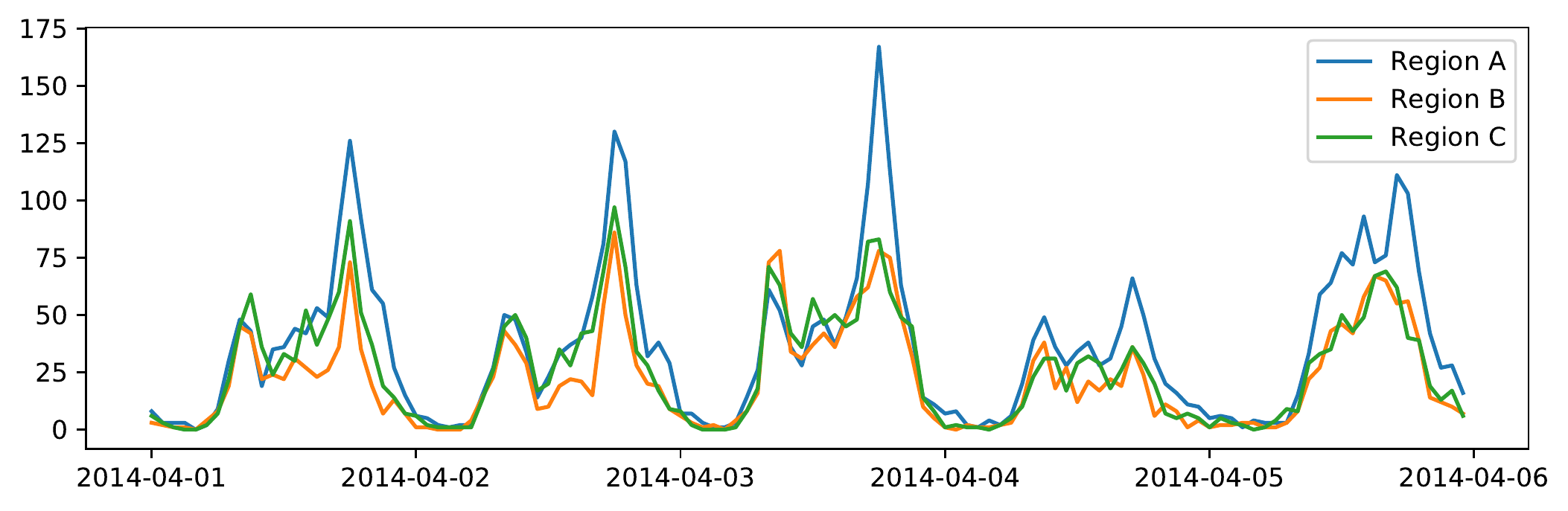}
    \caption{Example data selected to study the output of intermediate representations from the spatial view.}
    \label{fig:case_sn_sample}
\end{figure}

\begin{figure}[tb]
\centering
\subfloat[]{
\includegraphics[width=0.4\linewidth]{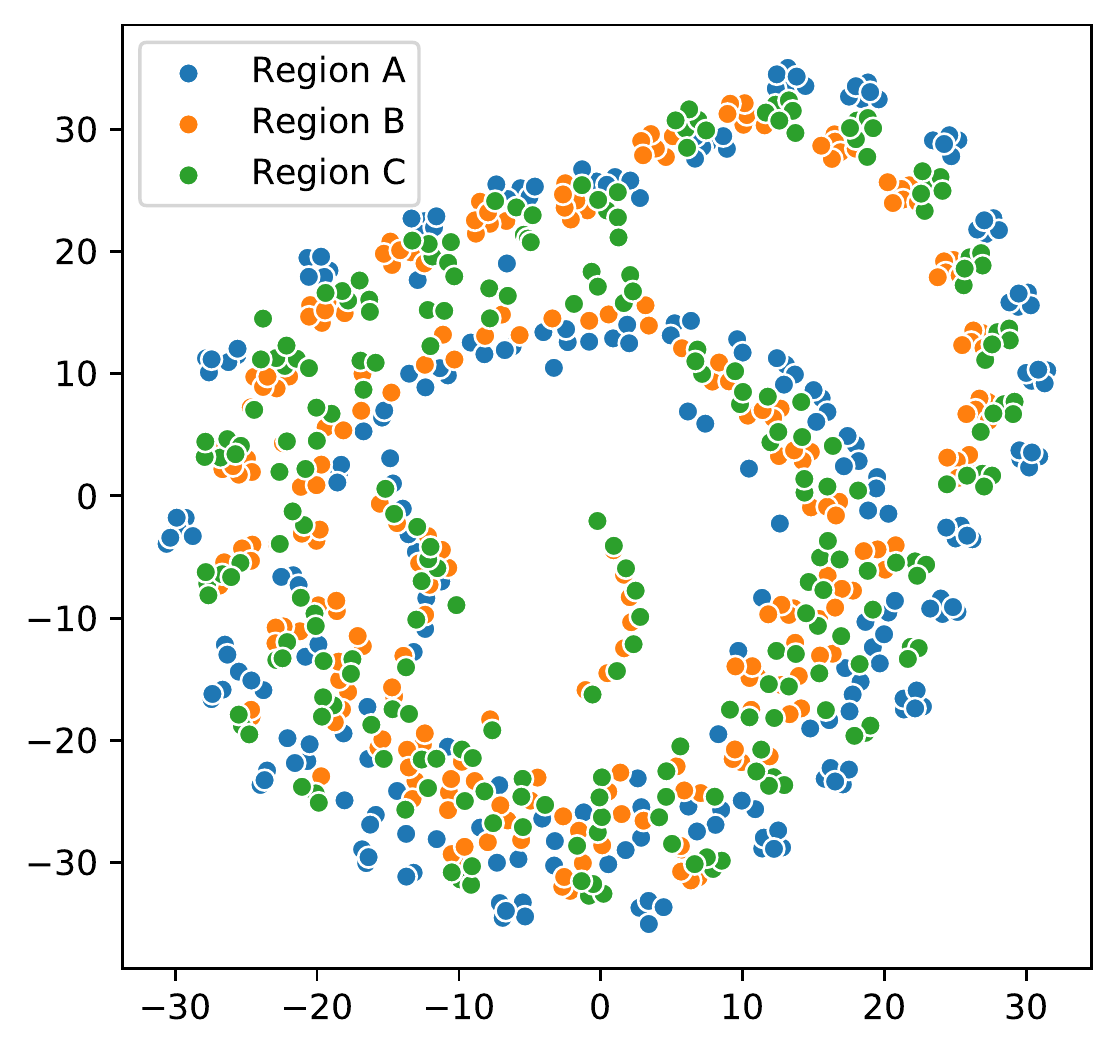}
\label{fig:case_sn_original}}
\subfloat[]{
\includegraphics[width=0.4\linewidth]{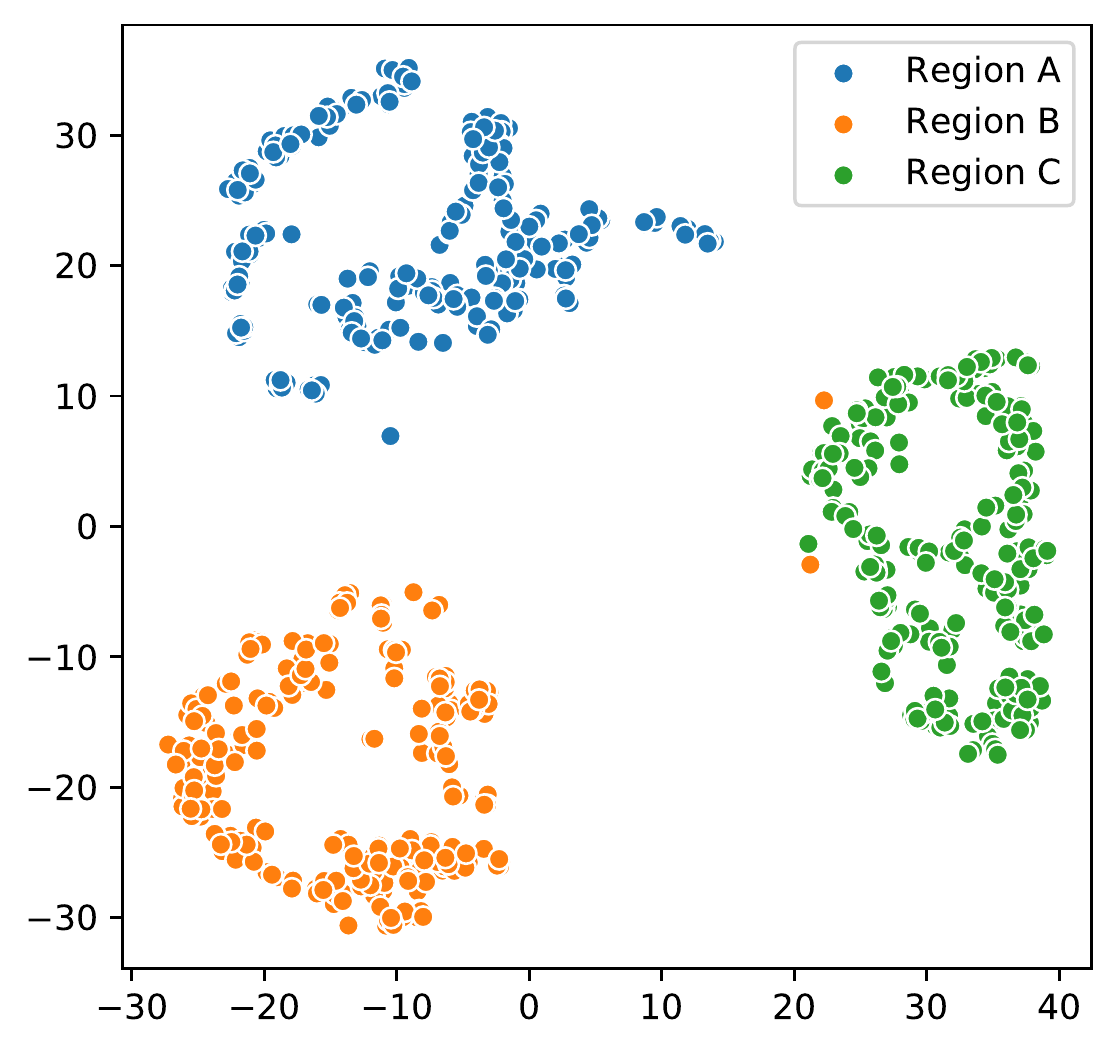}
\label{fig:case_sn}}
\caption{Intermediate representations produced by: (a) the vanilla Wavenet; (b) the spatial view in MVMT.}
\label{}
\end{figure}

\begin{figure}[tb]
    \centering
    \includegraphics[width=0.8\linewidth]{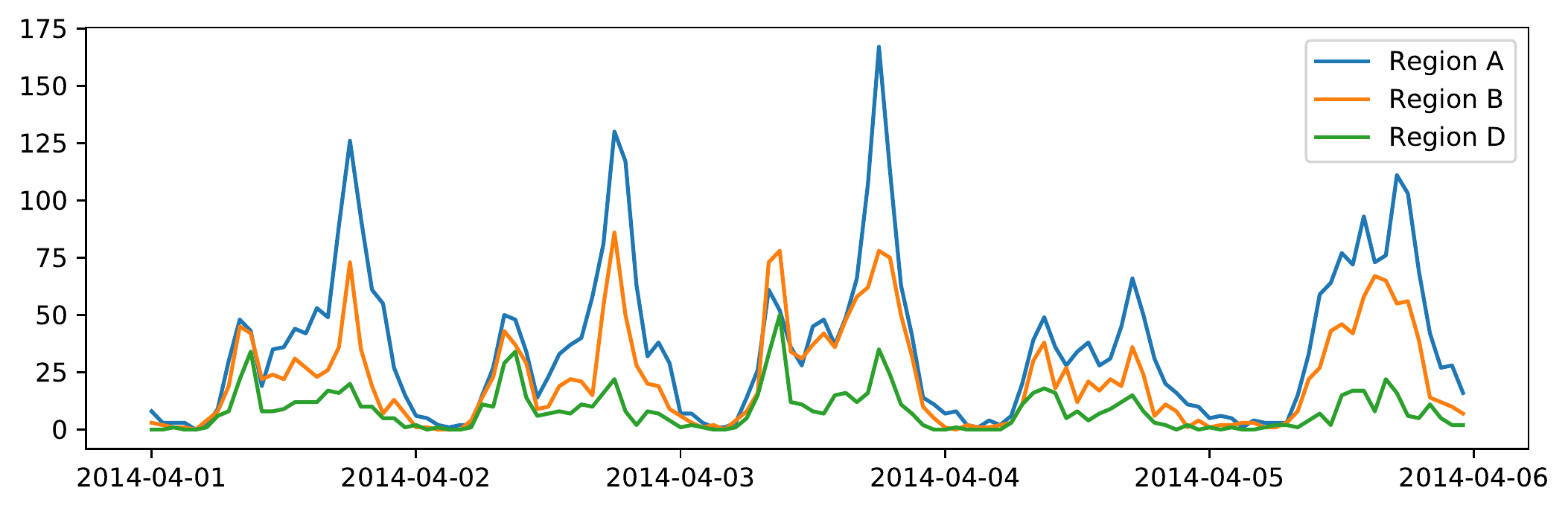}
    \caption{Example data picked out to study intermediate representations output from the temporal view.}
    \label{fig:case_tn_sample}
\end{figure}

\begin{figure}[tb]
\centering
\subfloat[]{
\includegraphics[width=0.4\linewidth]{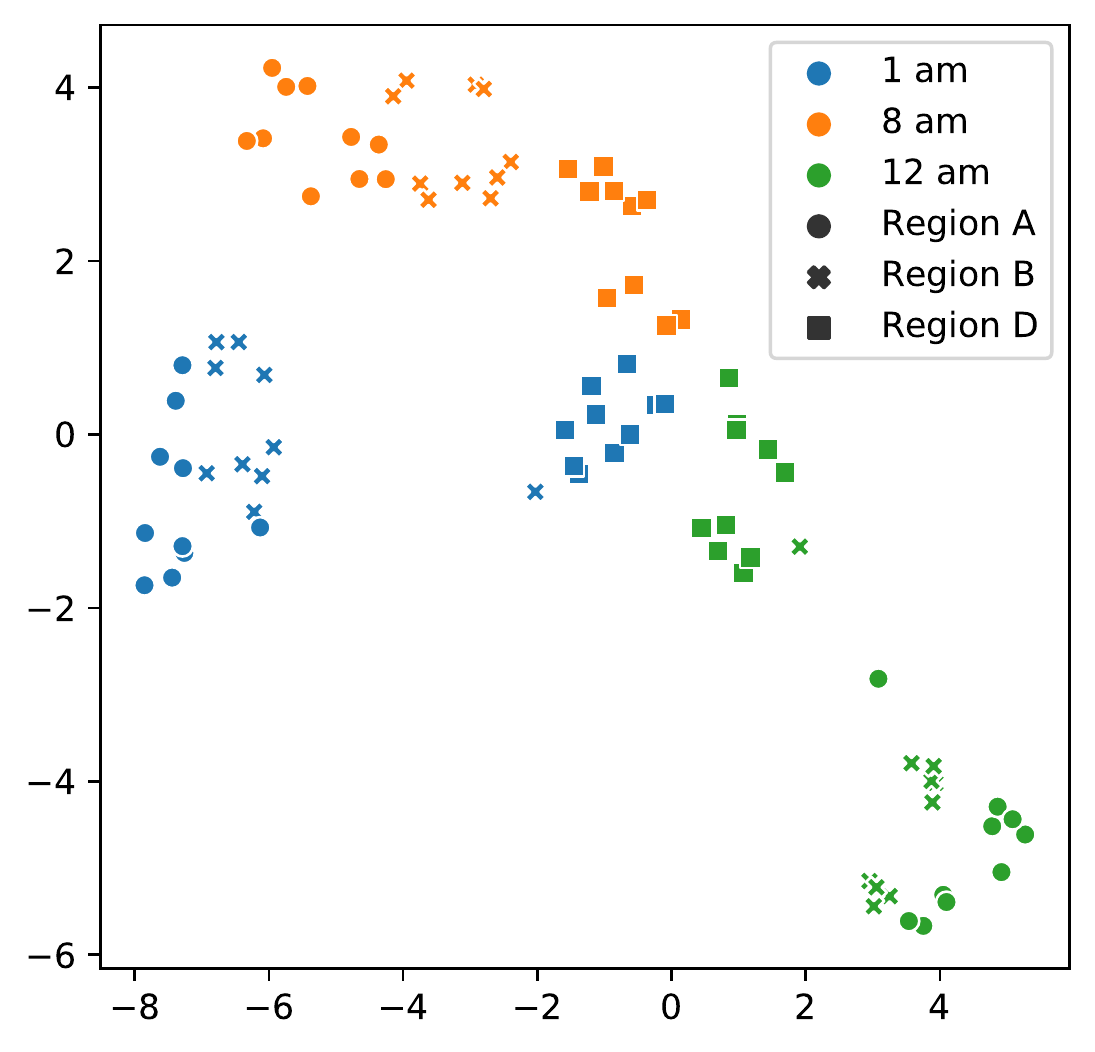}
\label{fig:case_tn_original}}
\subfloat[]{
\includegraphics[width=0.4\linewidth]{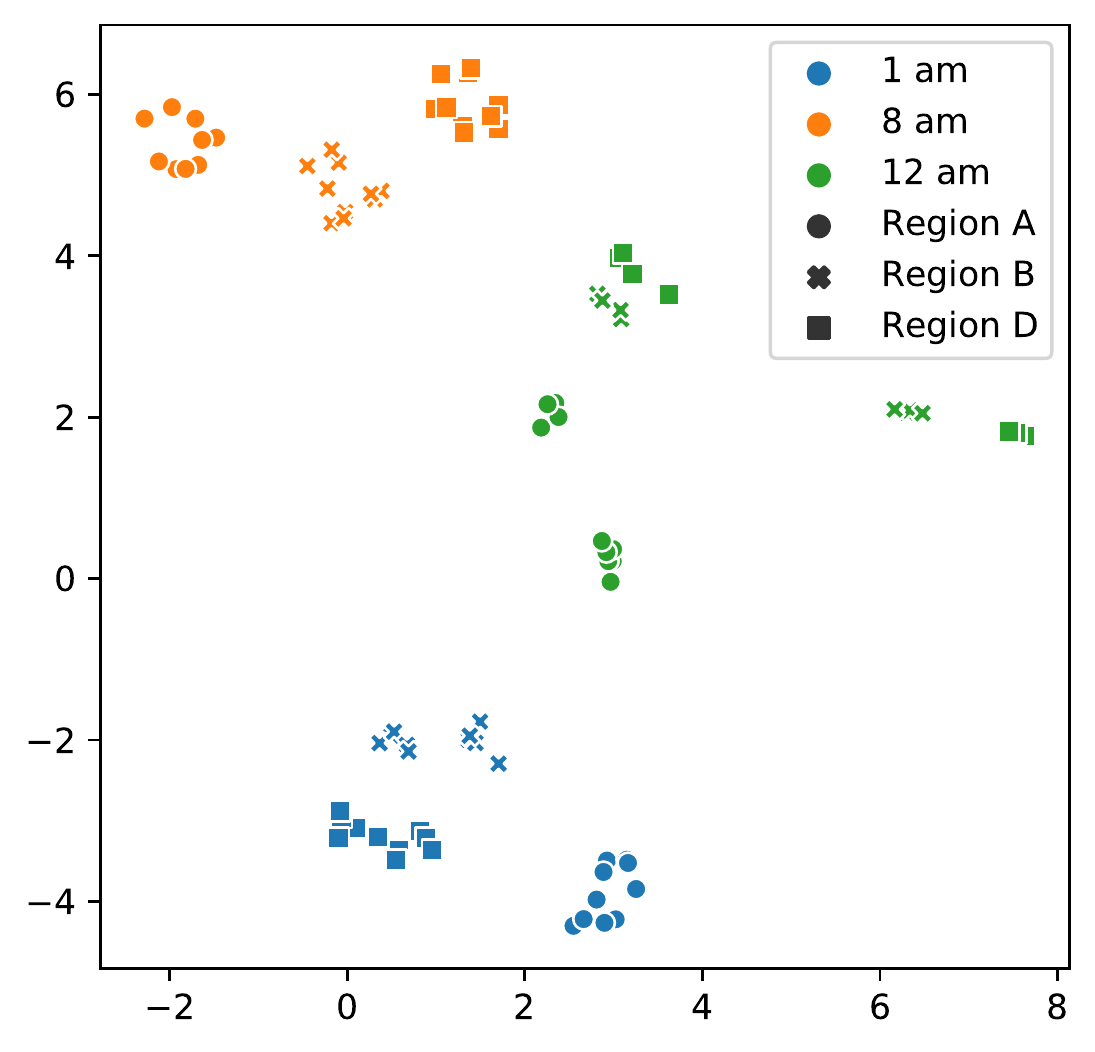}
\label{fig:case_tn}}
\caption{Intermediate representations produced by: (a) the vanilla Wavenet; (b) the temporal view in MVMT.}
\end{figure}

In this part, we investigate the output from the spatial branch and the temporal branch in MVMT. We start by discussing the property that these intermediate representations are supposed to express. The output from the spatial branch is expected to encode the local component which reflects the temporary pattern at a single timestamp. The reason for this is that task-wise normalization eliminates the global component from the input representation, where the global component encodes the long-term pattern regarding a region, which has nothing to do with a specific timestamp. With respect to the temporal branch, the resulting representation is deemed to show the region-wise pattern, which plays the role of a local component from this view.

We select three representative regions at specified times to reflect what the operations extract from the data. We collect the intermediate representation output from the two views in the top residual block and compress them via t-SNE for the sake of visualization. For comparison, we also inspect their associated input representations, each of which is a concatenation of raw measurements. Next, we discuss separately the outcomes from these two views in details.

\textbf{Temporal View.} In Fig. \ref{fig:case_sn_sample}, we display the demand evolution during a given period over the three investigated regions. We can observe that the three regions have similar evolution patterns, especially regions B and C. The representations concatenated by the original measurements are plotted in Fig. \ref{fig:case_sn_original}, and the output of the intermediate representations from the temporal view are plotted in Fig. \ref{fig:case_sn}. We can observe that the representations are completely rearranged in accordance with the regional identity. This observation demonstrates that the local components are roughly invariant within the group belonging to the same region. This coincides with our understanding that some regional attributes, such as population and functionality, are stable over time.

\textbf{Spatial View.}  To reflect the characteristics of the output of representations from the spatial view, we take another region D into consideration, as shown in Fig. \ref{fig:case_tn_sample}. Noticeably, the magnitude of the demand over region D is substantially smaller than those over regions A or B. Here, we account for three different times in a day, consisting of 1am, 8am and 12pm. Likewise, the input representations are plotted in Fig. \ref{fig:case_tn_original}, and the intermediate representations are plotted in Fig. \ref{fig:case_tn}. As shown in Fig. \ref{fig:case_tn_original}, instances belonging to region D are mixed up without separation between different times, which signifies that the model will struggle to differentiate the times at which those instances occurred. In contrast, this issue is mitigated in the representation space from the spatial view, as it forms clusters of instances with the same occurrence time.

\subsubsection{Final Representation}

In this part, we investigate the representations directly used for prediction. The major purpose is to examine the spatial distinguishability and temporal distinguishability of the representations after processing by MVMT. These two types of distinguishability are separately illustrated with two cases.

\begin{figure}
    \centering
\includegraphics[width=0.8\linewidth]{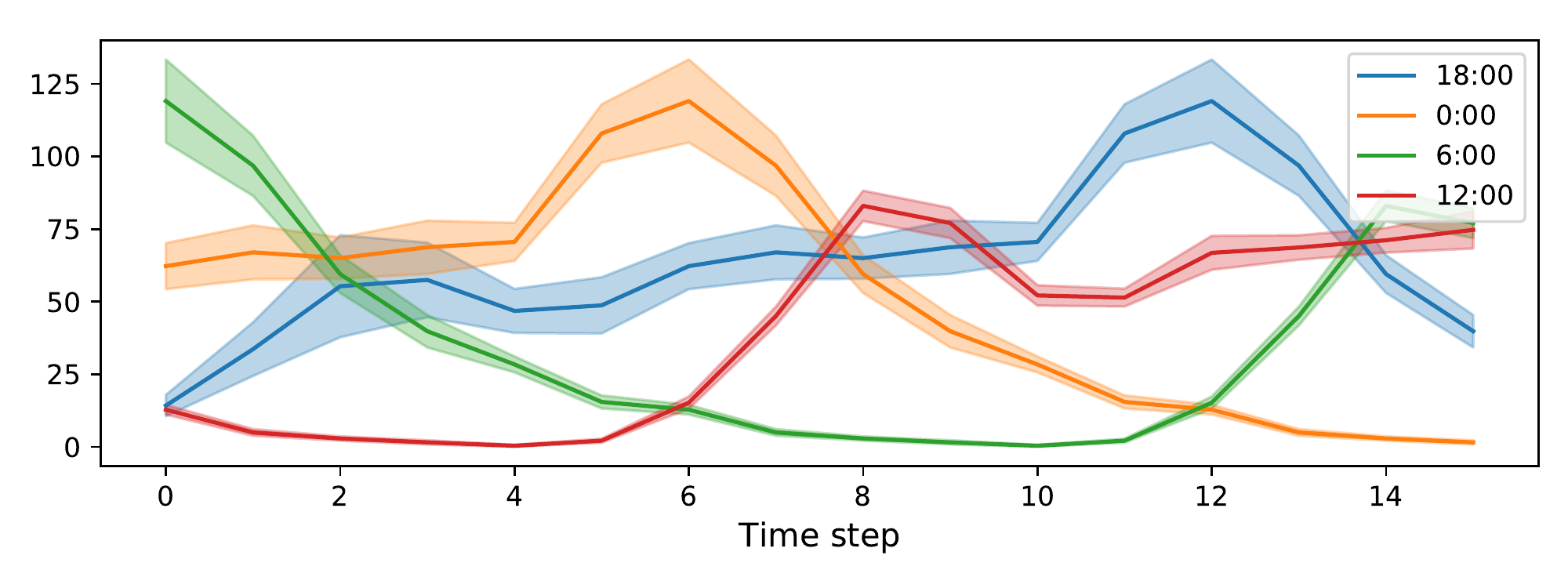}
\caption{Four groups of samples selected to inspect temporal distinguishability.}
\label{fig:pca_time_samples}
\end{figure}

\textbf{Temporal View.} In the first case, we examine the distinguishability from the temporal view, concentrating on a specific region. With prior knowledge that bike demand shows daily periodicity where the same evolving pattern appears iteratively at the same time every day, we select four representative times of a day which exhibit entirely different patterns. Then, we extract the input sequences at these times, and visualize them in Figure \ref{fig:pca_time_samples}, grouped by time.

\begin{figure}[tb]
\centering
\includegraphics[width=0.8\linewidth]{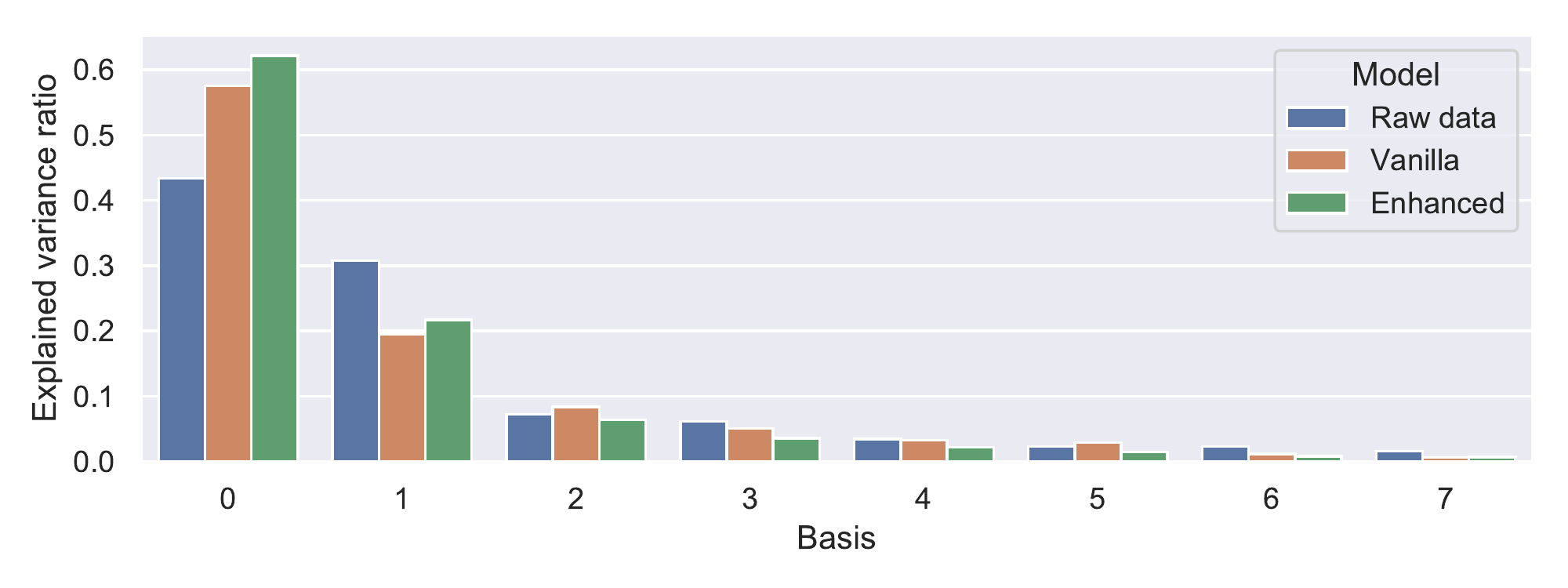}
\caption{The percentages of variance explained by the first seven components.}
\label{fig:pca_time_variance}
\end{figure}

\begin{figure}[tb]
\subfloat[Raw data]{
\includegraphics[width=0.32\linewidth]{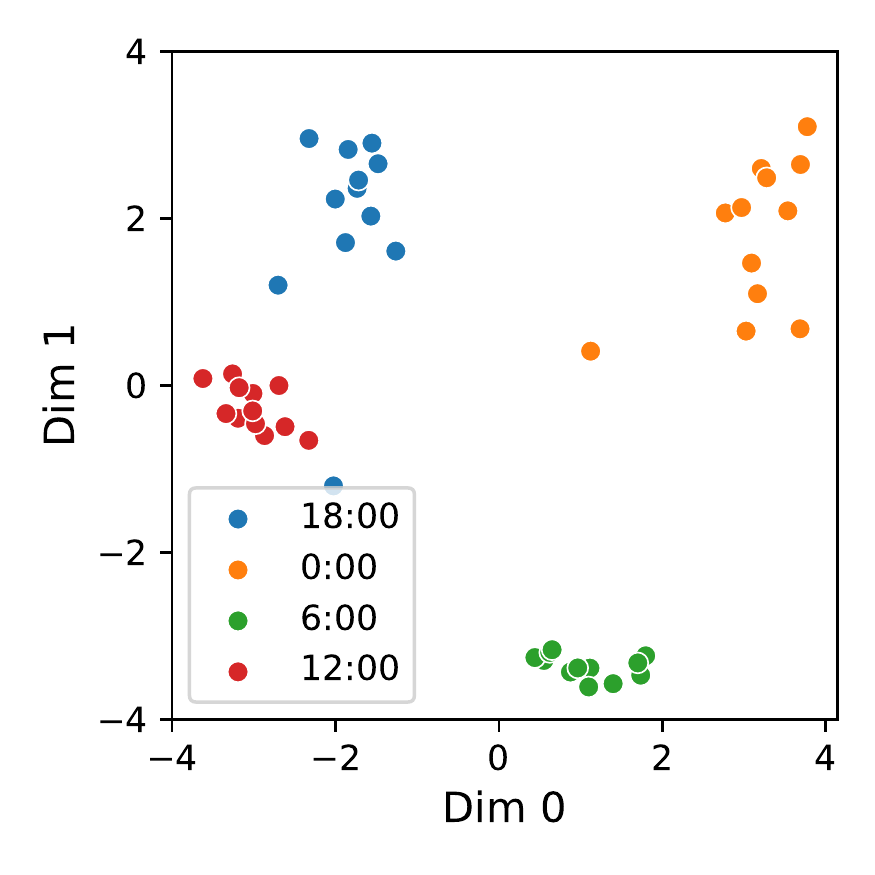}}
\subfloat[Vanilla]{
\includegraphics[width=0.32\linewidth]{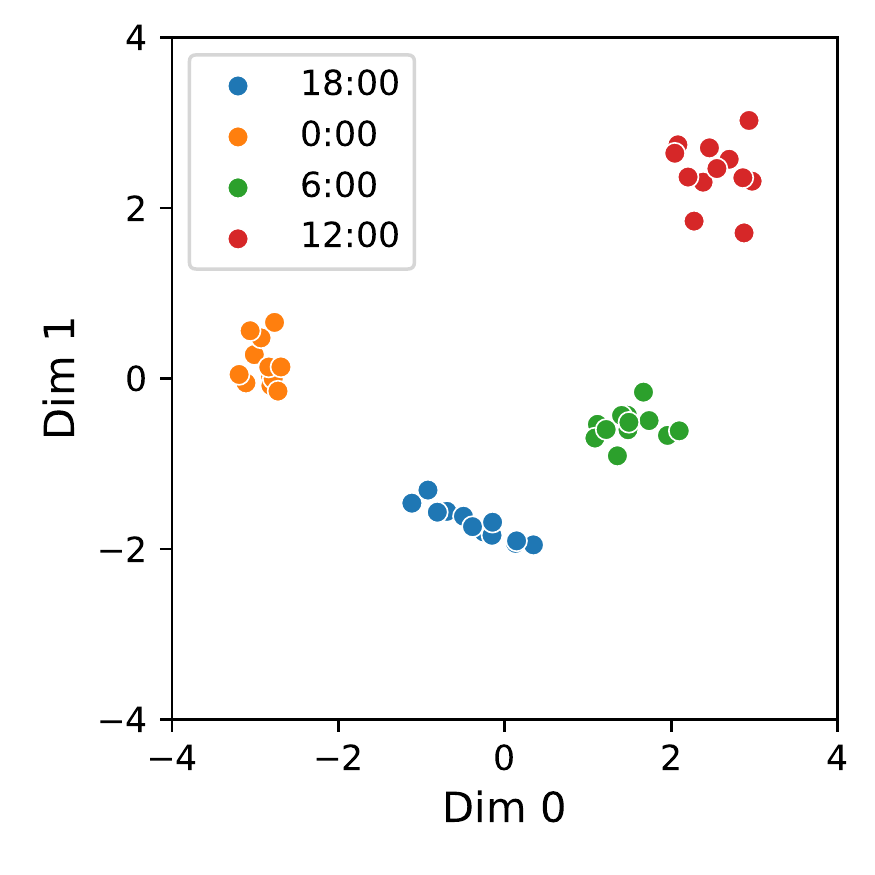}}
\subfloat[Enhanced]{
\includegraphics[width=0.32\linewidth]{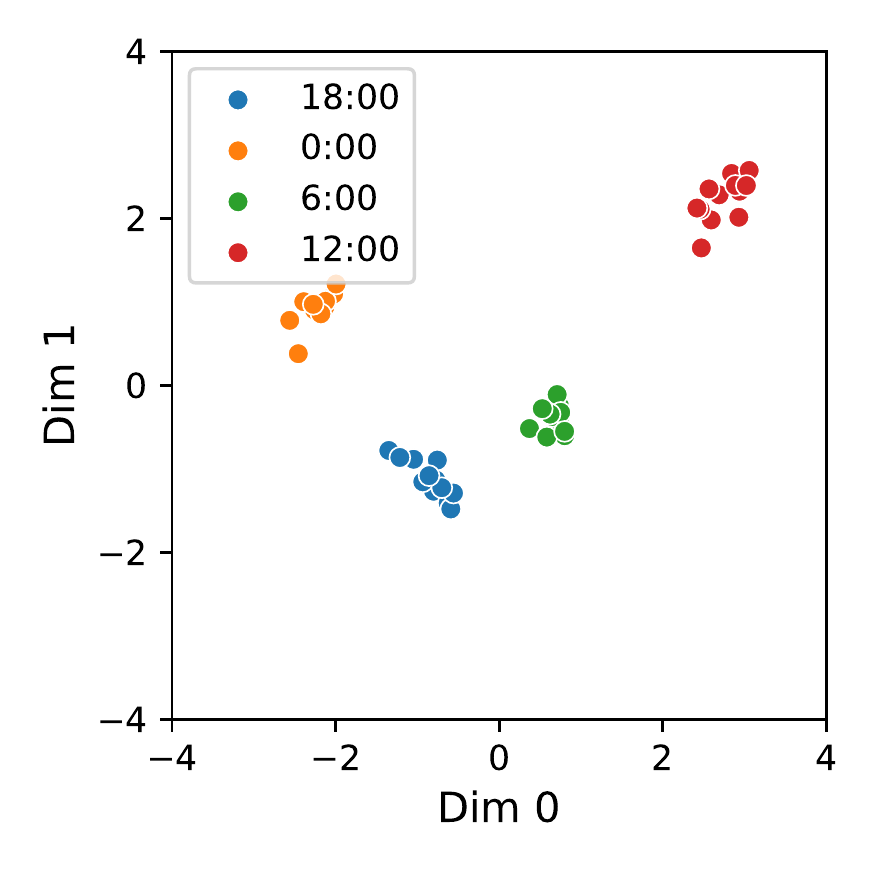}}

\subfloat[Raw data]{
\includegraphics[width=0.32\linewidth]{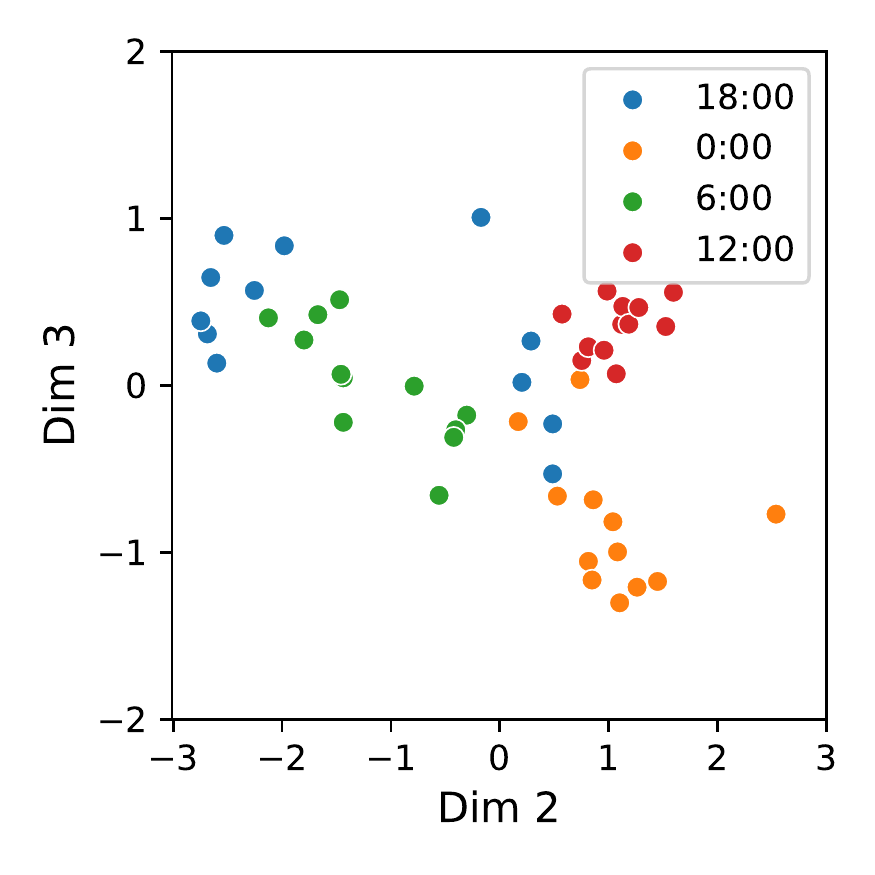}}
\subfloat[Vanilla]{
\includegraphics[width=0.32\linewidth]{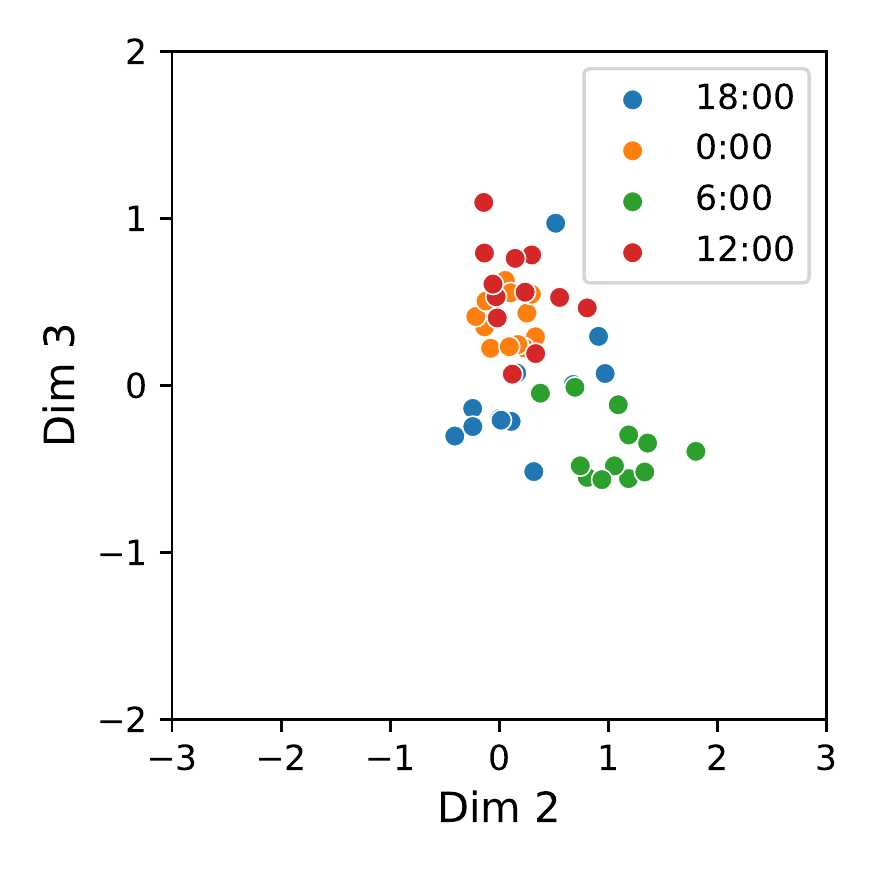}}
\subfloat[Enhanced]{
\includegraphics[width=0.32\linewidth]{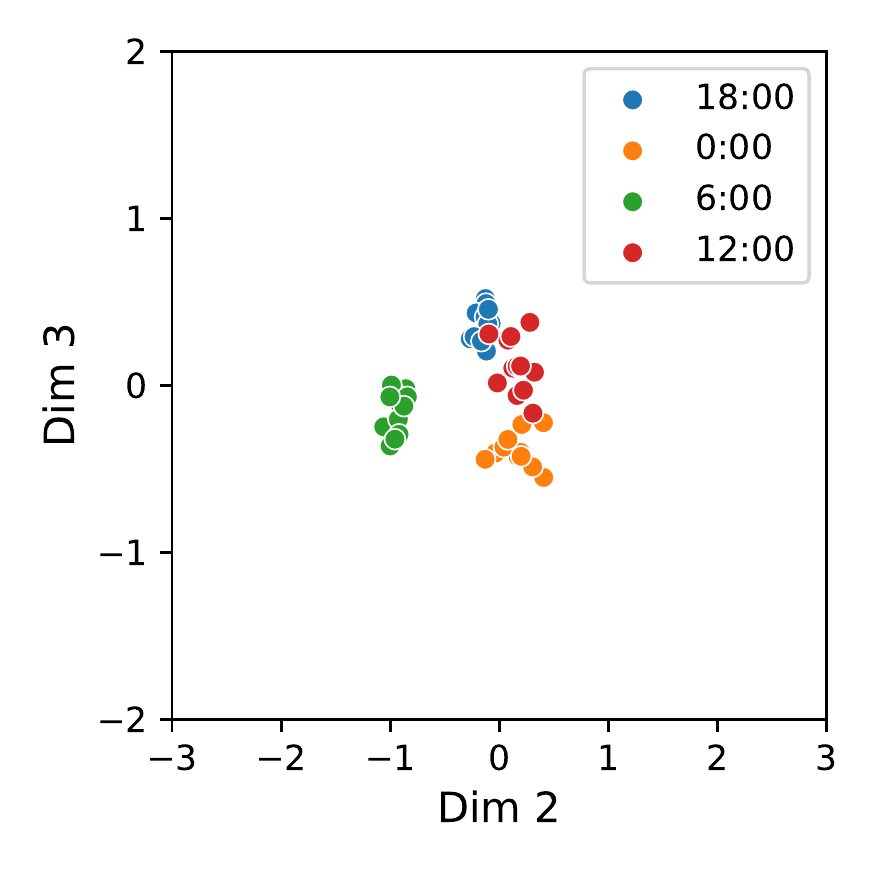}}
\caption{Projections of representations on the subspace spanned by (a), (b), (c): the first two components; (d), (e), (f):  the third component and fourth component.}
\label{fig:pca_time}
\end{figure}

To demonstrate that the distinguishability of representations can be greatly enhanced by the additional operations, we employ raw sequence and representation produced by vanilla Wavenet as two benchmark representations. We firstly perform principal component analysis (PCA) aiming at reducing the dimensionality of the representations. The reason for adopting PCA rather than t-SNE is that PCA can preserve the linearity on the representation space. The percentage of variance explained by each of the selected components is plotted in Figure \ref{fig:pca_time_variance}, from which we can observe that only a few components are effective as the variance over the others is extremely small. According to this finding, we visualize in Figure \ref{fig:pca_time} the projections on the first four components which explain almost all the variance over the representations. It is obvious that the enhanced model produces representations showing the strongest intra-task correlation and the weakest inter-task correlation of the three models. Hence, the distinguishability from temporal view is improved by applying the proposed operations.

\begin{figure}[tb]
    \centering
\includegraphics[width=0.8\linewidth]{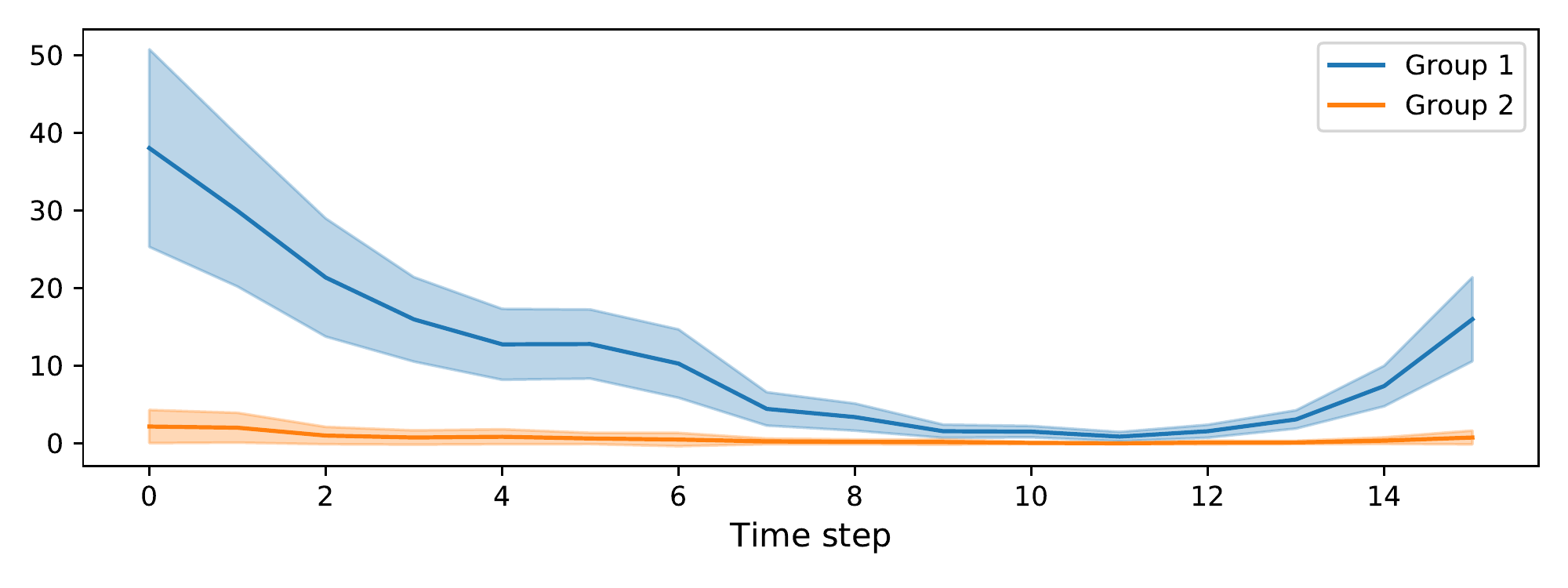}
\caption{Two groups of samples selected to inspect spatial distinguishability.}
\label{fig:pca_space_samples}
\end{figure}

\textbf{Spatial View.} In the second case, we examine the distinguishability from the spatial view, concentrating on a particular time. We roughly divide the regions into two groups based on the evolving patterns as shown in Figure \ref{fig:pca_space_samples}. Basically, bike demand over the first group of regions experiences a v-shape variation during the period being visualized and will continue rising in the forthcoming time steps; on the contrary, the second group of regions approximately remains constant e.g. 0. We clarify that the assignment is not rigorous and unique, but such fuzziness will not affect the conclusion we can draw.

\begin{figure}[tb]
\centering
\includegraphics[width=0.8\linewidth]{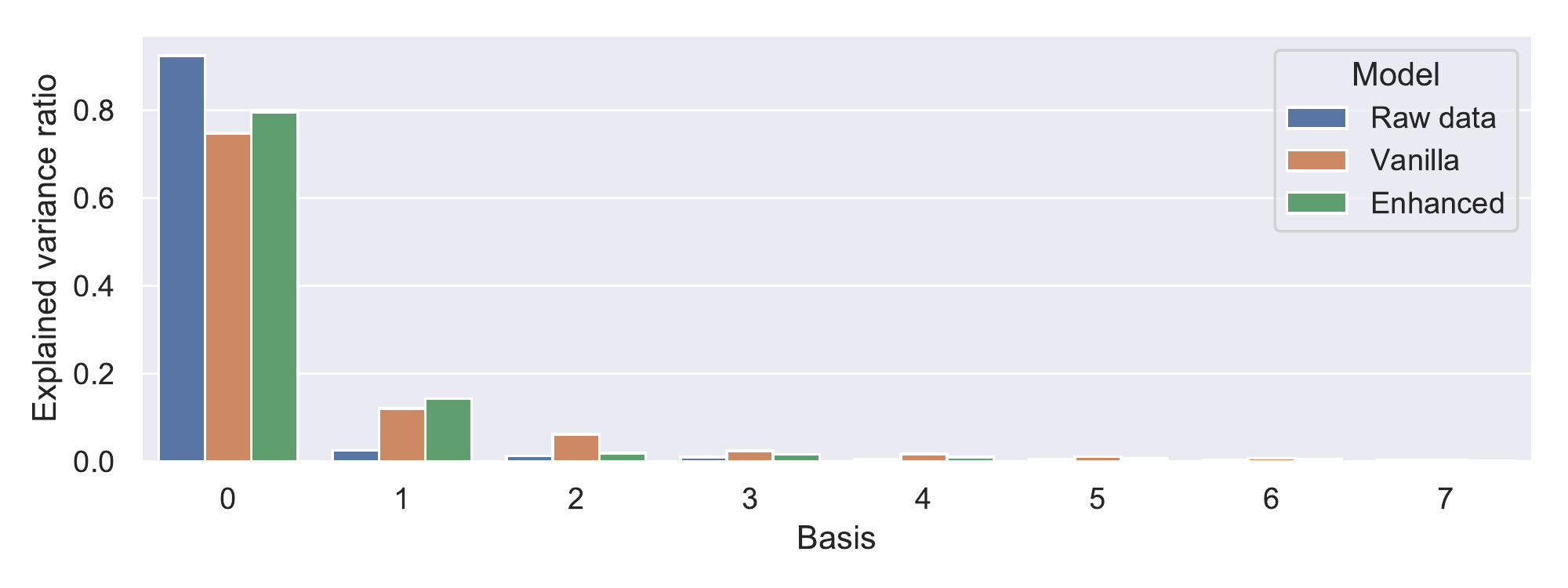}
\caption{The percentage of variance explained by the first seven components.}
\label{fig:pca_space_variance}
\end{figure}

\begin{figure}[tb]
\subfloat[Raw data]{
\includegraphics[width=0.32\linewidth]{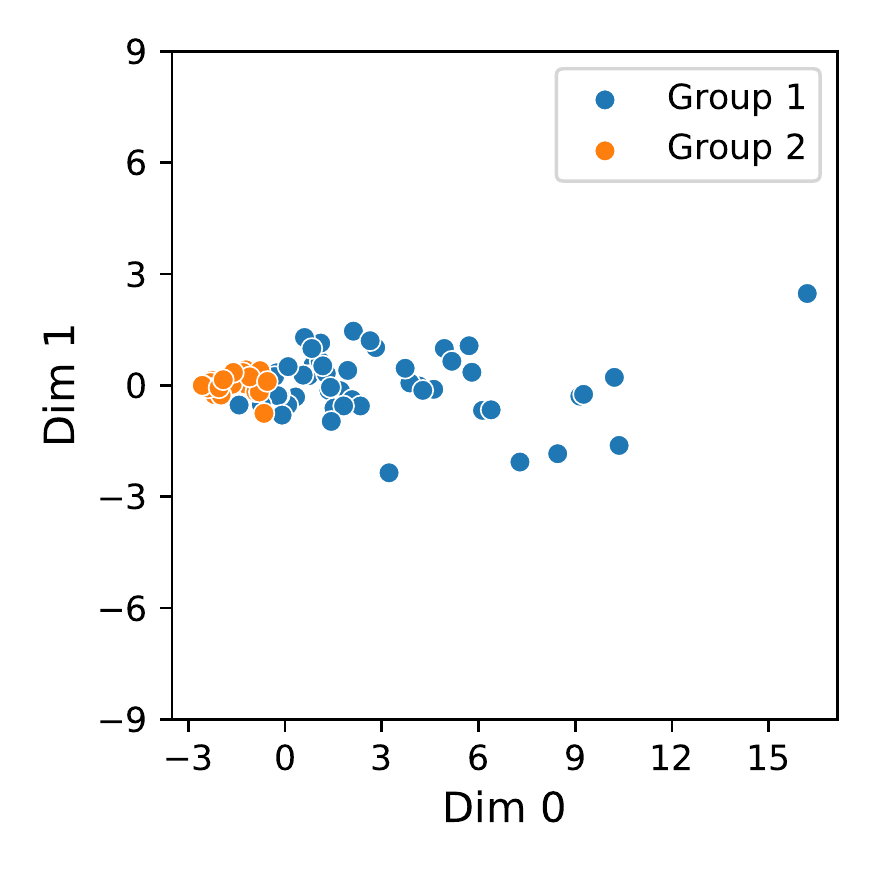}}
\subfloat[Vanilla]{
\includegraphics[width=0.32\linewidth]{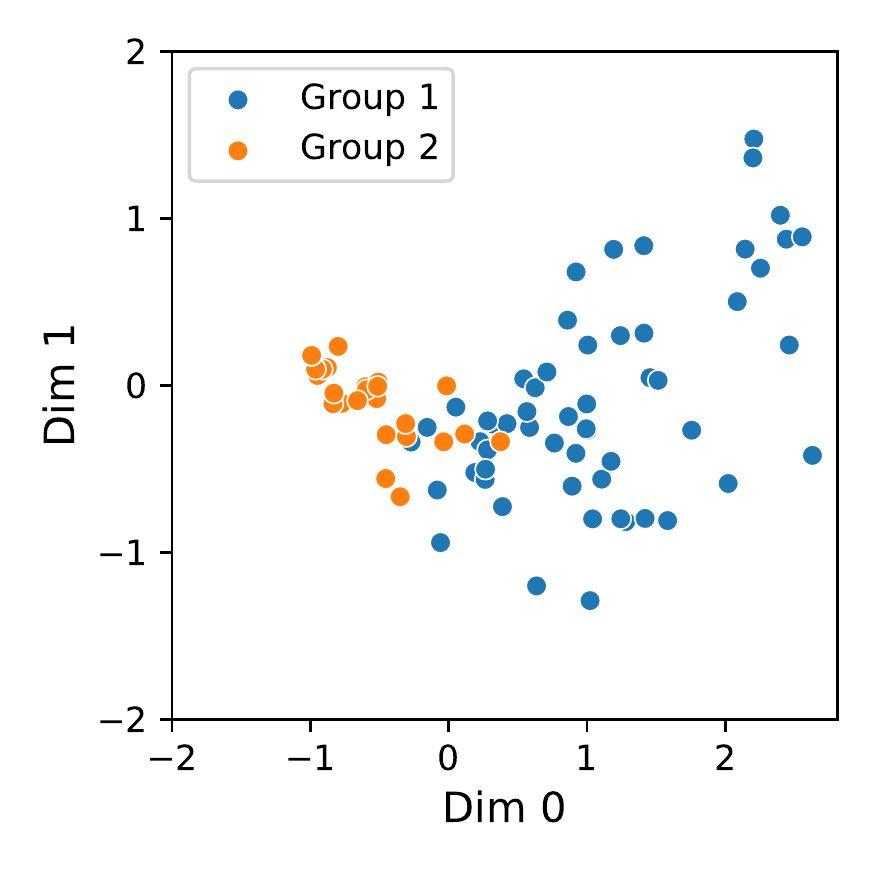}}
\subfloat[Enhanced]{
\includegraphics[width=0.32\linewidth]{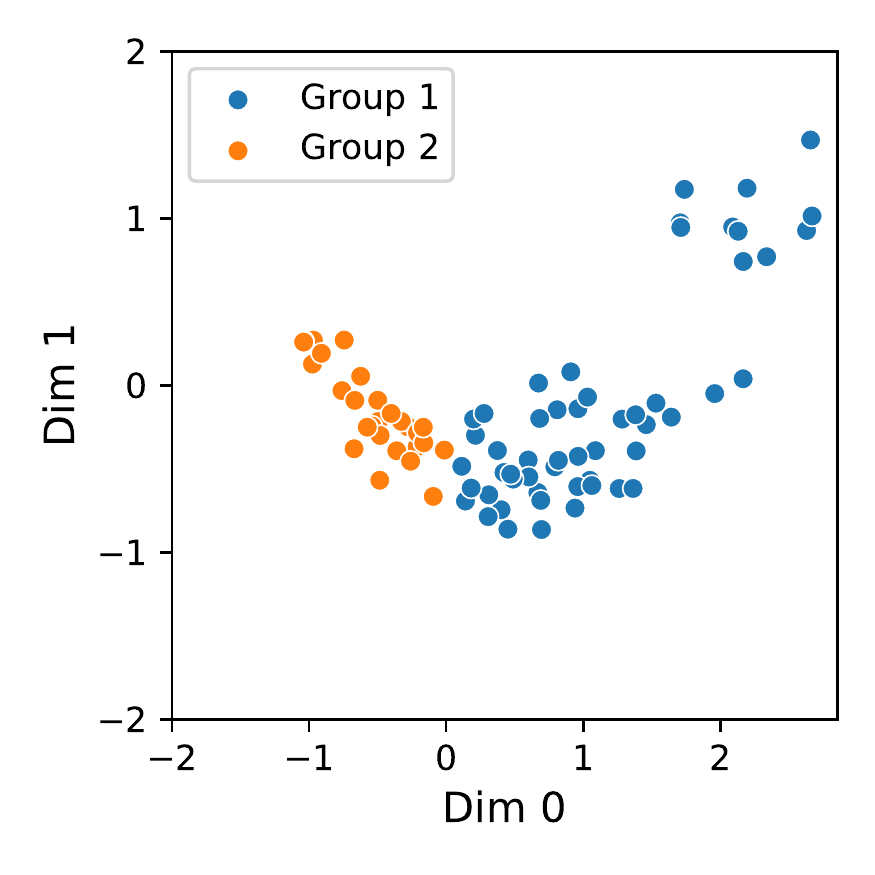}}
\caption{Projections of representations on the subspace spanned by the first two components.}
\label{fig:pca_space}
\end{figure}

\begin{figure}[tb]
\centering
\subfloat[Raw data]{
\includegraphics[width=0.32\linewidth]{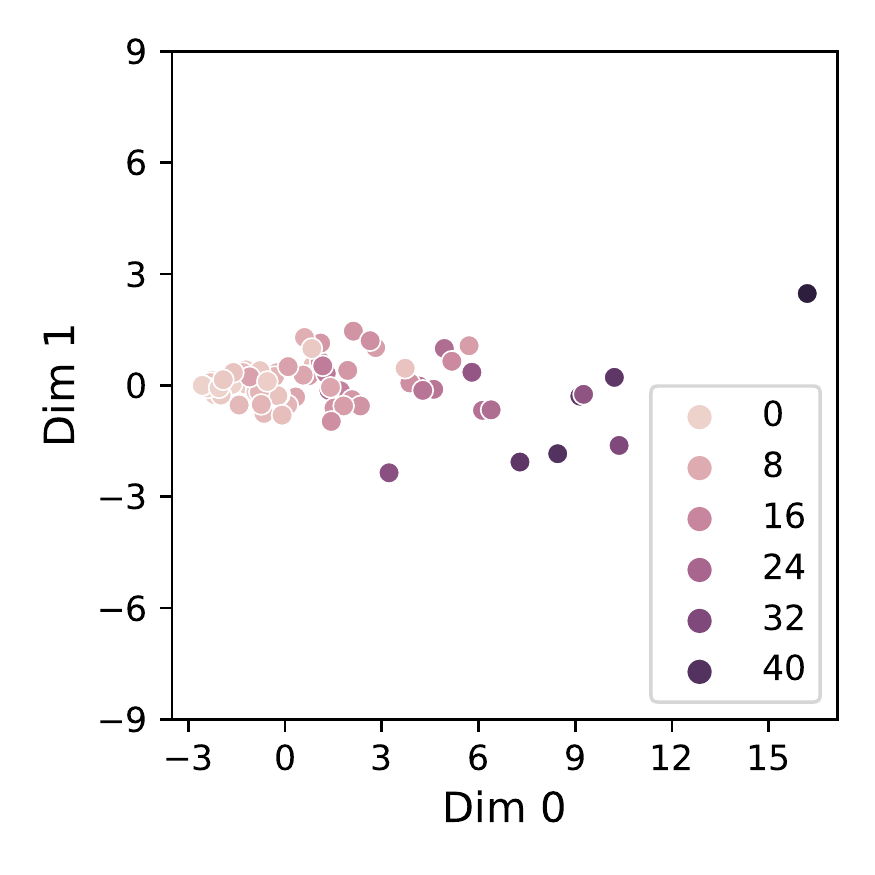}}
\subfloat[Vanilla]{
\includegraphics[width=0.32\linewidth]{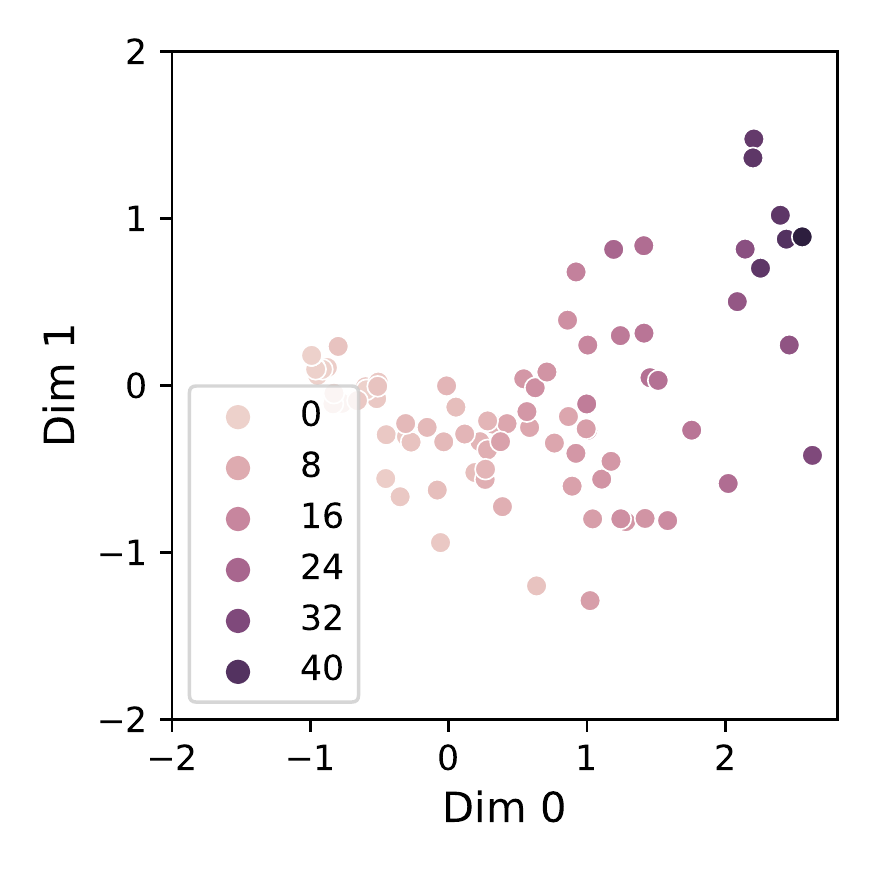}}
\subfloat[Enhanced]{
\includegraphics[width=0.32\linewidth]{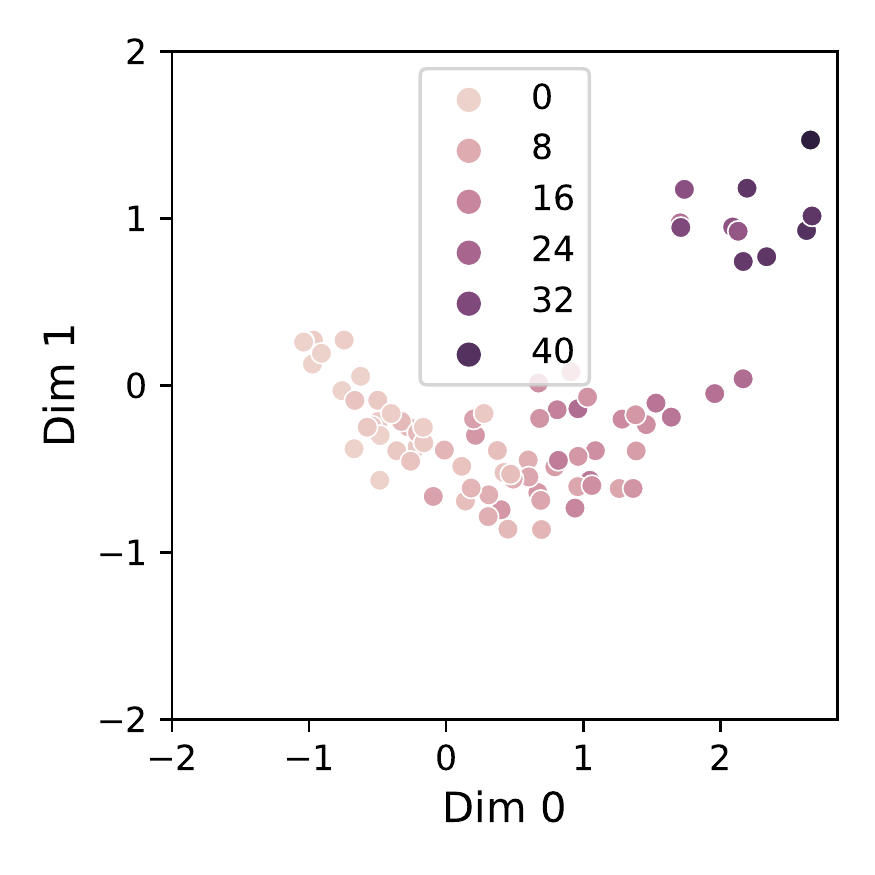}}
\caption{Projections of representations on the subspace spanned by the first two components.}
\label{fig:pca_scale_space}
\end{figure}

Similar to the first case, we perform PCA, plot the percentage of variance explained by each of the selected components in Figure \ref{fig:pca_space_variance} and visualize in Figure \ref{fig:pca_space} the projections on the first two components which explain most of the variance. At this step, it is apparent that the enhanced model yields the most distinguishable representation space from the spatial view. Moreover, we recolor the sample points based on the scale of the observations as shown in Figure \ref{fig:pca_scale_space}, and the visualization results show that the representations also encode the information on scale.

\section{Conclusion and Future Work}

In this work, we develop a novel multi-view multi-task learning framework for multi-variate time series forecasting. In our design, this framework consists of the original view, the spatial view and the temporal view, but it is flexible to account for more views depending on the specific application. Forecasting from each view is accomplished in a multi-task manner via two task-dependent operations, namely task-wise normalization and task-wise affine transformation. Diverse experiments were performed to quantitatively show the effectiveness, the efficiency and the robustness of the framework. Furthermore, we conducted multiple case studies on representations produced at different stages in the forecasting procedure. The outcome qualitatively demonstrates that this framework strengthens the intra-task correlation, while weakening the inter-task correlation.

In our future work, we will explore how to perform MTS forecasting under the circumstance of long-term pattern shift. Or in other words, the long-term pattern of each variable will change over time. In this new scenario, the current implementation of the MVMT framework will not be suitable or at least have large room to be improved, since the spatial-view affine transformation enforces constraint on a fixed long-term pattern for each variable.

\section{Acknowledgments}

This work was supported in part by ARC under Grants DP180100106 and DP200101328.

\bibliographystyle{IEEEtran}
\bibliography{sample-bibliography-biblatex}

\begin{IEEEbiography}[{\includegraphics[width=1in,height=1.25in,clip,keepaspectratio]{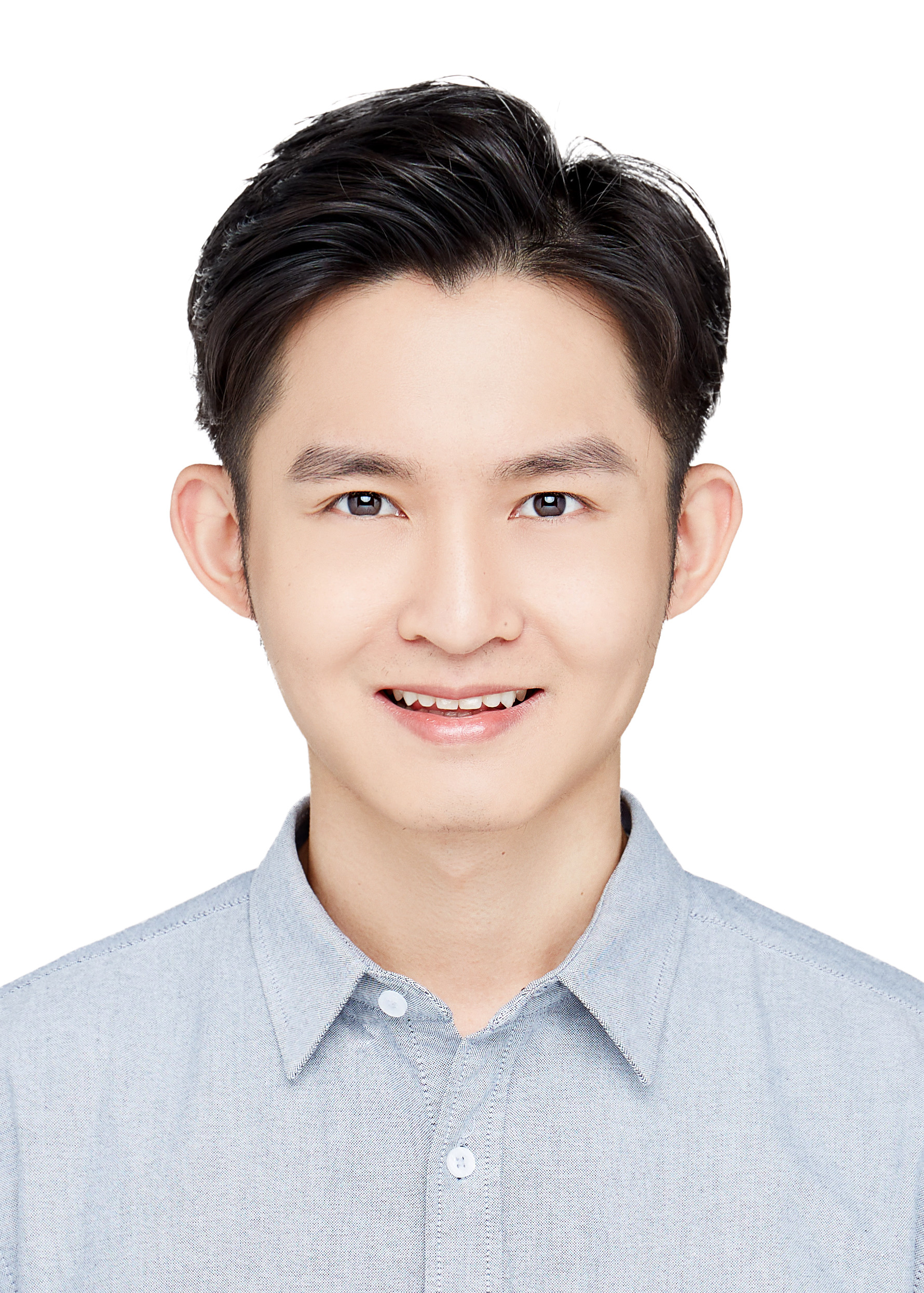}}]{Jinliang Deng} received a B.S.
degree in computer science from Peking University in 2017, and a M.S. degree in computer science from The Hong Kong University of Science and Technology in 2019. He is currently a Ph.D. candidate in the Australian Artificial Intelligence Institute, University of Technology Sydney and the Department of Computer Science and Engineering, Southern University of Science and Technology. His research interests include time series forecasting, urban computing and deep learning.
\end{IEEEbiography}

\begin{IEEEbiography}[{\includegraphics[width=1in,height=1.25in,clip,keepaspectratio]{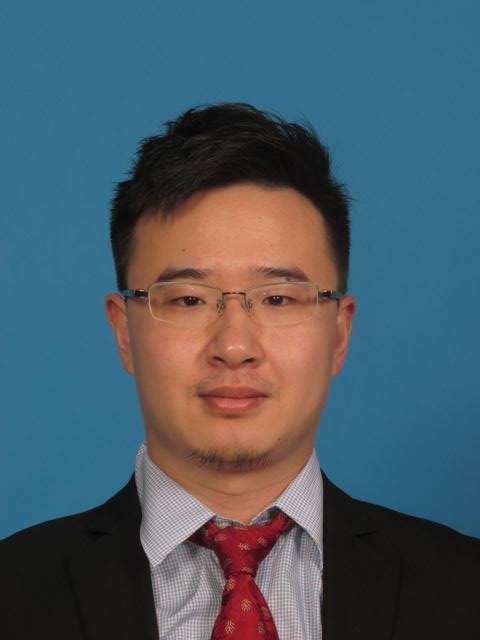}}]{Xiusi Chen} received a B.S.
degree and a M.S. degree in computer science from Peking University, in 2015 and 2018, respectively. He is currently a Ph.D. candidate in the Department of Computer Science, University of California, Los Angeles. His research interests include natural language processing, knowledge graph, neural maching reasoning and reinforcement learning.
\end{IEEEbiography}

\begin{IEEEbiography}[{\includegraphics[width=1in,height=1.25in,clip,keepaspectratio]{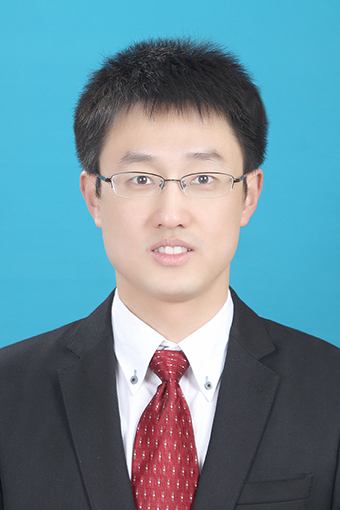}}]{Renhe Jiang} received a B.S. degree in software engineering from the Dalian University of Technology, China, in 2012, a M.S. degree in information science from Nagoya University, Japan, in 2015, and a Ph.D. degree in civil engineering from The University of Tokyo, Japan, in 2019. From 2019, he has been an Assistant Professor at the Information Technology Center, The University of Tokyo. His research interests include ubiquitous computing, deep learning, and spatio-temporal data analysis.
\end{IEEEbiography}

\begin{IEEEbiography}[{\includegraphics[width=1in,height=1.25in,clip,keepaspectratio]{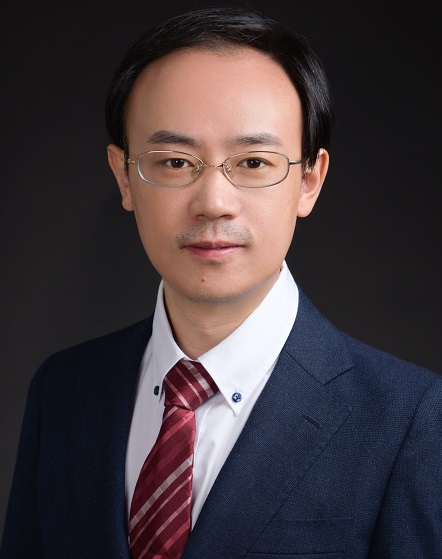}}]{Prof. Xuan Song} received a Ph.D. degree in signal and information processing from Peking University in 2010. In 2017, he was selected as Excellent Young Researcher of Japan MEXT. He has served as Associate Editor, Guest Editor, Area Chair, Senior Program Committee Member for many prestigious journals and top-tier conferences, such as IMWUT, IEEE Transactions on Multimedia, WWW Journal, ACM TIST, IEEE TKDE, Big Data Journal, UbiComp, IJCAI, AAAI, ICCV, CVPR etc. His main research interests are AI and its related research areas, such as data mining and urban computing. To date, he has published more than 100 technical publications in journals, book chapters, and international conference proceedings, including more than 60 high-impact papers in top-tier publications for computer science. His research has been featured in many Chinese, Japanese and international venues, including the United Nations, the Discovery Channel, and Fast Company Magazine. He received the Honorable Mention Award at UbiComp 2015.
\end{IEEEbiography}

\begin{IEEEbiography}[{\includegraphics[width=1in,height=1.25in,clip,keepaspectratio]{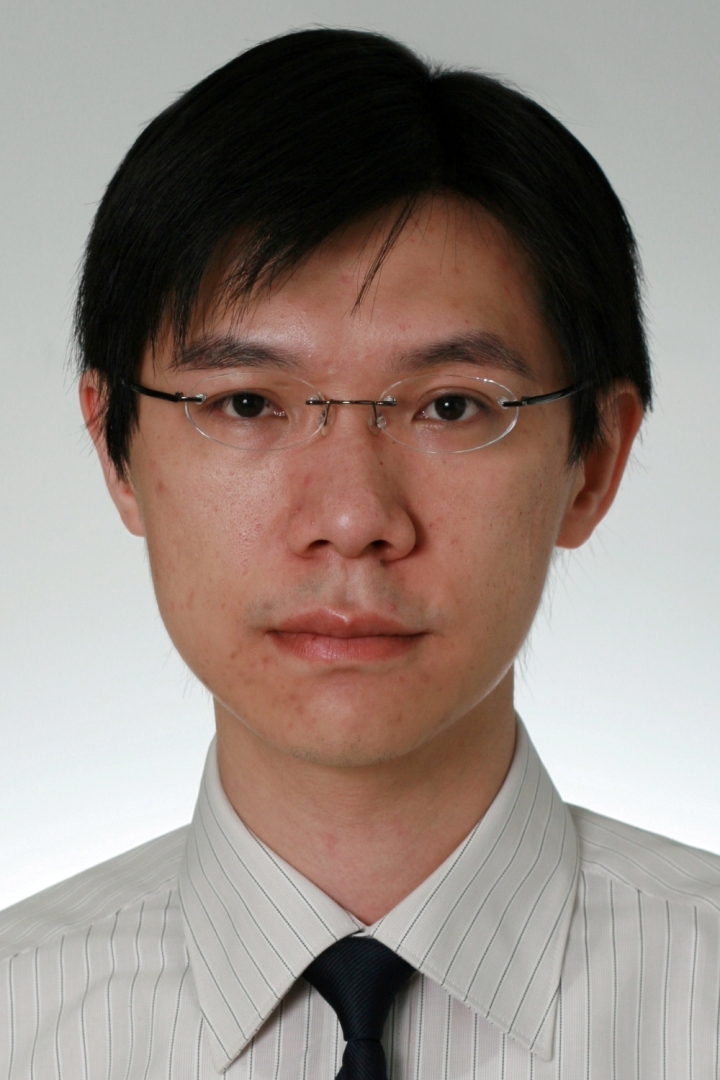}}]{Ivor W. Tsang}
is an ARC Future Fellow and Professor
of Artificial Intelligence with the University
of Technology Sydney (UTS), Australia. He is also
the Research Director of the Australian Artificial
Intelligence Institute. His research interests include
transfer learning, generative models, and big data analytics for data with extremely
high dimensions. In 2013, Prof Tsang received his
prestigious ARC Future Fellowship for his research
regarding Machine Learning on Big Data. In 2019,
his JMLR paper titled ”Towards ultrahigh dimensional
feature selection for big data” received the International Consortium
of Chinese Mathematicians Best Paper Award. In 2020, Prof Tsang was
recognized as the AI 2000 AAAI/IJCAI Most Influential Scholar in Australia
for his outstanding contributions to the field of AAAI/IJCAI between 2009
and 2019. His research on transfer learning granted him the Best Student
Paper Award at CVPR 2010 and the 2014 IEEE TMM Prize Paper Award. In
addition, he received the IEEE TNN Outstanding 2004 Paper Award in 2007.
He serves as a Senior Area Chair/Area Chair for NeurIPS, ICML, AISTATS,
AAAI and IJCAI, and the Editorial Board for JMLR, MLJ, and IEEE TPAMI.

\end{IEEEbiography}

\end{document}